\documentclass[10pt,journal,compsoc]{IEEEtran}

\usepackage[nocompress]{cite}
\usepackage{amsmath,amssymb,amsthm,bm}
\usepackage{booktabs,tabularx,array,multirow,makecell}
\usepackage{xcolor}
\usepackage{pifont}
\usepackage{graphicx}
\usepackage{url}
\usepackage{algorithm,algorithmic}
\usepackage[hidelinks,hypertexnames=false]{hyperref}

\newtheorem{definition}{Definition}
\newtheorem{lemma}{Lemma}
\newtheorem{proposition}{Proposition}
\newtheorem{theorem}{Theorem}
\newtheorem{corollary}{Corollary}
\theoremstyle{remark}
\newtheorem{remark}{Remark}

\newcommand{\R}{\mathbb{R}}
\newcommand{\Tr}{^{\top}}
\newcommand{\II}{\mathrm{II}}
\newcommand{\Nm}{N$\cdot$m}
\newcolumntype{Y}{>{\raggedright\arraybackslash}X}
\newcolumntype{L}[1]{>{\raggedright\arraybackslash}p{#1}}


\edef\FLSdbltopnumber{\the\value{dbltopnumber}}
\begin{document}
\makeatletter
\let\FLSmaketitle\maketitle
\let\FLS@maketitle\@maketitle
\let\FLSthanks\thanks
\edef\FLStabcolsep{\the\tabcolsep}
\makeatother
\bstctlcite{IEEEexample:BSTcontrol}

\title{FoLaSim: A Form--Law Twin Simulator for Dynamics Understanding in World Action Models}

\author{Qibin~Gu and Yunfeng~Yan%
\IEEEcompsocitemizethanks{\IEEEcompsocthanksitem Q.~Gu is with Nanjing Tech University, Nanjing, China.\protect\\
E-mail: guqibin123@njtech.edu.cn
\IEEEcompsocthanksitem Y.~Yan is with Zhejiang University, Hangzhou, China.\protect\\
E-mail: yyff@zju.edu.cn}}

\markboth{IEEE Transactions on Pattern Analysis and Machine Intelligence}%
{FoLaSim: A Form--Law Twin Simulator for World Action Models}

\IEEEtitleabstractindextext{%
\begin{abstract}
World action models increasingly act and plan through forecasts, yet a plausible video can hide whether a forecast responds correctly to physical interventions. We present FoLaSim, the first form--law twin simulator, making physical law controllable across visible forms and measurable from predicted motion. Its forward half synthesizes and certifies systems that share a law across forms or change the law within one form, including rigid robots in contact. Its inverse half reads state and video forecasts into unexplained effort in SI units and an implied law. Four complementary tests on these twins jointly characterize correctness within a declared law class. Across nineteen robots, certified law twins agree to machine precision in contact. With exact states, unexplained effort detects faults with larger margins than trajectory error, identifies laws through unknown contact forces, and anticipates decision regret. Canonical-coordinate models benefit from law-matched evidence, whereas four native video models fail to adjust to changed laws in controlled out-of-distribution tests. Training studies show how law variation and excitation shape the use of motion and effort. On an arm corpus that varies the law within one form, a compact response model reads the law from evidence, with mean error below $0.06^\circ$ and millisecond batch inference; trained on one law, it ignores this evidence. With frozen weights, it predicts real arm motion and selects commands that reduce measured cost by 23\% across three reaching tasks. On the real arm, a payload's law appears in the servo effort, not the motion.
\end{abstract}

\begin{IEEEkeywords}
World action models, world models as simulators, robot dynamics, physical simulation, dynamics understanding, form--law twins, Gauss's principle.
\end{IEEEkeywords}}

\maketitle
\IEEEdisplaynontitleabstractindextext
\IEEEpeerreviewmaketitle

\IEEEraisesectionheading{\section{Introduction}\label{sec:intro}}

\IEEEPARstart{W}{orld} action models (WAMs) predict how scenes evolve under robot actions, either conditioned on those actions or generating them jointly~\cite{yang2024unisim,bruce2024genie,agarwal2025cosmos,du2023unipi,assran2025vjepa2,li2026causal}. These models have quickly moved from prediction into the control loop. They act as zero-shot policies~\cite{dreamzero2026,fastwam2026,gigaworldpolicy2026,tau0wm2026}, select actions before execution~\cite{acid2026,wise2026}, and evaluate policies through predicted rollouts~\cite{worldeval2025,worldgym2025,ctrlworld2025,dworldeval2026}. They also serve as learned simulators for reinforcement learning of vision--language--action policies~\cite{cosmospredict2025,cosmos3_2026,pelicansim2026,wmpo2025,worldenv2025,vlarft2025}. All of these uses rest on one requirement: forecasts must respond correctly to applied forces. A planner comparing two torque programs, for example, needs the difference between their predicted futures to reflect the difference between their physical effects.

\begin{figure}[t]
\centering
\includegraphics[width=\columnwidth]{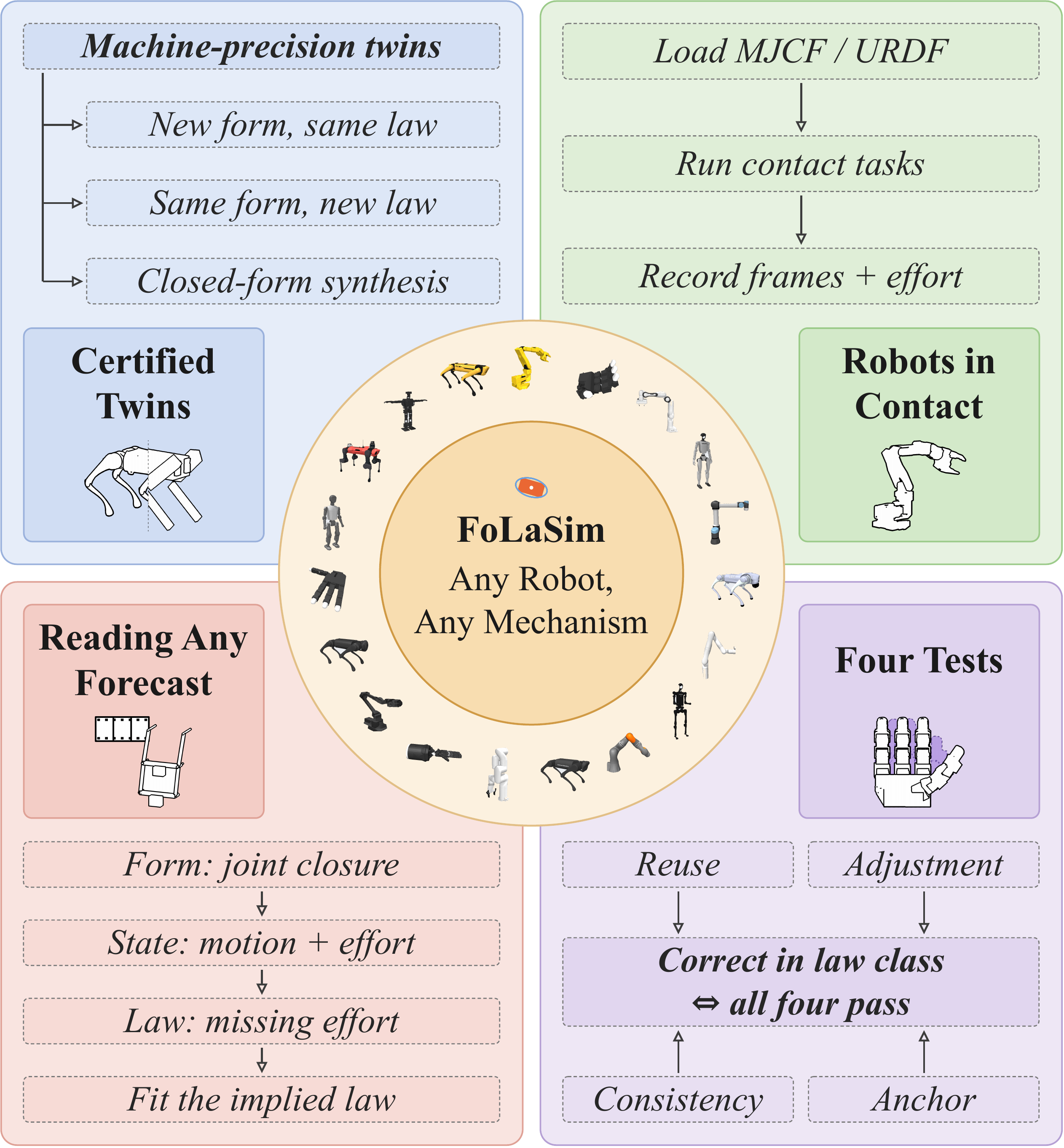}
\caption{FoLaSim connects robot descriptions, certified twins, and physical diagnosis. The center links the four modules. At top left, law twins change form while preserving dynamics; form twins change dynamics within one form. At top right, MJCF/URDF robots run with effort recorded beside each frame. The bottom-left reader converts forecasts into joint verdicts, unexplained effort, and an implied law. The bottom-right tests assess reuse, adjustment, consistency, and the physical anchor. The Go2 example contrasts the original with its equimomental twin.}
\label{fig:folasim}
\end{figure}

Visual plausibility, however, gives only a partial account of this ability. Video models weight color, size, velocity, and shape differently when they generalize~\cite{kang2025howfar}. Action-conditioned models struggle with articulated motion~\cite{acwmphys2026} and transfer across embodiments by visual rather than kinematic similarity~\cite{xeworld2026}, and in joint video--action models, action accuracy depends on video quality~\cite{gigaworldpolicy2026}. Recent approaches therefore separate appearance from dynamics in model design~\cite{mask2real2026,xeworld2026} and training data~\cite{phyparam2026,phyco2026}. Yet a forecast can look intact and still move wrongly. A learned Lagrangian model, for instance, can preserve the joints of a closed chain while falling far behind its true motion~\cite{paper1}. Benchmarks of physical plausibility~\cite{bansal2024videophy,meng2024phygenbench,motamed2026physicsiq} and embodied prediction~\cite{ewmbench2025,worldarena2026,smallworlds2025} mainly evaluate what a forecast shows. A deeper question thus remains open: \emph{has a world action model learned the law that turns effort into motion, or the form visible in its frames?}

A single forecast cannot answer this question, because form and law are entangled in every frame. A mechanism's \emph{form} describes how its bodies look and are joined. Its \emph{law} consists of the inertia, potential, dissipation, and input port that determine its response to effort. Different forms can realize the same law, and identical forms can obey different laws. Observation adds a second ambiguity. Free motion reveals the law only up to scale, while tight tracking moves information about the law from motion into actuator effort (Section~\ref{sec:invariances}). An evaluation must therefore control both the relation between form and law and the channel through which the law is observed.

Our central idea is to separate these factors with \emph{certified twins}. Law twins keep the law and change the form; form twins keep the form and change the law. Crossing the two lets us hold a query fixed and change only whether its evidence matches in form or in law. Such an evaluation needs (i) certified twins, (ii) known effort recorded with each frame, (iii) a physical reading of the forecast against a measurement floor, and (iv) tests that jointly establish correctness. Physics engines provide known efforts at scale~\cite{todorov2012mujoco,mujocoplayground2025,isaaclab2025}, and physics data engines vary mass, friction, force, or gravity within a scene~\cite{phyparam2026,phyco2026,physeditworld2026,physinone2026}. FoLaSim adds the two missing pieces, certified law twins and a characterization of correctness, and unites all four components.

We introduce FoLaSim, the first \emph{form--law twin simulator}, built on the unmodified MuJoCo engine~\cite{todorov2012mujoco}. Its forward half makes physical law controllable, and its inverse half makes it measurable (Fig.~\ref{fig:folasim}). The forward half solves for rigid parts that realize a target law on different topologies. For robots, a closed-form construction produces certified twins directly from MJCF or URDF descriptions, including contact scenes. Certification bounds the mismatch between the simulated systems, so comparing a predictor across twins isolates its response to form and to law.

The inverse half travels the opposite way, from a forecast back to a physical diagnosis. It checks joint closure, recovers the motion together with its effort, and measures the torque left unexplained by the known law, in \Nm{} against a measurement floor. From the same motion, it also identifies the implied law. Across twins, four tests then assess transfer between forms (\emph{reuse}), response to a changed law (\emph{adjustment}), a shared explanation across interventions (\emph{consistency}), and agreement with the true law under known effort (\emph{anchor}). Together, they characterize correctness within a declared law class (Theorem~\ref{thm:characterization}). They also expose misleading successes, such as a nearly stationary predictor that appears invariant across twins.

Our experiments first establish that both halves are reliable. Twin generation stays faithful from designed mechanisms to robots in contact. Unexplained effort exposes dynamic faults that tracking can conceal, identifies robot laws through unknown contact forces, and anticipates decision regret.

With these tools in place, we study how models use physical evidence. Canonical-coordinate context models transfer law information across visibly different mechanisms. In contrast, the four tested native video models fail to adjust to a changed law and favor evidence that matches the query's form. Controlled training corpora then reveal where effective transfer comes from. Variation of the law within a form promotes law-sensitive prediction, while the excitation decides whether motion or effort carries the useful information. Finally, the same principle yields a compact response model that reads the law from evidence. It trades accuracy for online cost favorably against calibrated physics engines (Section~\ref{sec:efficiency}) and transfers to a real arm, where a payload's law appears in the servo effort rather than in the motion (Section~\ref{sec:real}).

In summary, this work makes four contributions:
\begin{itemize}
\item A novel form--law twin simulator that synthesizes and certifies different forms of a target law. Its forward construction yields an unbounded set of certified twins, with closed-form twins for rigid robots on fixed bases or in frictional contact (Section~\ref{sec:constructions}).
\item The first form-to-law reading of world-model forecasts. It combines unexplained effort in SI units, per-forecast measurement floors, and implied-law identification under unknown contact forces, and four complementary tests jointly characterize correctness within the declared law class (Sections~\ref{sec:why} and~\ref{sec:measurement}).
\item An extensive study of eleven mechanisms, nineteen robots, and state, image, and native video models. It separates evidence use from physical correctness and identifies how training data make the law accessible (Section~\ref{sec:experiments}).
\item An open-source platform for scalable data generation and model evaluation, with automatic form generation, MJCF/URDF import, and batched, differentiable execution (Supplementary Section~\ref{sup:software}).
\end{itemize}

\section{Related Work}\label{sec:related}

\begin{table*}[t]
\centering
\caption{Comparison of FoLaSim with representative simulators, data engines, and benchmarks, by the four components (i)--(iv) of Section~\ref{sec:intro}. $\checkmark$: provided as part of the method; (iv): how many of the four tests of Definition~\ref{def:capabilities} the approach applies. Supplementary Table~\ref{tab:capabilities-compared} gives, for each kind of approach, what it provides instead.}
\label{tab:compare}
\footnotesize
\setlength{\tabcolsep}{4.5pt}
\begin{tabular*}{\textwidth}{@{\extracolsep{\fill}}lccccccccc@{}}
\toprule
 & & \multicolumn{3}{c}{(i) Instances} & (ii) & \multicolumn{3}{c}{(iii) Reading a forecast} & (iv) \\
\cmidrule(lr){3-5}\cmidrule(lr){7-9}
Approach & \makecell{Renders\\video} & \makecell{Law varied,\\form fixed} & \makecell{Form varied,\\law fixed} & \makecell{Certified\\twins} & \makecell{Effort with\\every frame} & \makecell{Scores a\\forecast} & \makecell{In SI units,\\with a floor} & \makecell{Reads back\\the law} & \makecell{Tests\\applied} \\
\midrule
MuJoCo~\cite{todorov2012mujoco} & $\checkmark$ &  &  &  & $\checkmark$ &  &  &  &  \\
Isaac Lab~\cite{isaaclab2025} & $\checkmark$ &  &  &  & $\checkmark$ &  &  &  &  \\
\midrule
PhysEditWorld~\cite{physeditworld2026} & $\checkmark$ & $\checkmark$ &  &  &  &  &  &  &  \\
PhysInOne~\cite{physinone2026} & $\checkmark$ & $\checkmark$ &  &  &  &  &  &  &  \\
\midrule
VideoPhy~\cite{bansal2024videophy} &  &  &  &  &  & $\checkmark$ &  &  &  \\
Physics-IQ~\cite{motamed2026physicsiq} &  &  &  &  &  & $\checkmark$ &  &  &  \\
EWMBench~\cite{ewmbench2025} &  &  &  &  &  & $\checkmark$ &  &  &  \\
WorldArena~\cite{worldarena2026} &  &  &  &  &  & $\checkmark$ &  &  &  \\
XEWorld~\cite{xeworld2026} &  &  &  &  &  & $\checkmark$ &  &  &  \\
\midrule
PokeWorld~\cite{pokeworld2026} &  & $\checkmark$ &  &  &  & $\checkmark$ &  &  &  \\
CALIPER~\cite{caliper2026} &  & $\checkmark$ &  &  &  & $\checkmark$ &  &  &  \\
\midrule
ACID~\cite{acid2026} &  &  &  &  &  & $\checkmark$ &  &  &  \\
OneWorld~\cite{oneworld2026} &  &  &  &  &  & $\checkmark$ &  &  & 1 \\
IMPLY~\cite{imply2026} &  &  &  &  &  & $\checkmark$ &  & $\checkmark$ & 2 \\
Closed-chain evaluator~\cite{paper1} &  &  &  &  & $\checkmark$ & $\checkmark$ & $\checkmark$ &  &  \\
\midrule
\textbf{FoLaSim} & $\checkmark$ & $\checkmark$ & $\checkmark$ & $\checkmark$ & $\checkmark$ & $\checkmark$ & $\checkmark$ & $\checkmark$ & \textbf{4} \\
\bottomrule
\end{tabular*}
\end{table*}

\subsection{Simulators, Data Engines, and Benchmarks}
Physics engines generate motion from known laws at scale~\cite{todorov2012mujoco,mujocoplayground2025,isaaclab2025} and form the foundation of robot-learning frameworks~\cite{dmcontrol2020,robosuite2020}. Physics data engines build on them by varying forces, masses, friction, or gravity within a fixed form~\cite{phyparam2026,phyco2026,physeditworld2026,physinone2026}, and real-to-sim methods reconstruct physical twins of observed objects~\cite{phystwin2025}. None of these tools, however, varies the form while holding the law fixed. FoLaSim adds this complementary operation by constructing visibly different mechanisms that realize the same certified law. Combined with form twins, these law twins yield controlled comparisons along both dimensions (Table~\ref{tab:compare}).

Evaluation methods approach physical prediction from several directions. Video benchmarks assess plausibility~\cite{bansal2024videophy,meng2024phygenbench,motamed2026physicsiq,upadhyay2026worldbench}, and embodied benchmarks assess world-model fidelity~\cite{ewmbench2025,worldarena2026,acwmphys2026,smallworlds2025}. Controlled studies go further: they vary mass, stiffness, or physics under identical appearance~\cite{pokeworld2026,physicallyviable2026}, hold out embodiments~\cite{xeworld2026}, or provide calibration evidence in context~\cite{physicalworld2026,caliper2026}.

Physical diagnostics come closest to the dynamics by connecting predicted motion to its governing laws. Inverse dynamics checks transitions against their actions~\cite{acid2026,wise2026,wav2026}. Long-rollout studies distinguish kinematic behavior from dynamic consistency~\cite{kinematicnotdynamic2026}, and relational tests examine whether futures under different actions admit a common physics~\cite{oneworld2026,imply2026}. A model-agnostic closed-chain evaluator reports SI-unit errors against a measurement floor~\cite{paper1}. FoLaSim couples such a physical readout to certified changes in form and law. As a result, the four tests of Section~\ref{sec:intro} explain both individual forecast errors and how these errors relate across mechanisms and robots.

\subsection{Mechanical Structure and Identifiability}
Mechanical structure is a useful inductive bias for prediction, and Lagrangian, Hamiltonian, and constrained models encode it directly~\cite{lutter2019delan,cranmer2020lagrangian,greydanus2019hamiltonian,daems2024keycld}. Whether its physical parameters can be identified, however, depends on the observations and the excitation~\cite{phast2026}. Closed-loop data require external excitation~\cite{ljung1999system,forssell1999closed}, and controlled latent world models require state-conditional action variation~\cite{controlledwmid2026}. FoLaSim makes these conditions explicit through recorded effort, controlled interventions, and a physical anchor for the inferred structure.

Robot dynamics are linear in the inertial parameters, which enables dynamic identification~\cite{atkeson1986estimation,gautier1990minimum,swevers1997optimal} with convex physical-consistency constraints~\cite{wensing2018lmi,lee2020geometric}. FoLaSim exploits this structure in both directions. Synthesis prescribes a law and solves for the parts that realize it, which reverses identification. The inverse readout identifies the law while treating contact forces as unknowns. Both operations build on classical constrained and port-based mechanics~\cite{gauss1829,bullo2005geometric,vanderschaft2014port}. Supplementary Section~\ref{sup:related} provides an extended review.

\section{Form, Law, and Twins}\label{sec:why}

We now make the question of Section~\ref{sec:intro} precise: we define form, state, and law spaces, examine what motion and effort reveal about the law, and use twins to build four complementary tests. Table~\ref{tab:notation} summarizes the notation, and Supplementary Sections~\ref{sup:why} and~\ref{app:proofs} give derivations and proofs.

\begin{table}[t]
\centering
\caption{Notation. Subscript $S$ marks a quantity reduced to the degrees of freedom of a mechanism.}
\label{tab:notation}
\footnotesize
\setlength{\tabcolsep}{4pt}
\begin{tabularx}{\columnwidth}{@{}lY@{}}
\toprule
Symbol & Meaning \\
\midrule
$q\in Q$, $\phi(q)=0$ & poses of all bodies; joint and transmission constraints \\
$S$, $s\in\R^d$, $N$ & admissible poses, their coordinates, tangent basis $\partial q/\partial s$ \\
$u$ & effort of the actuators, in SI units \\
$M_S$, $C_S$ & kinetic-energy metric (mass matrix) and its velocity terms \\
$V_S$, $d_S$, $B_S$ & potential, dissipation, and input port \\
$\lambda$, $a^\ast_\lambda$ & law $(M_S,V_S,d_S,B_S)$ and the acceleration it gives \\
$c$ & density factor of a form twin \\
$(\psi,K)$ & certified correspondence of law twins: coordinates, inputs \\
$\omega=(m,\lambda)$ & instance: mechanism $m$ realizing law $\lambda$ \\
$\iota$, $e$ & intervention (state, program, channel); evidence \\
$\hat a$, $\Theta$ & acceleration field implied by a model; declared law class \\
$\rho$, $r$ & Gauss residual and its norm in the kinetic metric \\
$\Lambda$, $\delta u$ & response to effort; effort left unexplained (\Nm{}) \\
$H$, $r_E$ & stored energy; power residual \\
$E_{\rm ch}$ & error of a predicted change of law \\
$\gamma$ & gain scale of a tracking controller \\
\bottomrule
\end{tabularx}
\end{table}

\subsection{Form, State, and Law}\label{sec:manifold}
We start from a robot--object system whose bodies have poses $q\in Q$, with mass matrix $M(q)\succ0$ and ideal bilateral constraints $\phi(q)=0$ for its joints and transmissions. The admissible poses form the constraint manifold $S=\phi^{-1}(0)$, with coordinates $s\in\R^d$ for its degrees of freedom, poses $q=q(s)$, and tangent basis $N=\partial q/\partial s$. D'Alembert's principle eliminates the joint reactions and leaves the dynamics on $S$~\cite{bullo2005geometric},
\begin{equation}
M_S(s)\ddot s+C_S(s,\dot s)\dot s+\nabla V_S(s)+d_S(s,\dot s)=B_S(s)\,u,
\label{eq:reduced}
\end{equation}
where $M_S=N\Tr MN$ is the kinetic-energy metric on $S$, $C_S$ collects its Christoffel symbols, $V_S$ is the potential, $d_S$ the dissipation, and $B_S$ the input port for the effort $u$.

The \emph{form} of a mechanism consists of its visible appearance and the embedding $s\mapsto q(s)$, that is, its bodies, their connections, and the configurations the assembly admits. Its \emph{law} is $\lambda=(M_S,V_S,d_S,B_S)$ on a declared domain and in declared coordinates, and it determines the acceleration $a^\ast_\lambda(s,\dot s,u)$ in \eqref{eq:reduced}. Together, form and law define the physical chain from effort to frames. The law produces motion in $s$, the embedding turns that motion into body poses, and rendering turns the poses into images.

The three spaces follow this chain. \emph{Form space} contains images and their geometric properties. \emph{State space} contains motion $(s,\dot s)$ together with the query effort $u$. \emph{Law space} contains acceleration fields $a(s,\dot s,u)$, compared in the kinetic-energy norm. The forward half realizes a law as visible motion; the inverse half uses that motion and its effort to assess the law.

\subsection{Observability of the Law}\label{sec:invariances}
\label{sec:split-fact}
Respecting the form and respecting the law are different conditions. A forecast respects the form when it stays on $S$, i.e., it satisfies the closure $\phi(\hat q)=0$ and the tangency $J\dot{\hat q}=0$, with $J=\partial\phi/\partial q$. It respects the law only if, in addition, the residual of \eqref{eq:reduced} vanishes. We call
\begin{equation}
r^2=\rho\Tr M_S^{-1}\rho,\qquad \rho=M_S\ddot{\hat s}+C_S\dot{\hat s}+\nabla V_S+d_S-B_Su,
\label{eq:residual}
\end{equation}
the \emph{Gauss residual}. Gauss's formula splits acceleration on $S$ into a normal part fixed by the constraints and a tangential part fixed by the applied forces~\cite{gauss1829,bullo2005geometric}. The residual $r$ measures the tangential mismatch in the kinetic-energy metric, independently of the coordinates or the tangent basis $N$. This split explains why intact geometry can coexist with incorrect dynamics. Re-timing an admissible path as $\hat q(\sigma(t))$ keeps it admissible, and a force error $J\Tr\mu$ along the reaction directions leaves the motion unchanged. Joint closure alone therefore cannot tell whether a mechanism moves at the physically correct speed.

Effort removes ambiguities that motion alone leaves open. Multiplying $(M_S,V_S,d_S)$ by a common factor $c$ preserves every free motion but changes the response $M_S^{-1}B_S$ to a given effort by $1/c$. Tracking introduces a different ambiguity, because the controller pulls the motion toward the commanded reference. With controller gain scale $\gamma$, the Fisher information about the law in the motion decays as $F_{\rm motion}=O(\gamma^{-2})$, whereas the information in actuator effort approaches a gain-independent limit, $F_{\rm effort}=O(1)$ (Supplementary Theorem~\ref{thm:channels}; see also~\cite{ljung1999system,forssell1999closed}). Tighter tracking thus moves information from the motion into the effort. Identifying the law requires known effort and informative excitation, which is why controlled changes across twins are essential to the evaluation.

\subsection{Law Twins and Form Twins}\label{sec:moves}
Twins turn this separation into an experiment. We call two mechanisms, or two instances of one mechanism, \emph{twins} when they are certified to share form or law while differing in the other. Section~\ref{sec:constructions} constructs both types as finite rigid assemblies.

\begin{definition}[Law twins: same law, different form]\label{def:equivalence}
Mechanisms $A$ and $B$ with laws $\lambda_A$ on $D_A\subset\R^d$ and $\lambda_B$ on $D_B$ are law twins under $(\psi,K)$ if $\psi:D_A\to D_B$ is a diffeomorphism, $K$ is an invertible input map ($u_B=Ku_A$), and for all $s\in D_A$ and $\dot s$:
\begin{align}
&\text{(T1)}\quad M_A(s)=D\psi(s)\Tr M_B(\psi(s))\,D\psi(s),\nonumber\\
&\text{(T2)}\quad V_A(s)=V_B(\psi(s))+\text{const},\nonumber\\
&\text{(T3)}\quad d_A(s,\dot s)=D\psi(s)\Tr d_B\big(\psi(s),D\psi(s)\dot s\big),\nonumber\\
&\text{(T4)}\quad B_A(s)=D\psi(s)\Tr B_B(\psi(s))\,K.\nonumber
\end{align}
\end{definition}

Condition (T1) makes $\psi$ an isometry of the kinetic metrics. Conditions (T2)--(T4) then match the potential, dissipation, and input port, so the two motions correspond under $(\psi,K)$. The twins may differ in their embeddings, reactions, and appearance, yet they share the Gauss residual and the power balance. Geometric residuals, in contrast, need not be preserved. Because the correspondence is fixed during construction, it provides an independent reference for comparing forecasts.

\begin{definition}[Form twins: same form, different law]\label{def:formtwins}
Instances $(m,\lambda)$ and $(m,\lambda')$ of one mechanism $m$ are form twins if they share the embedding and the appearance of every body and $\lambda'\neq\lambda$. Scaling the density of every moving body by $c\neq1$, with damping and friction scaled alike, gives the form twin $\lambda'=(cM_S,cV_S,cd_S,B_S)$.
\end{definition}

Density form twins mirror law twins. They keep appearance and free motion but change the response to effort. Combined with law twins, they support a controlled comparison of evidence: the query can receive a matching law from another form or a different law from its own form. Section~\ref{sec:crossing} uses this crossing to measure the two influences separately.

\subsection{Problem Formulation}\label{sec:formulation}
We can now state what dynamics understanding means for a predictor. Let an instance $\omega=(m,\lambda)$ be a mechanism $m$ realizing law $\lambda$. An intervention $\iota\in\mathcal I$ specifies an initial state, input program, and control channel, and evidence $e$ is the interaction history provided to the model. For a predictor queried at given states, its accelerations at matched states define an implied field $\hat a_{\omega,e}(\cdot\,;\iota)$ in the public coordinates of $m$, which are declared and available to every model. We use the conditional mean for stochastic predictors. A declared law class $\Theta$ contains the admissible explanations, including the true laws, and $(\psi,K)_\ast$ transports acceleration fields through a certified correspondence. With these objects, dynamics understanding decomposes into reuse, adjustment, and consistency, together with a physical anchor.

\begin{definition}[Capabilities and anchor]\label{def:capabilities}
\leavevmode
\begin{itemize}
\item[(C1)] \textbf{\textup{Reuse.}} For every certified pair $(A,B)$ under $(\psi,K)$ realizing the same law, $\hat a_{(B,\lambda),e}=(\psi,K)_\ast\,\hat a_{(A,\lambda),e}$ for all $\iota$, with corresponding evidence.
\item[(C2)] \textbf{\textup{Adjustment.}} For laws $\lambda_1,\lambda_2$ realized on the same mechanism with declared evidence budgets, $\hat a_{(m,\lambda_2),e_2}-\hat a_{(m,\lambda_1),e_1}=a^\ast_{\lambda_2}-a^\ast_{\lambda_1}$ for all $\iota$.
\item[(C3)] \textbf{\textup{Consistency.}} For each $\omega$ there is $\theta\in\Theta$ with $\hat a_{\omega,e}(\cdot\,;\iota)=a^\ast_\theta$ for all $\iota$, up to the declared equivalences of $\Theta$.
\item[(A)] \textbf{\textup{Anchor.}} At a reference instance $\omega_0=(m_0,\lambda_0)$, $\hat a_{\omega_0,e}(\cdot\,;\iota)=a^\ast_{\lambda_0}$ for $\iota\in\mathcal I_A$, where $\mathcal I_A$ is \emph{identifying} for $\Theta$: any $\theta,\theta'\in\Theta$ whose fields agree on $\mathcal I_A$ are equivalent.
\end{itemize}
\end{definition}

\begin{theorem}[Characterization]\label{thm:characterization}
Assume that (a) the tested instances $\Omega$ are connected to $\omega_0$ by certified twin pairs and by law changes within mechanisms, (b) all true laws lie in $\Theta$, and (c) $\mathcal I_A$ is identifying for $\Theta$. Then the predictor is correct on $\Omega\times\mathcal I$, that is $\hat a_{\omega,e}(\cdot\,;\iota)=a^\ast_\lambda$ for all $\omega=(m,\lambda)\in\Omega$ and $\iota\in\mathcal I$, if and only if (C1), (C2), (C3), and (A) hold.
\end{theorem}

Intuitively, the anchor fixes the physical law at one reference instance. Reuse and adjustment carry that agreement through the connected set, and consistency extends it across interventions. Each condition matters. A predictor using $c\lambda$, for instance, reproduces free motion, transfers across law twins, and remains mechanically consistent, yet it is wrong. Known effort in SI units resolves this scale ambiguity and anchors the predictions to the physical law.

For finite errors, the quantitative counterpart bounds prediction error by the anchor error, the relation tolerances, and the length of the connecting path (Supplementary Theorem~\ref{thm:quantitative}). For finite evidence, a sufficiency test requires the laws consistent with the supplied evidence to agree on the query; their disagreement gives an irreducible prediction error (Supplementary Proposition~\ref{prop:decidable}). FoLaSim can therefore distinguish a model that lacks information from one that fails to use it.

\section{The Forward Half: From Law to Form}\label{sec:constructions}

\begin{figure*}[!t]
\centering
\includegraphics[width=\textwidth]{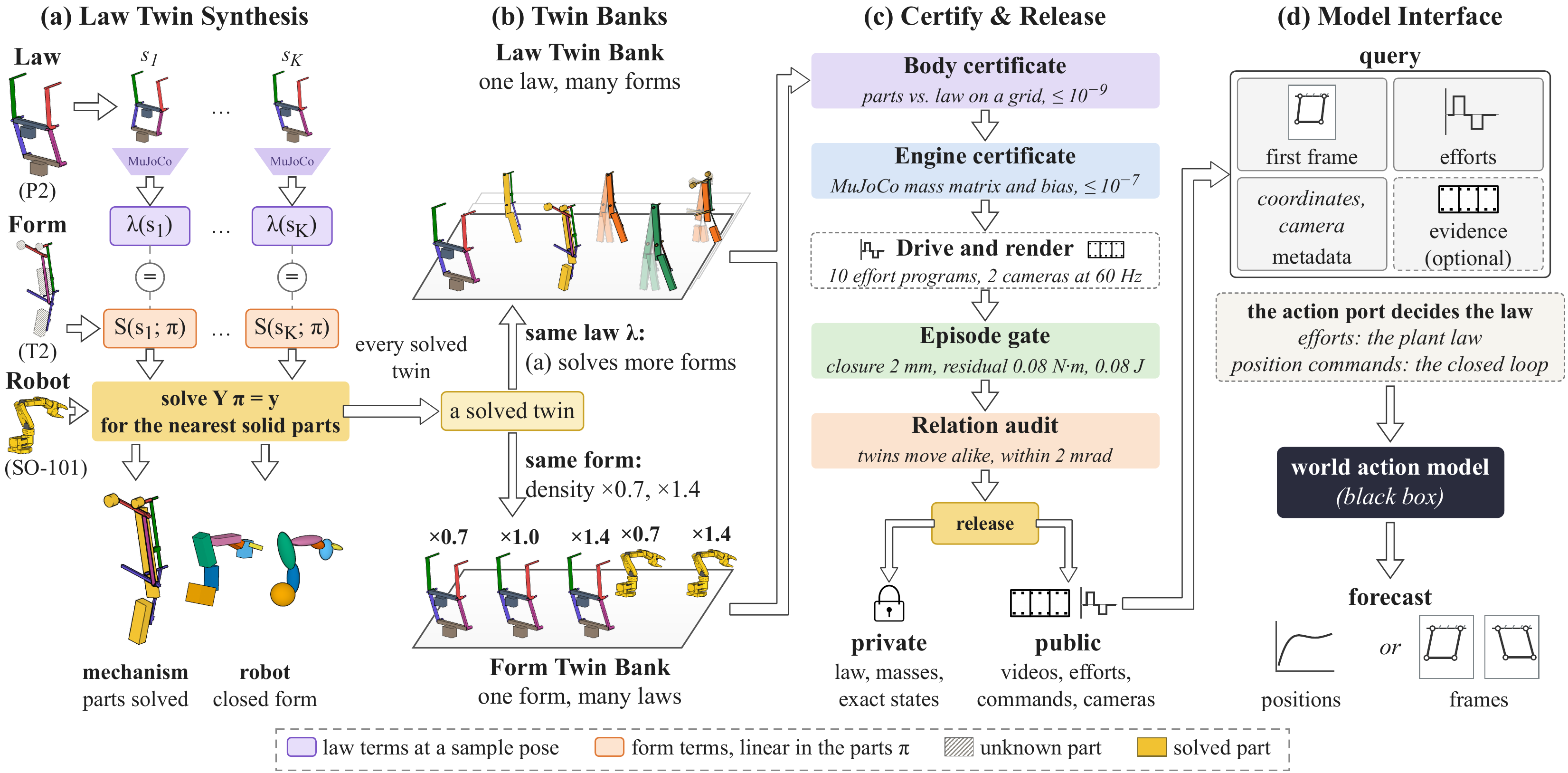}
\caption{From a target law to certified queries. \textbf{(a)} The engine samples P2's law, and $Y\pi=y$ determines realizable parts nearest a proposed design. Hatched T2 parts are tunable; robot twins have a closed-form equimomental construction. \textbf{(b)} The resulting banks vary form at fixed law or density at fixed form. \textbf{(c)} Body and engine certificates precede simulation; episode and relation checks verify the executed motion. The release separates model observations from evaluation truth. \textbf{(d)} A model receives the query, actions, and optional evidence, then returns frames or positions. Effort inputs test the plant law; position commands test the closed loop. The forecast enters the inverse half in Fig.~\ref{fig:overview}.}
\label{fig:forward}
\end{figure*}

The forward half turns a target law into certified, visible motion (Fig.~\ref{fig:forward}). FoLaSim builds and excites rigid assemblies on the unmodified MuJoCo engine~\cite{todorov2012mujoco}, following the integration approach of robot-learning frameworks~\cite{dmcontrol2020,robosuite2020,mujocoplayground2025}. MuJoCo's constrained dynamics follow Gauss's principle~\cite{todorov2014convex}, which also underlies our residual, so reading the reduced dynamics from the engine ties the law used for synthesis to the bodies actually simulated. The same files also run on MJX for batched, differentiable generation, with implementation details in Supplementary Section~\ref{sup:simulator}.

\subsection{Law Twin Synthesis}\label{sec:twins}
Law twin synthesis answers a design question: which realizable parts give a chosen topology the target dynamics? A specification declares the assembly, constraints, actuators, public coordinates, and tunable parts. Fixed-shape ballasts, flywheels, and sliding blocks contribute parameters that enter the reduced dynamics linearly. Dynamic identification estimates such parameters from observations~\cite{gautier1990minimum,swevers1997optimal}; synthesis reverses this direction, prescribing the law and solving for the parts. For parameters $\pi$, matching at sampled states $s_k$ gives

\begin{equation}
M_S(s_k;\pi)=M_\lambda(s_k),\qquad \nabla V_S(s_k;\pi)=\nabla V_\lambda(s_k),
\label{eq:matching}
\end{equation}
These equations form a linear system $Y\pi=y$ whose coefficients come from the engine. Generalized coordinate partitioning~\cite{wehage1982partitioning} solves the constraints for the dependent joints, so FoLaSim can compute $M_S=N\Tr MN$ and $\nabla V_S=N\Tr g$, where $g$ is the gravity component of the MuJoCo bias force. The solver selects the solution nearest the nominal, as-drawn design and checks that every part is a realizable solid using the planar inertial-consistency conditions~\cite{wensing2018lmi}. The rank of $Y$ separates the parameter directions fixed by the target law from those left free for design variation. If no realizable solution exists, FoLaSim reports that the topology cannot realize that law.

\label{sec:built}
The built-in mechanisms exercise three aspects of dynamics of increasing difficulty. Six constant-inertia topologies realize one scalar law through different transmissions and mass arrangements. The coupled class matches a two-stage parallel mechanism P2 to a serial arm R2. Adding a gear constraint gives the variable-inertia pair P1/R1, whose configuration-dependent inertia links the effort response to the velocity forces in free motion. The evaluation thus progresses from scalar response to coupled and configuration-dependent dynamics. Supplementary Sections~\ref{sup:families} and~\ref{sup:coupled} give their formulas and part inventories.

The specification interface is also open to new topologies. T2, for example, drives an arm's elbow through a base-mounted crank and rod. Its additional moving parts introduce terms absent from the target law, and the synthesizer cancels them with two counterweights, producing a third law twin of P2 and R2 (Fig.~\ref{fig:forward}a; Supplementary Section~\ref{sup:t2}).

Automatic generation applies the same solve to sampled geometry (Fig.~\ref{fig:forward}b). Parameters constrained by the law stay fixed, while the free geometric and inertial directions create new forms. Each candidate enters the corpus only after synthesis and certification. The result is an unbounded set of certified twins that supports evaluation on previously unseen forms, beyond the built-in topologies (Supplementary Section~\ref{sup:forms}).

\subsection{Form Twin Synthesis}\label{sec:formtwins-forward}
Form twins are simpler to build. A uniform density change preserves every visible part, and damping and friction scale with density, as Definition~\ref{def:formtwins} requires. Additional controls vary a single body mass, load, friction, or damping. Applying these changes across the law-twin bank yields the crossed comparison of Section~\ref{sec:crossing}.

\subsection{Excitation, Control, and Rendering}\label{sec:programs}
Excitation decides which channel reveals the law. Releases test passive motion, pulses and oscillatory torques expose the response to effort, and compliant and stiff tracking shift information toward the controller's effort (Section~\ref{sec:invariances}). Every rendered frame therefore carries synchronized shaft efforts and, under tracking, commanded positions. Supplementary Section~\ref{sup:programs-gates} gives program definitions, sampling rates, and rendering settings.

\subsection{Robot Twins}\label{sec:robots}
\begin{figure*}[!t]
\centering
\includegraphics[width=\textwidth]{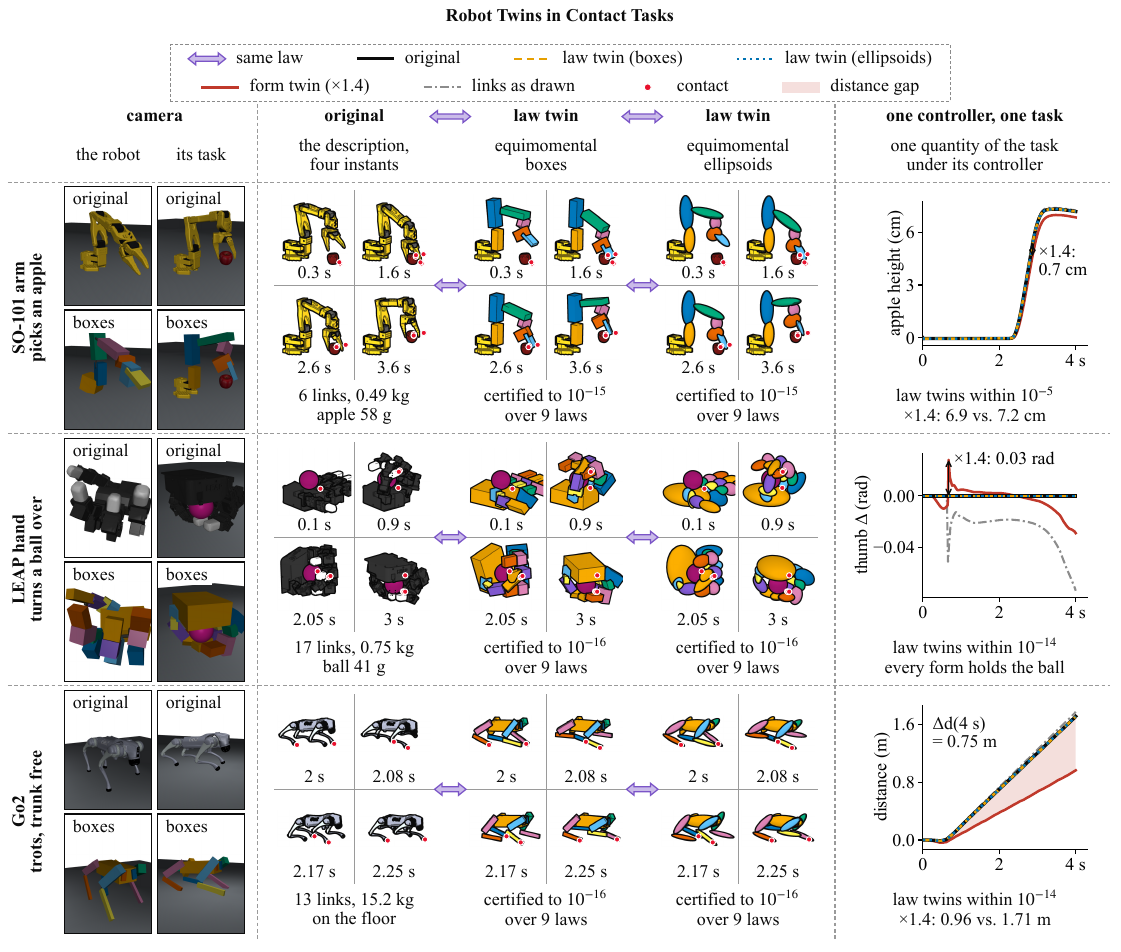}
\caption{Certified robot twins in contact (E1). Rows show the SO-101 apple pickup, LEAP ball turn, and Go2 trot with a free trunk. From left, camera frames compare the original with its equimomental boxes; ghosted assemblies show the original, boxes, and ellipsoids through the task, with contacts and certificate errors; task traces compare both law twins with a density-1.4 form twin and mesh-bounding-box links. The red shading marks the Go2 form twin's distance deficit: 0.96 versus 1.71\,m in the displayed run. The coincident law-twin traces show preserved dynamics despite changed form. Fixed-base comparisons appear in Supplementary Fig.~\ref{fig:robots-fixed}.}
\label{fig:robots}
\end{figure*}
Beyond designed mechanisms, FoLaSim reads robot dynamics directly from MJCF or URDF descriptions: link inertials, reflected motor inertia, joint effort ports, damping, and smooth dry friction. The included SO-101 arm~\cite{soarm2025}, LEAP hand~\cite{shaw2023leap}, and Unitree Go2 cover manipulation, dexterity, and locomotion; the latter two use MuJoCo Menagerie descriptions~\cite{menagerie2022}. Other descriptions enter through the same interface without per-robot code.

For robots, the matching problem \eqref{eq:matching} has an especially efficient, closed-form solution. A link contributes to the law only through ten inertial parameters, namely its mass, center of mass, and inertia tensor. Any solid with the same parameters therefore preserves that contribution, as in classical equimomental systems~\cite{routh1905elementary}. For link $\ell$ with mass $m_\ell$ and principal moments $I_1,I_2,I_3$ about its center of mass, we construct a uniform box or ellipsoid centered there and aligned with the principal axes. Their half edges $a_i$ and semi-axes $b_i$ are
\begin{equation}
a_i=\sqrt{\frac{3(I_j+I_k-I_i)}{2m_\ell}},\qquad b_i=\sqrt{\frac{5(I_j+I_k-I_i)}{2m_\ell}},
\label{eq:equimomental}
\end{equation}
for $\{i,j,k\}=\{1,2,3\}$. Both solids exist for every physically consistent, nonplanar link~\cite{wensing2018lmi}. Replacing each link with its box or ellipsoid thus keeps its inertial contribution while changing the visible form. Uniform scaling of link densities, rotor armatures, and dissipation produces form twins, and independent link-density changes generate the robot's law family. This construction supplies exactly the certified pairs and law changes that Theorem~\ref{thm:characterization} requires, within a rigid-body class that contains the true law and is linear in its inertial parameters (Section~\ref{sec:reports}).

Contact scenes keep the robot's collision primitives and replace only its visible inertial form. The tasks in Fig.~\ref{fig:robots} combine the robots with YCB objects~\cite{calli2015ycb}. Law twins preserve the complete contact dynamics, while density form twins scale every moving body, including the manipulated objects. MuJoCo's mass-normalized soft contact then preserves the density relation: free motion agrees, and actuated motion agrees when efforts scale with density. In each task, one nominal controller drives all twins and laws. Supplementary Sections~\ref{sup:robots-open}, \ref{sup:contact}, and~\ref{sup:zoo} describe fixed-base programs, contact settings, and the general robot-import procedure.

\subsection{Certification and Data Splits}\label{sec:certify}
Certification links the intended law to the released observations through four checks, from the bodies to the motion. A \emph{body certificate} checks the assembly's reduced inertia, velocity coupling, and gravity, and an independent \emph{engine certificate} checks the projected engine dynamics. An \emph{episode gate} then verifies closure, force balance, and energy balance along the executed motion. Finally, a \emph{relation audit} checks twin trajectories under matched interventions. Together, these checks bound the simulation mismatch in a model comparison; Supplementary Section~\ref{sup:programs-gates} gives the thresholds.

Public assets provide videos, recorded efforts and commands, coordinate and input maps, and camera calibration. The law, masses, and exact states serve as evaluation truth, and only canonical-state experiments explicitly expose states as model inputs. Development data set the readout and training choices, pilot laws establish the protocols, and separate confirmation laws evaluate the frozen models and settings. Supplementary Sections~\ref{sup:corpus} and~\ref{sup:protocol} list corpus inventories and split identifiers.

\subsection{Model Interface}\label{sec:interfaces}
A common black-box interface connects the simulator to predictors. Each query supplies an initial state or image prefix, actions, public metadata, and optional evidence, and the model returns positions or frames. Its observations, history interface, and action port determine which tests apply. Image and video outputs enter form space, whereas state outputs enter state space.

The action port also determines which dynamics are evaluated. Effort inputs test the plant law. For position commands, the reader computes the known controller's effort at the predicted states and tests the resulting closed loop. Because this distinction carries through every reported score, controller compensation has an explicit physical interpretation.

\section{The Inverse Half: From Form to Law}\label{sec:measurement}

\begin{figure*}[t]
\centering
\includegraphics[width=\textwidth]{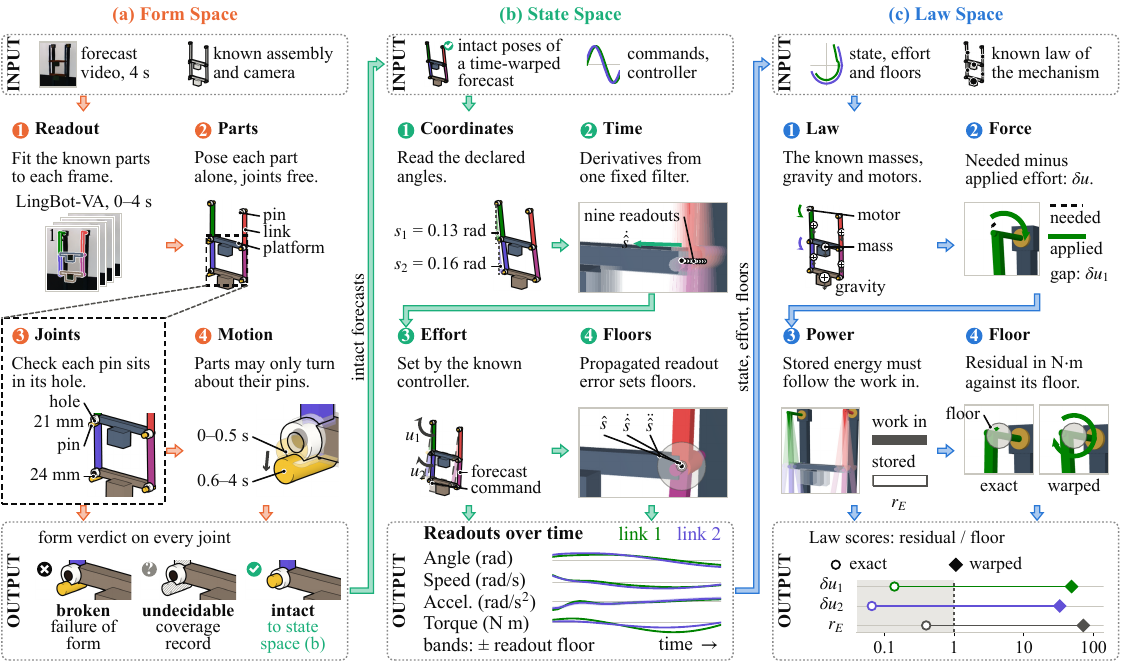}
\caption{Reading a P2 forecast from form to law (law 2009, density 1.4, servo commands). \textbf{(a)} Independently fitted bodies expose a displaced lower link in a LingBot-VA forecast; joint verdicts combine geometric gaps with pixel evidence. \textbf{(b)} A separate, designed time-warped forecast $s(0.8t)$ preserves the assembly. Its angles, derivatives, and controller effort share a time axis; rings and shaded bands show propagated readout floors. Dots trace the upper platform and the arrow gives its velocity. \textbf{(c)} Required effort (dashed) is compared with applied effort (solid), and energy change with actuator work. The warped forecast leaves 5.9 and 2.9\,\Nm{} unexplained at the two ports, while the exact forecast stays within its floors. Supplementary Section~\ref{sup:single} gives the complete numerical example.}
\label{fig:overview}
\end{figure*}

\begin{figure*}[t]
\centering
\includegraphics[width=\textwidth]{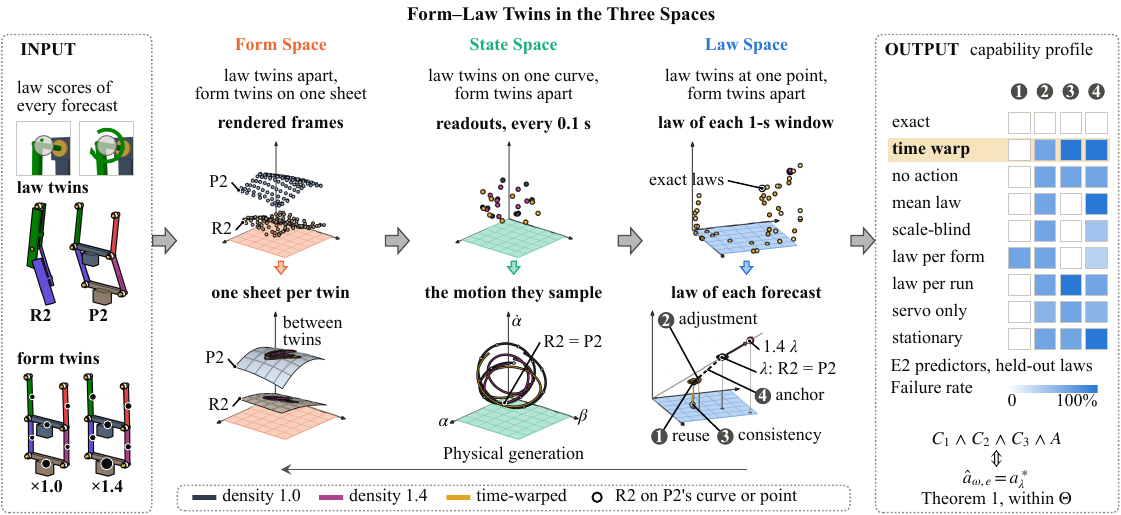}
\caption{From individual diagnoses to tests across twins. On the left, certified correspondences connect the law twins P2/R2 and the density form twins of P2. The center shows their relation in form, state, and law spaces; the bottom arrow follows the physical chain from law to state to form, which the reader traverses in reverse. The four tests operate on the resulting law-space comparisons. On the right, the nine designed predictors of E2 have distinct capability profiles: the time-warped predictor passes reuse but fails adjustment, consistency, and anchor; only the exact predictor passes all four.}
\label{fig:twins}
\end{figure*}

The inverse half travels the chain of Section~\ref{sec:manifold} backward. It reads a single forecast through form, state, and law spaces (Fig.~\ref{fig:overview}), then compares forecasts across certified twins (Fig.~\ref{fig:twins}). The known law and query effort provide the physical reference, so no recorded future is needed. Measurement floors and readout coverage accompany every diagnosis and tie each conclusion to the precision of the observations. Supplementary Section~\ref{sup:measurement} details the procedure.

\subsection{Form Space: Pose Reading and Joint Verdicts}\label{sec:read}
The reader first recovers physical poses from pixels, using the known assembly geometry. It renders each body from the calibrated camera and fits its planar pose over the corresponding image region. A joint estimate of color gain and bias separates appearance changes from motion. Bodies are fitted independently of the joint constraints, so a displaced link remains visible to the test. Unreadable frames are marked unmeasurable and reduce coverage.

Each joint then receives an intact, broken, or undecidable verdict from two kinds of evidence. The geometric quantity is the gap that its equality constraint leaves between the fitted bodies. The photometric quantity is the increase in pixel cost when the joint is enforced. We require both to support a break, because a pose-fitting error can produce a spurious gap without a corresponding pixel penalty.

A persistent excess in both quantities marks a joint as \emph{broken}. Reliable readings within the calibrated gap threshold establish an \emph{intact} joint, and insufficient or conflicting evidence gives an \emph{undecidable} verdict. Broken forecasts identify form failures, while intact forecasts proceed to the law-space evaluation. Supplementary Section~\ref{sup:readout} specifies the temporal and coverage rules.

\subsection{State Space: Coordinates, Derivatives, and Effort}\label{sec:state}
State space bridges the visual readout and the physical evaluation. Forecasts are expressed in public coordinates $\hat s_t$, with law twins related by $\psi$. A fixed linear differentiator estimates velocity and acceleration, and residuals are evaluated at interval midpoints. The query effort is then attached to this motion, through the known controller for position commands. Since differentiation and dynamics amplify readout errors, we propagate the errors measured on development videos through the differentiator, mass matrix, constraints, and controller. Each derived quantity therefore enters law space with its own measurement floor.

\subsection{Law Space: Unexplained Effort}\label{sec:core}
\textbf{Unexplained effort gives dynamic error a physical scale.} Let $\Lambda=B_S\Tr M_S^{-1}B_S$ be the response to effort, i.e., the Gram matrix of actuator fields in the kinetic metric. For an invertible port, $\delta u=B_S^{-1}\rho$ is the effort left unexplained by the forecast, and
\begin{equation}
r^2=\delta u\Tr\Lambda\,\delta u,
\label{eq:gram}
\end{equation}
expresses its norm in the response metric. Reported per actuator in \Nm{}, $\delta u$ turns a physical failure into an interpretable quantity, namely missing torque.

\textbf{Energy balance checks the work implied by a forecast.} With storage $H=\tfrac12\dot s\Tr M_S\dot s+V_S$, the power balance~\cite{vanderschaft2014port} yields
\begin{equation}
r_E(t)=H(t)-H(t_0)-\int_{t_0}^t\left(u\Tr B_S\Tr\dot s-\dot s\Tr d_S\right)d\tau,
\label{eq:energy}
\end{equation}
which compares stored energy with actuator work and dissipation. Force and power residuals catch different failures. A stationary forecast, for example, can satisfy power balance while violating force balance.

\textbf{Momentum balance accommodates unknown contact forces.} We evaluate it over the differentiator window, using $p=M_S\dot s$ and the zero-acceleration bias and dissipation forces $F_0$. For a weight $w\ge0$ vanishing at both window endpoints, integrating $\dot p=u-F_0+\dot M_S\dot s+J\Tr\lambda$ gives
\begin{equation}
-\!\int\! \dot w\,p\,dt-\!\int\! w\,(u-F_0+\dot M_S\dot s)\,dt=\!\int\! w\,J\Tr\lambda\,dt,
\label{eq:momentum}
\end{equation}
This generalized-momentum balance, also used for robot collision detection~\cite{deluca2005sensorless,haddadin2017robot}, needs only velocities, represents impacts as impulses, and remains linear in inertial parameters. We fit admissible contact impulses within the friction cones and score the unexplained remainder in the kinetic metric. The contact directions follow their configurations through the window, so the balance also covers changing support (Supplementary Section~\ref{sup:contact}).

The remaining residual reveals motion that no admissible contact can explain, such as an unsupported object.

\subsection{Four Tests over the Twins}\label{sec:relational}
A single diagnosis judges one forecast. The four tests instead use controlled sets of forecasts to separate complementary properties of the model. Reuse compares law twins at matched states and ports after applying the certified correspondence $(\psi,K)$. Each forecast also keeps its own physical residual, since two forecasts can agree by sharing the same wrong law. Adjustment asks whether a change of law within one form produces the correct change in acceleration. At matched states and inputs, with protocol-fixed kinetic metric $M_{\rm ref}$, its error is
\begin{equation}
E_{\rm ch}=\left\|(\hat a_{\lambda_2}-\hat a_{\lambda_1})-(a^\ast_{\lambda_2}-a^\ast_{\lambda_1})\right\|_{M_{\rm ref}};
\label{eq:change}
\end{equation}
A density change leaves free motion unchanged, so adjustment requires actuated evidence. For consistency, we fit one law $\hat\theta\in\Theta$ to support interventions and evaluate it on held-out interventions without refitting. A complementary local test derives a mass matrix from the model's effort response and uses it to predict free motion (Supplementary Section~\ref{sec:response-geometry-sup}). Finally, the anchor evaluates force and power residuals on identifying actuated interventions at a reference instance.

\label{sec:crossing}
Crossing twins then isolates what the evidence contributes. The four instances are $(A,\lambda_1)$, $(B,\lambda_1)$, $(A,\lambda_2)$, and $(B,\lambda_2)$, with a certified correspondence under both laws. Evidence comes from one cell, while unseen query interventions target $(B,\lambda_2)$, a combination excluded from training and checkpoint selection. With the query held fixed, same-law evidence from the twin tests transfer across forms, and other-law evidence from the target tests the influence of matching appearance. Comparing the two reveals whether a model benefits more from a matching law or from a matching form.

\subsection{Reports of Form and Law}\label{sec:reports}
The report closes the inverse half by turning residuals into an explanation of the dynamics. Since \eqref{eq:residual} is linear in the law parameters, the reader can fit the constants implied by a forecast. Comparing them with constants recovered from true motion on the same clock controls differentiation bias. The report then distinguishes the true law, a uniform density change, and changes in individual constants, and separately tests re-timing and motion amplitude. Each decision uses its own measurement floor.

For robots, the declared class contains the ten inertial parameters of every link, rotor armatures, and each joint's viscous and dry dissipation. The momentum balance \eqref{eq:momentum} is linear in these quantities, with columns computed by MuJoCo inverse dynamics. A weak prior centered on the description yields a maximum a posteriori estimate under convex physical-consistency constraints~\cite{wensing2018lmi}. Only the description's kinematics are trusted, so motion and effort can correct its inertial table while contact impulses remain unknowns.

The estimate also states how much the observations determine, namely the number of identified parameter combinations. Different inertial realizations can preserve the same reduced law, and contact can absorb further directions. The observable law is therefore the natural output of identification. Passing it to the forward half produces certified equimomental solids, which completes the path from measured motion to a robot twin.

\section{Experiments}\label{sec:experiments}

Our experiments follow the structure of the simulator. E1 and E2 first verify the two halves, namely the fidelity of twin synthesis and the reliability of physical diagnosis. On these validated comparisons, E3 and E4 study how models exploit evidence and how that behavior affects decisions and training. E5 and E6 then carry the same principle to an arm, from accurate and efficient simulation-trained prediction to frozen deployment and action selection on physical hardware. Supplementary Section~\ref{sup:experiments} provides detailed settings and complete tables.

\begin{figure*}[t]
\centering
\includegraphics[width=\textwidth]{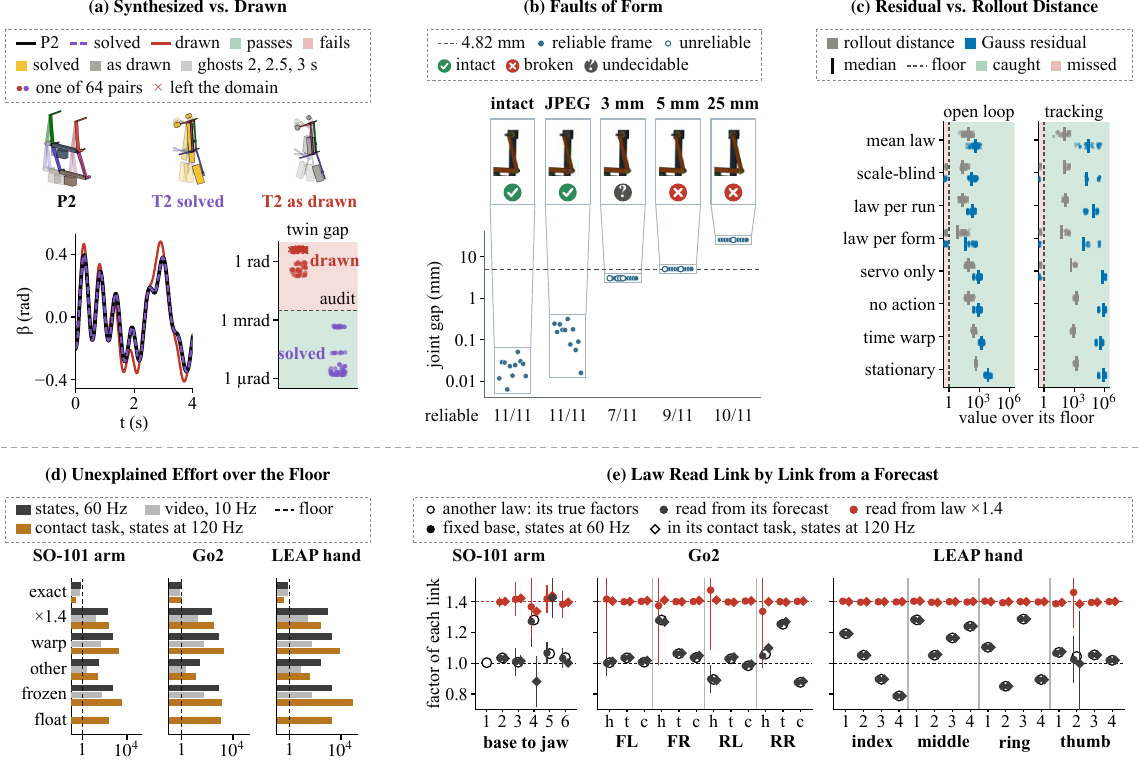}
\caption{Self-check of FoLaSim on mechanisms and robots. \textbf{(a)} P2--T2 upper-link gaps under the program of Fig.~\ref{fig:forward}: 0.4\,mrad after synthesis and 0.20\,rad as drawn. All 64 E1 pair gaps are compared with the 3\,mrad audit. \textbf{(b)} Five A1 law-25 clips paired with ground--arm1 gaps. Dots are frames, open dots indicate unreliable readings, and counts give reliable frames. The 4.82\,mm threshold uses intact law-24 clips under all degradations; verdicts also require pixel evidence and temporal coverage. \textbf{(c)} Rollout distance and Gauss residual over per-probe floors for eight failing predictors. At exact states, the residual's detection margin is 5--15 times larger under open-loop effort and 130--1,300 times larger under tracking. \textbf{(d)} Geometric-mean unexplained effort over its floor, for five forecasts across three held-out programs (60\,Hz states, 10\,Hz video) and six per contact task (120\,Hz states, unknown forces). The floating fault raises the apple, ball, or trunk by 2\,cm and exceeds the floor by 220--2,100 times despite exact joint angles. \textbf{(e)} Link factors for another family law and the density-1.4 form twin. Open circles give truth; dots and diamonds give fixed-base and contact estimates, with bars of two standard errors and two standard deviations over readout noise, respectively. Supplementary Tables~\ref{tab:formfaults}, \ref{tab:contact-reading}, and~\ref{tab:contact-law} give the per-family and per-task values.}
\label{fig:selfcheck}
\end{figure*}

\label{sec:protocol}
Evaluation uses held-out laws and excludes changed-density target combinations from training and checkpoint selection. We compare two kinds of learned predictors. Canonical-coordinate models comprise a history-free MLP, a context GRU, and a mechanical-context model, with DeLaN~\cite{lutter2019delan}, a dissipative extension, and a canonical MLP as published-architecture controls. RGB context models and their no-context controls receive images and efforts. The first kind uses evidence through a known coordinate correspondence, and the second through visual motion.

We also evaluate four native video models zero-shot through their released interfaces: the joint video--action policy LingBot-VA~\cite{li2026causal}, the action-conditioned BWM~\cite{bwm2026} and Ctrl-World~\cite{ctrlworld2025}, and the omnimodal Cosmos3-Nano~\cite{cosmos3_2026}. We measure their adjustment by the response gain $g=\langle\Delta\hat s,\Delta s\rangle/\|\Delta s\|^2$, which projects the predicted density-change response onto the true response. Exact adjustment gives $g=1$, and no response gives $g=0$.

Readout settings are fixed on development laws. We estimate uncertainty by bootstrapping laws after averaging probes, densities, and seeds within each law, so repeated frames do not inflate the sample size. Supplementary Section~\ref{sup:protocol} gives model counts, splits, and statistical settings, and Supplementary Section~\ref{sup:robots-open} gives the robot protocols.

\begin{table}[t]
\centering
\caption{Robot results across E1--E3. E1 reports the largest certificate error of the boxes and ellipsoids over nine laws of each family and the largest joint-angle gap to the original. E2 gives unexplained effort over its floor for the exact forecast and four designed faults. E3 reports the ratio of the original's other-law error to the boxes' same-law error across three seeds. For contact tasks, E1 measures joint-angle and body-position gaps under the task controller. E2 uses states at 120\,Hz with unknown contact forces and reports which link factors of another family law are recovered within 0.05.}
\label{tab:robots}
\footnotesize
\setlength{\tabcolsep}{2.6pt}
\begin{tabular}{@{}lccc@{}}
\toprule
 & SO-101 & Go2 & LEAP \\
\midrule
E1: certificate, nine laws & $4\times10^{-16}$ & $8\times10^{-16}$ & $6\times10^{-16}$ \\
E1: law twins apart (rad) & {\boldmath$1\times10^{-15}$} & {\boldmath$9\times10^{-16}$} & {\boldmath$4\times10^{-14}$} \\
E1: density 1.4 apart (rad) & 0.72--0.86 & 0.40--0.47 & 1.71--1.79 \\
E1: links as drawn apart (rad) & 0.14 & 0.36 & diverges \\
\midrule
E2: exact forecast, at most & 0.77 & 0.96 & 0.73 \\
E2: faults at 60 / 10\,Hz, at least & \textbf{27 / 2.2} & \textbf{25 / 2.1} & \textbf{275 / 7.4} \\
\midrule
E3: other law $\div$ same-law twin & \textbf{9.4--11.2} & \textbf{3.9--5.5} & \textbf{5.1--5.6} \\
\midrule
Contact task & apple & trot & ball \\
E1: law twins apart (rad, m) & $4\times10^{-6}$ & $7\times10^{-15}$ & $5\times10^{-15}$ \\
E1: density 1.4 apart (rad) & 0.097--0.102 & 0.17--0.19 & 0.032--0.035 \\
E2: exact forecast, at most & 0.18 & 0.96 & 0.28 \\
E2: another law of the family & 24 & 13 & 39 \\
E2: other faults, at least & \textbf{220} & \textbf{309} & \textbf{284} \\
E2: its links read within 0.05 & \textbf{3 of 6} & \textbf{12 of 12} & \textbf{15 of 16} \\
\bottomrule
\end{tabular}
\end{table}

\subsection{E1: Fidelity and Scalability of Twin Synthesis}\label{sec:e1}

\textbf{Synthesis turns nominal geometry into faithful law twins.} Regenerating the built-in families reproduces the prescribed laws and passes both certificates. Comparing designs before and after synthesis shows why matching is necessary (Fig.~\ref{fig:selfcheck}a). Nominal designs depart substantially from the reference and often leave the admitted domain, whereas every synthesized pair stays within the relation tolerance. The certificates thus predict agreement in executed motion, not merely a fit at sampled states.

The same process also handles a new topology and detects a physically incompatible one. T2 realizes P2's law from its specification. Changing its upper-link length, however, violates the coupling between inertial and gravitational terms and makes the target law unrealizable. Form twins provide the complementary check. Their passive motions coincide, yet identical applied efforts separate them, directly exposing the density ambiguity that passive observation leaves unresolved (Supplementary Table~\ref{tab:regeneration}).

\textbf{Automatic generation and batching make certified variation scalable.} Sampled forms span new geometry while retaining the source law under shared efforts (Supplementary Fig.~\ref{fig:forms-all}). In this study, synthesis and certification take less than a second per accepted form. MJX preserves the C engine's trajectories within the relation tolerance and generates over 400 four-second A1 episodes per second on one GPU. Differentiating the rollout residual with respect to density further supports optimization through the generated dynamics. Supplementary Section~\ref{sup:e1-forward} reports acceptance rates, certificate errors, throughput, and gradient checks.

\textbf{Closed-form twins preserve robot dynamics through contact.} On all nineteen robots, equimomental forms reproduce the original dynamics to machine precision (Table~\ref{tab:robots}; Supplementary Table~\ref{tab:zoo}). Simpler substitutes fall short. Mesh-bounding-box links with the correct masses still depart substantially on the included robots, which shows that the full inertia tensor and center of mass matter. The closed-form construction preserves both automatically.

In contact, the law twins pick up the apple, turn the ball, and trot with their originals despite their different forms (Fig.~\ref{fig:robots}). Density form twins, in contrast, change the task outcome under the same controller. Scaling their efforts restores agreement, as the density relation predicts. This controlled recovery attributes the observed gap to the changed law rather than to the rendering or controller. The sixteen additional Menagerie descriptions extend the result across arms, hands, quadrupeds, and humanoids, each loaded and exercised in contact without per-robot code.

\subsection{E2: Physical Diagnosis and Law Identification}\label{sec:e2}\label{sec:metaval}\label{sec:reliability}

\textbf{Unexplained effort exposes faults that trajectory tracking can conceal.} We evaluate an exact predictor and eight designed failing predictors, each of which realizes an observability blind spot or violates a condition of Definition~\ref{def:capabilities}. Across the A1/B1 fault suite, every fault fails its intended test, and the exact predictor passes all tests (Fig.~\ref{fig:twins}). The joint profiles explain each failure. A mean-law predictor transfers across twins and remains mechanically consistent, but its force and power residuals reveal a coherent, incorrect law. Re-timing preserves geometry while violating both balances. A scale-blind predictor is exact in free motion but fails under actuation, and a stationary predictor passes power balance but fails force balance.

How does this compare with trajectory error? On exact state forecasts, unexplained effort gives a substantially larger detection margin than trajectory distance, especially under tracking (Fig.~\ref{fig:selfcheck}c). The controller corrects the motion produced by a wrong law and thereby conceals its trajectory error. The mean-law predictor, for example, stays within 0.01\,rad of the reference while leaving 0.26\,\Nm{} unexplained. At the video clock, differentiation amplifies readout noise and raises the residual floor, so trajectory distance detects moderate open-loop errors more often; under tracking, the two scores agree.

\textbf{Degradation-aware calibration separates visible breaks from readout noise.} We test intact and displaced-link clips under blur, compression, color shifts, reduced resolution, and pixel noise. Calibration on clean references detects small displacements but mistakes some degradations for breaks. Including degraded references removes false alarms on all intact test clips, while the detectable displacement depends on the assembly (Fig.~\ref{fig:selfcheck}b). The multi-body parallel mechanisms require larger displacements than the simpler linkages, and R2 remains direction-sensitive. Supplementary Table~\ref{tab:formfaults} gives the complete detection thresholds and directional results.

\textbf{The implied law identifies the physical source of a forecast's error.} On R2 and B1, state and video readouts distinguish density scaling, re-timing, changes in individual constants, and a frozen forecast that admits no explaining law. A displaced body is localized to its broken joint (Supplementary Table~\ref{tab:reports-validation}). The parameter estimate also recovers R2's density change when its scalar residual barely exceeds the video floor. A nearby law whose constants differ by at most 6\% is instead reported as a near-uniform scaling, which marks the resolution of this readout.

On the three robots, the same diagnosis detects every designed fault, while exact forecasts remain within their floors (Fig.~\ref{fig:selfcheck}d,e; Table~\ref{tab:robots}). At the state clock, the report recovers density and timing scales and the alternate law's link factors. At the video clock, the anchor verdicts remain stable, but identification becomes coarser: density factors no longer pass the uniformity test, and an arm law differing mainly at the wrist is identified as the true law. The comparison thus separates reliable fault detection from the finer task of locating its parameters.

\textbf{Actuated motion reveals robot dynamics through unknown contact forces.} In the contact tasks, we attach the servo effort to each forecast and solve for admissible contact impulses. Exact forecasts stay within their floors, while scaled, re-timed, alternate-law, and frozen forecasts all fail (Fig.~\ref{fig:selfcheck}d). The floating fault makes the distinction especially clear. Its joint angles are exact, yet no friction-cone impulse can support the displaced apple, ball, or trunk. The residual therefore detects a failure that joint-angle error misses.

Identification follows the directions that the task excites (Fig.~\ref{fig:selfcheck}e). The trotting quadruped's leg factors are accurately recovered, and so are the hand's finger factors, except at the thumb base pressing on the ball. Supporting contacts, in contrast, allow the arm's wrist and gripper to trade mass. The quadruped's density twin is identified as a uniform scaling of bodies, rotors, and dissipation. These differences tie the observed parameter uncertainty to the mechanics of support; Supplementary Table~\ref{tab:contact-law} reports the estimates and readout spreads.

\textbf{Recovered dynamics turn inaccurate robot descriptions into certified twins.} We perturb inertials, rotor armatures, and dissipation in twenty robot scenes, then recover the law from one episode of motion and servo effort with unknown contact forces. Only the description's kinematics are trusted (Supplementary Section~\ref{sup:law}). On six open-chain arms, including one loaded from URDF, and the Shadow hand, held-out force error falls from 1.5--46\% to 0.01--0.08\%. The recovered law also improves held-out rollouts; Supplementary Table~\ref{tab:law} reports these errors alongside the closed-linkage and contact-task results.

Passing the recovered law to the forward half yields certified equimomental boxes. How much of the description can be recovered depends on the excitation. A standing legged robot reveals fewer leg-parameter combinations, because ground impulses absorb the unexcited directions, and the reported rank states how many combinations the motion determines.

\subsection{E3: Form Versus Law in Model Predictions}\label{sec:e3}
\begin{figure*}[t]
\centering
\includegraphics[width=\textwidth]{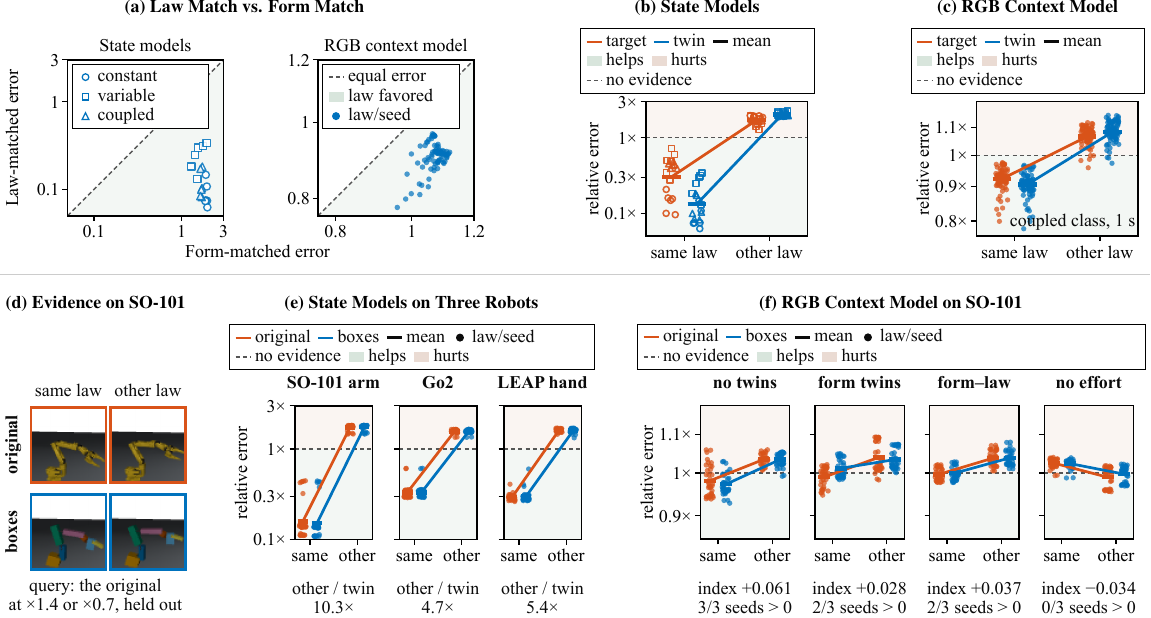}
\caption{Form-versus-law test (E3) on mechanisms and robots, with fixed query states and future efforts and disjoint evidence episodes. Errors are relative to no evidence. \textbf{(a)} Paired errors from (b), (c): original-form, other-law evidence on $x$ and twin-form, same-law evidence on $y$. Points below the diagonal favor law matching. \textbf{(b)} State models' predicted-change error by class, model, and seed. With twin evidence, this error is 7--28\% of the true change; original-form, other-law evidence gives 6--28 times larger errors. \textbf{(c)} RGB context model's 1\,s motion-region error on the coupled class, by law and seed. Bars mark geometric (b) and pooled (c) means; lines join means by form. \textbf{(d)} Held-out SO-101 evidence at the query's and the other density. \textbf{(e)} State models on three robots. Footnotes compare original-form, other-law error with boxes' same-law error. Individual-seed ratios range from 4--11 across robots. \textbf{(f)} SO-101 RGB model under four training corpora, with law-following indices from Table~\ref{tab:e4main}. Dots in (e), (f) are held-out laws and seeds. Close lines indicate transfer across forms. Detailed errors and RGB ablations are in Supplementary Tables~\ref{tab:state-transfer}, \ref{tab:robot-state}, and~\ref{tab:rgb-ablation}.}
\label{fig:formlaw}
\end{figure*}

With the simulator and readout validated, we return to the central question: what do models extract from evidence? Crossed twins answer it directly. A held-out density change rescales the target's kinetic metric, and actuated evidence reveals that change. The query state and future efforts remain fixed, while the evidence changes across the two-by-two design (Fig.~\ref{fig:formlaw}). This isolates the benefit of matching the law from the benefit of matching the form.

\textbf{State models transfer law information across certified changes in form.} Canonical-coordinate context models receive the mechanism identity and public correspondence. Across the three dynamics classes, same-law evidence from a twin reduces trajectory error from 7--34\,mrad without evidence to 2--5\,mrad (Fig.~\ref{fig:formlaw}b). Evidence from the target under another law, by contrast, behaves much like evidence mismatched in both form and law. The correspondence therefore lets the model use a shared physical response across different mechanisms, whereas matching the target's form alone provides little benefit (Supplementary Table~\ref{tab:state-transfer}).

This transfer persists on the SO-101 arm, Go2, and LEAP hand. Each context model reads two evidence episodes and predicts under the program's controller. When the law matches, evidence from the original robot and from its boxes gives similar improvements. Across every robot and seed, the boxes' same-law evidence outperforms the original's other-density evidence (Fig.~\ref{fig:formlaw}e; Table~\ref{tab:robots}). The useful match is thus physical, even when the evidence comes from another visible embodiment.

\label{sec:rgb-result}
\textbf{Under fixed pulses, RGB models read law information from motion.} Unlike state models, these models receive no states, masses, family labels, or coordinate maps, only image prefixes, evidence clips and actions, and future efforts. Even so, correct evidence outperforms other-law evidence in every class and seed at 1\,s (Fig.~\ref{fig:formlaw}c). On the variable-inertia and coupled classes, it also improves on nominal-density evidence from the same twin, which shows sensitivity to a change in kinetic-metric scale (Supplementary Table~\ref{tab:rgb-main}).

The evidence ablations locate this information in the observed motion. Freezing or shuffling frames causes a much larger loss than shuffling efforts (Supplementary Table~\ref{tab:rgb-ablation}). Because the open-loop pulses are fixed across densities, their effort records cannot distinguish the laws. The changed acceleration and displacement can, and the model uses that visual response. This finding motivates the comparison with tracking evidence in E4, where the informative channel changes.

\textbf{Mechanical consistency and a correct law are distinct outcomes.} A model can use evidence well without following one law. We fit one structure to each model's impedance and servo forecasts and test it on chirp and work forecasts. The variable-inertia GRU attains a trajectory error of 0.0031\,rad, yet its shared structure leaves 0.38\,\Nm{} unexplained on held-out interventions, compared with 0.0083\,\Nm{} on simulated data. Its accurate trajectories therefore combine responses that do not come from one law. The mechanical-context model admits a common structure to $6\times10^{-5}$\,\Nm{}, but that structure differs from the physical law.

Response-based identification reaches the same distinction on coupled systems. The GRU's effort response, free motion, and energy are incompatible with one structure, whereas those of the mechanical-context model are compatible (Supplementary Proposition~\ref{prop:response-structure}). DeLaN imposes a mechanical structure by design, yet it fails the anchor on two of three coupled seeds. Together, these tests separate three outcomes: learning from law-related evidence, producing mechanically consistent predictions, and recovering the correct physical law.

\begin{figure*}[t]
\centering
\includegraphics[width=\textwidth]{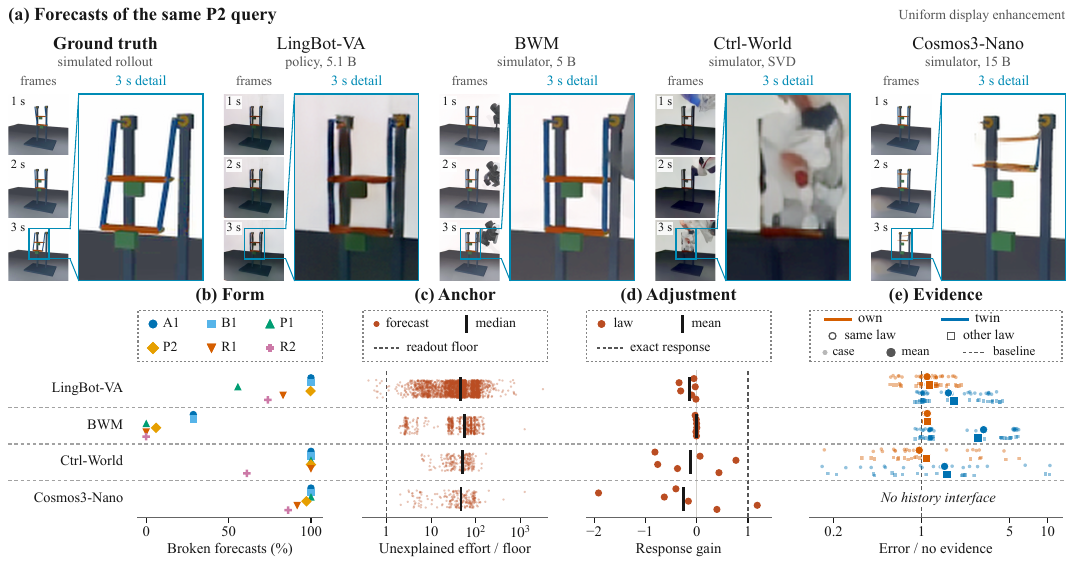}
\caption{Native video forecasts and their physical diagnoses. \textbf{(a)} The P2 query of Fig.~\ref{fig:overview} (law 2009, density 1.4, servo commands), shown at 1, 2, and 3\,s with cyan-boxed 3\,s enlargements. All displayed frames share the hue-preserving HSV transform $V'=V^{0.58}$, $S'=1-(1-S)^{1.18}$; measurements use the originals. \textbf{(b)} Broken-forecast share by family. \textbf{(c)} Unexplained effort over its floor, with one point per measurable forecast and bars at medians. \textbf{(d)} Density-response gain per law, with bars at means and a dashed reference at exact adjustment ($g=1$). \textbf{(e)} Evidence error relative to no evidence: small markers denote cases and large markers pooled ratios. Cosmos3-Nano has no history interface. Horizontal dashed lines separate models. Supplementary Table~\ref{tab:models} gives verdict counts, gains with confidence intervals, and evidence ratios. Model-specific horizons and readout clocks are detailed in Supplementary Sections~\ref{sup:e3-lingbot} and~\ref{sup:e3-models}; BWM uses 3.2\,s for the displayed sequence and 1.6\,s for the statistics.}
\label{fig:video}
\end{figure*}

\textbf{Abstract planar mechanisms provide controlled out-of-distribution diagnostic probes for native video models.} Evaluated zero-shot through their released action interfaces, all four models fail to adjust to a changed law (Fig.~\ref{fig:video}; Supplementary Table~\ref{tab:models}). LingBot-VA~\cite{li2026causal} receives commanded joint positions and accepts a separate episode through its interleaved video--action history, so we can evaluate its closed-loop law. Its forecasts move little. As a result, law twins agree with each other while both depart from the true motion. The anchor exposes this misleading reuse, and the model's near-zero density response shows that it does not adjust to the changed law. A structure fitted to two command branches also fails to explain the third.

The other video models show why form and law need separate tests. BWM preserves the form of every no-history forecast by barely moving the mechanism, while adding its own robot arms to the frame. Ctrl-World instead draws an arm across the mechanism and produces no intact forecasts. Cosmos3-Nano overshoots the motion and breaks most forecasts. Despite this markedly different visual behavior, all three receive the same physical diagnosis: large unexplained effort and a density response far from exact adjustment. Supplementary Section~\ref{sup:e3-models} gives the action adapters and full measurements.

Evidence reveals a further contrast with the context models trained on twins. For every video model that accepts history, evidence from the target's own form helps more than evidence from a physically matched twin (Fig.~\ref{fig:video}e). LingBot-VA's same-law twin history even performs worse than a still image of the twin across all six laws. For these models, the form of the evidence outweighs its physical relevance.

\begin{figure*}[t]
\centering
\includegraphics[width=.96\textwidth]{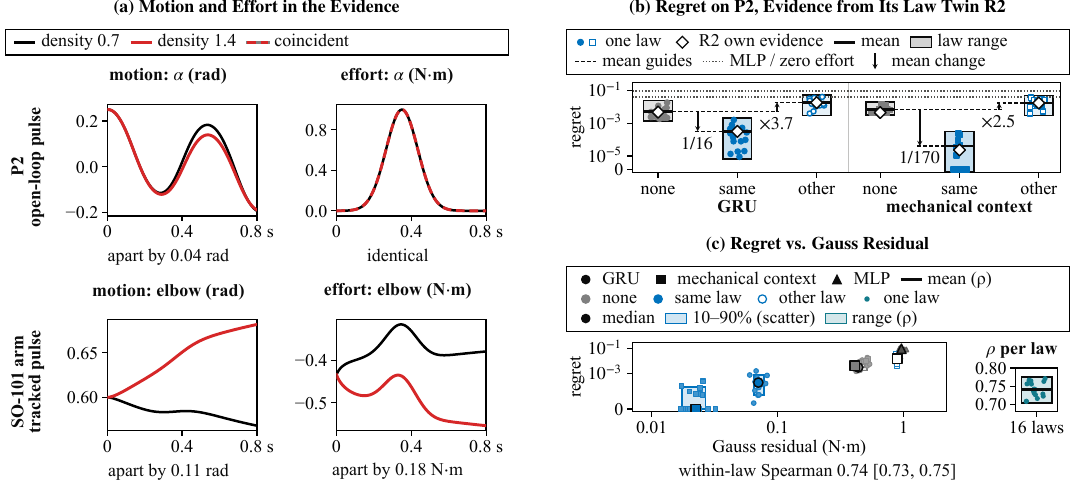}
\caption{Evidence and the cost of acting (E4). \textbf{(a)} The first 0.8\,s of a positive evidence pulse at densities 0.7 and 1.4. P2 receives identical open-loop efforts, so its law appears in motion; SO-101's tracked elbow effort changes by more than the pulse itself. \textbf{(b)} Regret on P2 under no evidence and same-law or other-law R2 evidence, for a fixed planner choosing among 64 constant-effort candidates for three targets. Arrows compare means with and without evidence; diamonds show R2 using its own evidence. \textbf{(c)} Regret versus rollout Gauss residual over 16 laws; boxes span 10--90\% around the median. Within-law Spearman correlations appear at right: mean 0.74, with 95\% law-bootstrap interval [0.73, 0.75]. Supplementary Table~\ref{tab:planning} gives the full costs and coverage.}
\label{fig:decisions}
\end{figure*}

\subsection{E4: Value of Twins for Decisions and Training}\label{sec:e4}
\label{sec:e5}
\textbf{Law-matched twin evidence improves the decisions a model supports.} We next ask whether better use of evidence matters downstream. A fixed planner selects a constant-effort candidate, and the forward half replays that choice to measure its actual cost. With R2 supplying the evidence for P2, the correct law reduces mean regret by 16 times for the GRU and 170 times for the mechanical-context model, consistently across training seeds (Fig.~\ref{fig:decisions}b). The twin's evidence serves P2 nearly as well as R2's own evidence serves R2. Other-law evidence, in contrast, increases regret relative to no evidence, which ties the decision benefit to the law conveyed by the support episode.

Unexplained effort anticipates this benefit before execution. Same-law evidence yields small residuals and nearly optimal choices, while absent or other-law evidence increases both residual and regret. The strong within-law association in Fig.~\ref{fig:decisions}c shows that this relationship persists when the underlying law is held fixed. Because the score uses the predicted rollout, known law, and query effort directly, it indicates decision quality physically, without a future replay.

\begin{table}[t]
\centering
\caption{Training on simulated corpora (E4). First-second error on held-out target densities, relative to no evidence, using evidence from the target (own) or its law twin (twin). The pairs are P2/R2 and SO-101/boxes. The law-following index is own-form, other-law error minus twin-form, same-law error; positive values favor law matching. Intervals are 95\% law-bootstrap intervals; Seeds counts runs with a positive index.}
\label{tab:e4main}
\footnotesize
\setlength{\tabcolsep}{2.2pt}
\begin{tabular}{@{}lcccccc@{}}
\toprule
 & \multicolumn{2}{c}{Same law} & \multicolumn{2}{c}{Other law} & & \\
\cmidrule(lr){2-3}\cmidrule(lr){4-5}
Corpus & Own & Twin & Own & Twin & Index & Seeds \\
\midrule
\multicolumn{7}{@{}l}{\emph{P2 (16 laws, five seeds)}}\\
No twins & 0.983 & 0.995 & 1.005 & 1.011 & \textbf{0.011} [0.005, 0.016] & \textbf{2/5} \\
Form twins & 0.946 & 0.956 & 1.053 & 1.066 & \textbf{0.097} [0.094, 0.100] & \textbf{5/5} \\
Form--law twins & 0.963 & 0.960 & 1.041 & 1.059 & \textbf{0.081} [0.079, 0.082] & \textbf{5/5} \\
\quad no effort & 0.958 & 0.952 & 1.037 & 1.061 & \textbf{0.085} [0.082, 0.088] & \textbf{5/5} \\
\midrule
\multicolumn{7}{@{}l}{\emph{SO-101 arm (12 laws, three seeds)}}\\
No twins & 0.982 & 0.973 & 1.034 & 1.034 & \textbf{0.061} [0.047, 0.075] & \textbf{3/3} \\
Form twins & 0.990 & 1.012 & 1.040 & 1.034 & \textbf{0.028} [0.022, 0.033] & \textbf{2/3} \\
Form--law twins & 0.997 & 1.000 & 1.038 & 1.038 & \textbf{0.037} [0.036, 0.039] & \textbf{2/3} \\
\quad no effort & 1.025 & 1.024 & 0.990 & 0.997 & \textbf{-0.034} [-0.036, -0.031] & \textbf{0/3} \\
\bottomrule
\end{tabular}
\end{table}

\setcounter{dbltopnumber}{2}
\begin{table*}[t]
\centering
\caption{Prediction accuracy and online cost on the SO-101/HX-30HM simulation corpus. Each forecast spans 3\,s after 18\,s of support. Lower is better; bold marks minima at displayed precision. RMSE averages eight queries and three seeds per configuration. Mean weights the 16 Test, eight OOD, and 16 Mass-test configurations equally. B1 predicts one trajectory; B32 predicts eight queries and 24 action candidates. Costs are medians across three fixed configurations. Relative costs divide each B32 measurement by FoLaSim's.}
\label{tab:so101-efficiency}
\small
\setlength{\tabcolsep}{3.0pt}
\renewcommand{\arraystretch}{1.12}
\begin{tabular*}{\textwidth}{@{\extracolsep{\fill}}lrrrrrrrrr@{}}
\toprule
 & \multicolumn{4}{c}{Prediction RMSE (${}^\circ$) $\downarrow$}
 & \multicolumn{2}{c}{Latency (ms) $\downarrow$}
 & \multicolumn{1}{c}{GPU energy $\downarrow$}
 & \multicolumn{2}{c}{B32 cost / ours $\downarrow$} \\
\cmidrule(lr){2-5}\cmidrule(lr){6-7}\cmidrule(lr){8-8}\cmidrule(l){9-10}
Method & Test & OOD & Mass-test & Mean & B1 & B32 & J / B32 call & Time & Energy \\
\midrule
TrajID-Param~\cite{kovalev2026trajid} & 0.1361 & 0.1316 & 0.0886 & 0.1162 & 550.284 & 638.755 & 60.0308 & 466.3$\times$ & 410.3$\times$ \\
Newton 1.8 + calibration~\cite{newton2026} & 0.0307 & 0.0386 & 0.0291 & 0.0316 & 2151.944 & 2749.496 & 254.7060 & 2007.3$\times$ & 1740.7$\times$ \\
Genesis 1.4.6 + calibration~\cite{genesis2026} & \textbf{0.0293} & \textbf{0.0378} & \textbf{0.0289} & \textbf{0.0309} & 5647.402 & 5829.053 & 545.2560 & 4255.5$\times$ & 3726.4$\times$ \\
Isaac Sim 6.2 + calibration~\cite{isaacsim2026} & 0.0294 & 0.0379 & \textbf{0.0289} & \textbf{0.0309} & 15949.977 & 16212.817 & 1423.6350 & 11836.1$\times$ & 9729.4$\times$ \\
\midrule
\textbf{FoLaSim response model} & 0.0633 & 0.0850 & 0.0425 & 0.0593 & \textbf{1.243} & \textbf{1.370} & \textbf{0.1463} & \textbf{1.0$\times$} & \textbf{1.0$\times$} \\
\bottomrule
\end{tabular*}
\par\vspace{5pt}
\begin{minipage}{\textwidth}\footnotesize
Timings: A800-SXM4-80GB GPUs in FP32, including transfers, synchronization, and our support encoding, after offline training, fitting, and compilation; the native engines are calibrated on the support set and integrate at 1,200\,Hz. Energy covers the GPU board, including idle draw (Supplementary Section~\ref{sup:so101-efficiency}).
\end{minipage}
\end{table*}

\label{sec:e6}
\textbf{Variation of law within a form teaches models to use physical evidence.} We now turn from decisions to training data. We train the E3 RGB context model on four coupled-class corpora matched in laws, programs, steps, and episodes. The no-twins corpus varies appearance without changing density. The form-twins corpus supplies R2 and T2 at three densities, as in physics data engines~\cite{phyparam2026,physeditworld2026}. The form--law corpus additionally pairs evidence across law twins, and the fourth corpus removes evidence efforts from those pairs.

Introducing density variation makes the law-following index positive in every seed on held-out P2 densities (Table~\ref{tab:e4main}). Pairing evidence across law twins then makes the twin's evidence as useful as the target's own, without further raising the index. The two ingredients play complementary roles: varying the law teaches sensitivity to physics, and pairing its realizations teaches transfer across forms. Removing effort has little effect here, because the fixed open-loop pulses expose the law through motion.

\textbf{Tracking shifts useful law information from motion into effort.} The picture changes when the robot tracks commands. We repeat the four-corpus comparison on SO-101, with boxes and ellipsoids as law twins and the same model with six effort channels (Fig.~\ref{fig:formlaw}d,f). Here every corpus already varies the law within a form through the robot family's link densities. All three corpora that retain efforts give positive mean law-following indices, with box evidence as useful as the arm's own. The effect is smaller than on P2 and varies across seeds in the corpora using twins (Table~\ref{tab:e4main}, lower block).

Removing evidence effort, however, reverses the preference in every seed. Under compliant tracking, the controller changes its effort to maintain the commanded motion, so the effort now carries the information that P2's fixed pulses leave in the motion (Fig.~\ref{fig:decisions}a). This contrast agrees with the channel theorem of Section~\ref{sec:invariances}.

\begin{figure*}[!t]
\centering
\includegraphics[width=.97\textwidth]{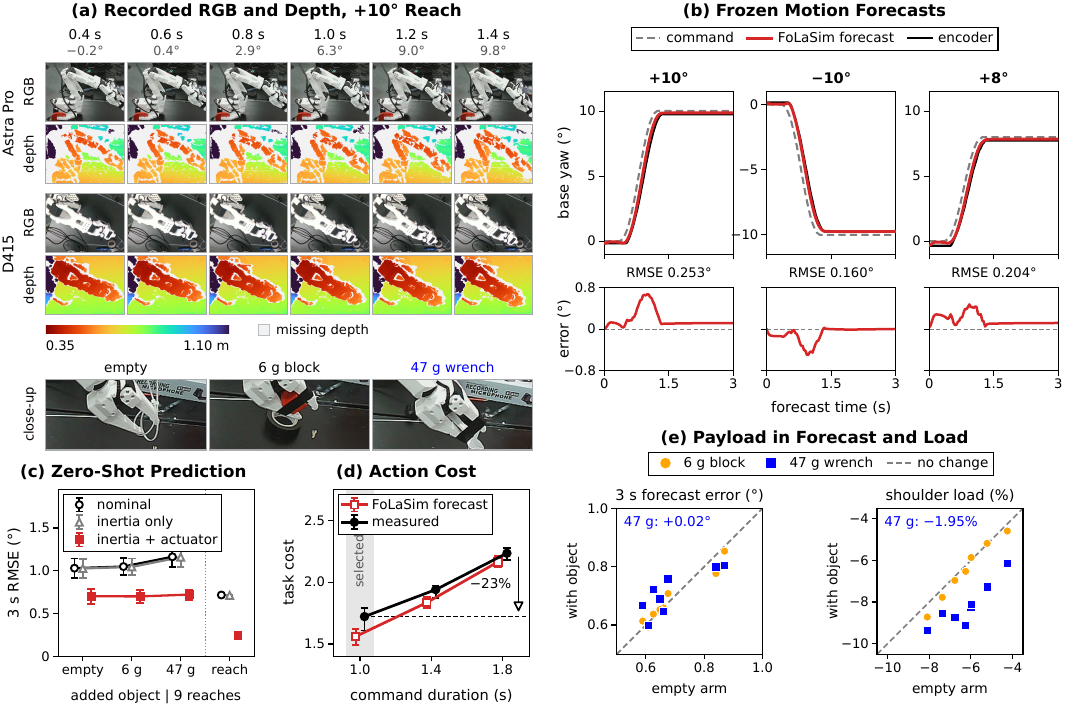}
\caption{Frozen prediction and decisions on the real SO-ARM101. \textbf{(a)} Recorded RGB/depth at six times during the $+10^\circ$ reach, with the attachments below. \textbf{(b)} All three selected reaches with the 47\,g wrench: base yaw above, prediction minus encoder below. \textbf{(c)} 3\,s RMSE over joints 1--5 after nominal, inertial, and full response variation (mean and 95\% interval over programs; three seeds). \textbf{(d)} Predicted and measured cost of the candidate durations (mean and range over three tasks); cost combines duration, peak tracking error, and task success (Supplementary Section~\ref{sup:so101-real-decisions}). \textbf{(e)} Per program, the deployed model's 3\,s error and the mean shoulder load with an attachment against the empty arm.}
\label{fig:real}
\end{figure*}

\subsection{E5: Accurate Arm Prediction at Low Online Cost}\label{sec:efficiency}

Candidate-action evaluation forecasts the same robot many times under different commands, so each forecast must be both accurate and cheap. We train a compact response model on FoLaSim's SO-101 corpus with varied inertia and HX-30HM actuator parameters (Table~\ref{tab:so101-efficiency}). It predicts 3\,s of joint motion from four initial position-command samples, future commands, and 18\,s of disjoint support history. Evaluation covers unseen configurations, out-of-range payloads, and inertial changes under a fixed servo.

We compare it with TrajID-Param~\cite{kovalev2026trajid}, adapted to the six-joint plant, and with the native Newton~\cite{newton2026}, Genesis~\cite{genesis2026}, and Isaac Sim~\cite{isaacsim2026} solvers calibrated on the same support observations (Supplementary Section~\ref{sup:so101-efficiency}).

\textbf{The learned response model combines low prediction error with sharply reduced online cost.} With an overall RMSE below $0.06^\circ$, it improves on TrajID-Param in every split, while the three calibrated physics engines attain lower errors. Its advantage lies in cost: a batch of 32 forecasts takes 1.37\,ms on an A800, with orders of magnitude lower latency and GPU energy than the measured baseline interfaces.

\textbf{As in E4, the response model follows the law because its corpus varies the law within one form.} Varying inertia and actuator response reduces Test RMSE by 82\% relative to nominal training. Since the configurations of a split share the arm and the commands, the test of E3 applies directly. With the support of another configuration, the varied model errs more than without support ($0.48^\circ$ against $0.27^\circ$ on Test; $0.06^\circ$ with its own), and its forecasts follow 90\% [85, 94] of the change of motion between configurations. Trained on one law, the same architecture ignores the support and does not adjust; in other words, it follows the form (Supplementary Tables~\ref{tab:so101-response-ablation} and~\ref{tab:so101-formlaw}).

\subsection{E6: Sim-to-Real Prediction and Action Selection}\label{sec:real}

The generated trajectories also support deployment on a physical SO-ARM101 with HX-30HM servos (Fig.~\ref{fig:real}). The deployed model keeps its simulation-trained weights and receives no support history, and each forecast is made before execution. Supplementary Section~\ref{sup:so101-real} gives the protocol and complete results.

\textbf{Varying simulated dynamics improves prediction on the real arm.} Across three attachments, full response variation consistently outperforms nominal pretraining on the same recorded programs (Fig.~\ref{fig:real}c). Small motions leave a constant-state forecast competitive, whereas larger base-yaw reaches expose the value of predicting the evolving response. On these reaches, the deployed model reduces mean 3\,s RMSE from $3.055^\circ$ for the constant-state forecast to $0.232^\circ$.

\textbf{The frozen forecasts also support effective action selection.} A fixed selector compares command durations by predicted tracking error and completion time. All three selected actions reach their joint-space targets and achieve the lowest measured cost in their respective candidate sets. Faster commands incur more tracking error, but their shorter duration yields a lower combined cost (Fig.~\ref{fig:real}d).

\textbf{On the real arm, the payload's law shows mainly in the effort.} The servos record their load with every encoder sample (Supplementary Section~\ref{sup:so101-real-effort}). The 6\,g block changes the law little and therefore sets the repeatability: $0.11^\circ$ RMS in the joint angles and at most 0.25\% of rated torque in the loads. Under the same commands, the 47\,g wrench shifts the motion by $0.33^\circ$, three times the change of the block, yet it changes the shoulder and wrist loads by 1.95\% and 0.82\% of rated torque, ten and forty times the change of the block. The channel theorem of Section~\ref{sec:invariances} predicts this pattern: the position servo moves the law into the effort and leaves only a small static offset in the motion. The frozen model, which reads positions and commands only, accordingly keeps its error ($+0.02^\circ$, Fig.~\ref{fig:real}e).

\section{Discussion}\label{sec:discussion}

Taken together, the experiments suggest a concrete principle for training data: vary the law within a form, then expose that variation through the channel that the excitation makes informative. This principle connects appearance-dependent generalization~\cite{acwmphys2026,xeworld2026} to the physical information available in a model's observations. FoLaSim supplies both the controlled variation and the diagnostic comparison, so a data choice can be judged by its effect on transfer, consistency, and decisions.

The physical anchor links these improvements to correctness. Evidence preference shows which observations help a model, and a common mechanical structure shows how its responses fit together. Only known effort, however, determines whether that structure is the right one. Because the Gauss residual is differentiable, coordinate-free, and independent of a recorded future, it is also a candidate training signal or post-training reward for learned simulators~\cite{wmpo2025,worldenv2025,vlarft2025}. The decision experiment motivates this use, and optimizing models with the residual is a direction for future work.

The current reader uses known assembly kinematics and calibrated measurement floors, with planar image readout and state-based robot evaluation. These inputs define a practical starting point for broader studies. Testing video models on robots closer to their training distribution, and fine-tuning them on certified corpora, would reveal how embodiment and training affect the zero-shot behavior reported here. Recorded robots additionally require calibrated anchors, while deformable bodies and flexible transmissions require richer laws and ports (Supplementary Section~\ref{sup:robots}).

\section{Conclusion}\label{sec:conclusion}

We asked whether a world action model has learned the law that turns effort into motion or the form visible in its frames. FoLaSim answers this question by making physical law controllable across visible forms and measurable from predicted motion. Certified twins provide the controlled comparisons, unexplained effort and implied-law identification explain their physical consequences, and four tests connect them to correctness within a declared law class. Our experiments establish faithful synthesis through robot contact, informative fault diagnosis, and a link to decision quality. They also show how variation of the law and the excitation make physical evidence useful for learning. Arm trajectories that vary the law within one form train a compact predictor that reads the law from evidence, forecasts in milliseconds, and transfers to a real arm with frozen weights, where a payload's law shows mainly in the effort. Together, these capabilities provide a scalable foundation for evaluating world action models and for constructing training data that carries the law across forms.

\clearpage
\setcounter{page}{1}
\renewcommand{\thepage}{S\arabic{page}}
\setcounter{dbltopnumber}{\FLSdbltopnumber}
\setlength{\tabcolsep}{\FLStabcolsep}
\setcounter{section}{0}
\setcounter{subsection}{0}
\setcounter{subsubsection}{0}
\setcounter{paragraph}{0}
\setcounter{equation}{0}
\setcounter{figure}{0}
\setcounter{table}{0}
\setcounter{theorem}{0}
\setcounter{proposition}{0}
\setcounter{lemma}{0}
\setcounter{corollary}{0}
\setcounter{definition}{0}
\setcounter{remark}{0}
\setcounter{algorithm}{0}
\setcounter{footnote}{0}
\renewcommand{\thesection}{S\arabic{section}}
\renewcommand{\thesubsection}{S\arabic{section}.\arabic{subsection}}
\renewcommand{\theequation}{S\arabic{equation}}
\renewcommand{\thetable}{S\arabic{table}}
\renewcommand{\thefigure}{S\arabic{figure}}
\renewcommand{\thetheorem}{S\arabic{theorem}}
\renewcommand{\theproposition}{S\arabic{proposition}}
\renewcommand{\thelemma}{S\arabic{lemma}}
\renewcommand{\thecorollary}{S\arabic{corollary}}
\renewcommand{\thedefinition}{S\arabic{definition}}
\renewcommand{\theremark}{S\arabic{remark}}
\renewcommand{\thealgorithm}{S\arabic{algorithm}}
\makeatletter
\let\maketitle\FLSmaketitle
\let\@maketitle\FLS@maketitle
\let\thanks\FLSthanks
\let\@IEEEtitleabstractindextext\relax
\makeatother
\title{Supplementary Material: FoLaSim}
\author{Qibin~Gu and Yunfeng~Yan}
\markboth{Supplementary Material}{FoLaSim: A Form--Law Twin Simulator for World Action Models}
\maketitle
This document accompanies \emph{FoLaSim: A Form--Law Twin Simulator for Dynamics Understanding in World Action Models}. S-prefixed numbers refer to this supplement; equations numbered without the prefix are those of the main paper.

This supplement is organized in the same order as the main paper. Section~\ref{sup:related} extends the related work. Section~\ref{sup:why} develops the mechanics of form, law, and twins, i.e., what a forecast can and cannot show of the law and what the four tests guarantee. Section~\ref{sup:simulator} details the forward half, from a specification to certified twins and their videos, and Section~\ref{sup:measurement} details the inverse half, from a forecast to its residuals and to the crossing of the twins. Section~\ref{sup:experiments} gives the protocols and the complete results of E1--E4, Section~\ref{sup:extensions} gives the extensions to contact and recorded robots, and Sections~\ref{app:proofs} and~\ref{app:numerics} give the proofs and the numerical checks.

\section{Extended Related Work}\label{sup:related}

\subsection{Physical Understanding of Video and World Models}
Benchmarks for video generation assess physical commonsense and principles~\cite{S:bansal2024videophy,S:meng2024phygenbench,S:motamed2026physicsiq,S:physicsiqverified2026}, isolate physics concepts~\cite{S:upadhyay2026worldbench}, compare against real Newtonian systems~\cite{S:tragoudaras2026morpheus}, and measure task-specific observable relations~\cite{S:gao2026lastexam}; ACWM-Phys studies physical interaction in action-conditioned prediction~\cite{S:acwmphys2026}. Controlled studies report case-based generalization in which color, size, velocity, and shape are weighted in that order~\cite{S:kang2025howfar}, show that latent prediction error can respond to appearance rather than physical possibility~\cite{S:latentsurprise2026}, and find that verdicts change with evaluator configuration~\cite{S:tetrastate2026} and across output modalities of one model~\cite{S:twostories2026}. WorldTest evaluates world-model learning through several kinds of environment-level queries rather than single trajectories~\cite{S:warrier2026worldtest}, and latent rollout consistency characterizes faithful latent dynamics~\cite{S:limitslatent2026}. We adopt their behavioral operationalization and multi-query view, and further ask whether the dynamics behind correct-looking forecasts are reusable laws.

\subsection{World Models as Simulators and Physics Data Engines}
World models are trained from autonomous robot play~\cite{S:playworld2026} and evaluated as embodied simulators for fidelity and functional utility~\cite{S:worldarena2026,S:worldarena2_2026,S:wowworldeval2026}. Physics data engines record the intermediate world states of graphics production~\cite{S:cgworld2026}, and related work post-trains video models toward physically faithful continuations~\cite{S:phyworld2026,S:li2025pisa}; scenes that keep their appearance while their physics varies expose visual world models~\cite{S:physicallyviable2026}. These works vary or recover the law of a fixed form. Different from them, the forward half of FoLaSim also provides certified changes of form under a fixed law.

\subsection{Action Consistency of World Action Models}
ACID scores predicted transitions by the residual of a separately trained inverse dynamics model and proves that, for linear latent dynamics and an ideal inverse model, the residual bounds the action-explainable error~\cite{S:acid2026}. WISE selects among jointly generated action--future candidates using inverse-dynamics consistency, action validity, and task progress~\cite{S:wise2026}. Related diagnostics test action--state consistency~\cite{S:futurecompatible2026}, decode with self-verification~\cite{S:wcd2026}, measure action following~\cite{S:followactions2026}, benchmark against real executions~\cite{S:worldoracle2026}, or reward out-of-distribution actions without paired video~\cite{S:pavxplorerl2026}. These methods judge a candidate against its action. Different from them, our measures concern relations among predictions across mechanisms, physics, and interventions. In particular, the identity of Section~\ref{sec:core} in the main paper is a closed-form mechanical counterpart of the action-explainable error of ACID.

\subsection{Adaptation to Physical Change}
CoPhy infers latent physical factors from a factual process to predict counterfactual outcomes~\cite{S:baradel2020cophy}, and CausalWorld exposes interventions on masses and other factors~\cite{S:ahmed2021causalworld}. PhysiCalWorld extracts in-context physics descriptors from exploratory interaction without test-time weight updates~\cite{S:physicalworld2026}; CALIPER tests whether calibration evidence is actually used~\cite{S:caliper2026}; InfiniteWorld separates visual evidence from familiar physics~\cite{S:infiniteworld2026}. Studies of parameter shifts separate knowing a parameter from using it~\cite{S:causalparamshift2026} and factorization from stability~\cite{S:conservefactor2026,S:stablecounterfactual2026}, and PokeWorld separates the availability of physical information in observations from its acquisition by latent models~\cite{S:pokeworld2026}. Transformers can also adapt in context across dynamical systems~\cite{S:transferoperators2026}. These works establish adaptation within an instance or a family of systems. Different from them, we test whether the adapted knowledge transfers across certified mechanical realizations.

\subsection{Shared Physical Explanations across Interventions}
OneWorld trains and evaluates world models so that futures under different actions share one physical explanation, reporting branch-wise, shared, and gap errors against independent simulation~\cite{S:oneworld2026}. IMPLY inverts a simulator to read the physics implied by rollouts, anchors the shared fit to observed calibration pushes, and shows that a model can be highly self-consistent yet wrong~\cite{S:imply2026}. We adopt their comparison of individual and shared fits and their anchoring, and we extend consistency across control channels, SI ports, and certified correspondences between mechanisms.

\subsection{Cross-Embodiment and Cross-System Dynamics}
XEWorld holds out robot embodiments in matched scenes~\cite{S:xeworld2026}; particle-based world models scale across hand embodiments~\cite{S:crossembodiment2025dex}; morphology-conditioned~\cite{S:qwm2026} and body-agnostic~\cite{S:hourglass2026} world models transfer across robots. Earlier work shared latent dynamics across systems~\cite{S:wang2021sidm}, learned cross-domain correspondences through dynamics cycle-consistency~\cite{S:zhang2021cycle}, and learned transferable physics with graph networks~\cite{S:sanchezgonzalez2018graph,S:sanchezgonzalez2020gns}. In these settings appearance, kinematics, action interfaces, and dynamics usually change together, or the correspondence is learned from the predictions themselves. Different from these settings, we fix the correspondence by mechanical construction and verify it independently, so the equivalence does not have to be inferred from the outputs being evaluated. Exact-state and visual conditions then separate failures of dynamical reuse from failures of observation or representation.

\subsection{Structure, Identifiability, and Physics Learning}
Hamiltonian~\cite{S:greydanus2019hamiltonian}, Lagrangian~\cite{S:cranmer2020lagrangian,S:lutter2019delan}, dissipative~\cite{S:zhong2020dissipative} and compositional~\cite{S:neary2023compositional} port-Hamiltonian, and constrained Lagrangian~\cite{S:daems2024keycld} models build mechanical structure into learning. PHAST shows that accurate prediction need not identify physical parameters and discusses anchoring and gauge freedom~\cite{S:phast2026}; linear port-Hamiltonian systems are learnable only up to an equivalence class~\cite{S:ortega2024learnability}; Helmholtz metrics quantify whether a learned field admits a Lagrangian~\cite{S:wolf2025helmholtz}; and the non-uniqueness of Lagrangians is classical~\cite{S:douglas1941inverse}. Closed-loop data identify a system only under sufficient excitation~\cite{S:ljung1999system,S:forssell1999closed}, control and learning conflict~\cite{S:feldbaum1960dual}, excitation trajectories are designed for robot identification~\cite{S:swevers1997optimal}, and only base inertial parameters are identifiable from motion~\cite{S:gautier1990minimum}. For controlled latent world models, identifying transitions requires state-conditional action variation~\cite{S:controlledwmid2026}. These results determine the evidence and excitation conditions under which our cross-mechanism queries are informative.

\subsection{Mechanics and Relational Testing}
Gauss's principle of least constraint~\cite{S:gauss1829} and its modern forms~\cite{S:udwadia1992new}, Gauss's geometry of surfaces~\cite{S:gauss1828disquisitiones}, geometric mechanics~\cite{S:bullo2005geometric}, port-Hamiltonian systems~\cite{S:vanderschaft2014port}, and multibody dynamics~\cite{S:kane1985dynamics,S:featherstone2008rigid} supply our physical language; nonsmooth mechanics~\cite{S:moreau1988unilateral,S:jean1999nonsmooth,S:desaxce1991bipotential} defines the additional requirements for extending the smooth-phase evaluation to contact. Testing a system through relations among its outputs is known as metamorphic testing~\cite{S:chen2018metamorphic}. Our contribution lies in the choice of the mechanical relations, their guarantees, and their diagnostic power for WAMs.

\subsection{Comparison of Capabilities}\label{sup:capabilities}
We compare simulators, data engines, and benchmarks by the capabilities they provide rather than by a common score, since no other approach produces certified law twins or reads a forecast back to a law, and there is therefore no shared metric to rank them. Table~\ref{tab:compare} of the main paper marks which representative approaches provide each of the components (i)--(iv) of the main-paper introduction; Table~\ref{tab:capabilities-compared} adds, by category, what each kind of approach provides instead, as described above.

\begin{table*}[t]
\centering
\caption{Requirements (i)--(iv) of the main-paper introduction against the approaches of Section~\ref{sup:related}, by category. $\checkmark$: provided as part of the method; --: not provided; text: what is provided instead.}
\label{tab:capabilities-compared}
\footnotesize
\setlength{\tabcolsep}{3pt}
\begin{tabularx}{\textwidth}{@{}L{3.6cm}ccccYL{1.6cm}@{}}
\toprule
 & \multicolumn{3}{c}{(i) Instances} & (ii) & (iii) & (iv) \\
\cmidrule(lr){2-4}
Approach & \makecell{Law varied,\\form fixed} & \makecell{Form varied,\\law fixed} & Certified & \makecell{Efforts\\known} & A forecast is scored by & Tests \\
\midrule
Physics engines and robot-learning frameworks~\cite{S:todorov2012mujoco,S:mujocoplayground2025,S:isaaclab2025,S:dmcontrol2020,S:robosuite2020} & by hand & -- & -- & $\checkmark$ & -- (they generate motion, they do not score it) & -- \\
Physics data engines~\cite{S:phyparam2026,S:phyco2026,S:physeditworld2026,S:physinone2026} & $\checkmark$ & -- & -- & forces or actions as conditions & -- & -- \\
Plausibility benchmarks for video~\cite{S:bansal2024videophy,S:meng2024phygenbench,S:motamed2026physicsiq,S:upadhyay2026worldbench} & -- & -- & -- & -- & plausibility judgments or agreement with recorded video & -- \\
Embodied world-model benchmarks~\cite{S:ewmbench2025,S:worldarena2026,S:smallworlds2025,S:acwmphys2026} & -- & -- & -- & actions & visual and task fidelity, or prediction error & -- \\
Identical looks, hidden physics~\cite{S:pokeworld2026,S:physicallyviable2026,S:physicalworld2026,S:caliper2026} & $\checkmark$ & -- & -- & pushes, strikes, or exploratory actions & prediction error or probes & -- \\
Held-out embodiments~\cite{S:xeworld2026} & -- & with its law & -- & actions & prediction error & -- \\
Inverse-dynamics checks~\cite{S:acid2026,S:wise2026,S:wav2026} & -- & -- & -- & actions & a learned inverse-dynamics residual & -- \\
Shared explanations~\cite{S:oneworld2026,S:imply2026} & -- & -- & -- & actions & one physics fitted across actions, anchored by calibration pushes in~\cite{S:imply2026} & consistency, anchor \\
Closed-chain evaluator~\cite{S:paper1} & -- & -- & -- & $\checkmark$ & an inverse-dynamics residual on closed chains & -- \\
\addlinespace[2pt]
FoLaSim & $\checkmark$ & $\checkmark$ & $\checkmark$ & $\checkmark$ & effort left unexplained, in SI units against a per-forecast floor, and the law it implies & reuse, adjustment, consistency, anchor \\
\bottomrule
\end{tabularx}
\end{table*}

\section{Form, Law, and Twins}\label{sup:why}

In this section, we give the mechanics behind Section~\ref{sec:why} in the main paper. Table~\ref{tab:invariances} summarizes the three spaces of Section~\ref{sec:manifold} in the main paper, i.e., what each space contains, what the inverse half measures there, and from which data it is decided.

\begin{table*}[t]
\centering
\caption{The three spaces of FoLaSim. Physics, and the forward half of FoLaSim, produce a video from law through state to form; the inverse half reads a forecast from form through state to law. Everything in form space is decided from the video alone; everything in law space needs the effort.}
\label{tab:invariances}
\footnotesize
\begin{tabularx}{\textwidth}{@{}L{1.1cm}L{1.9cm}YYL{3.2cm}@{}}
\toprule
Space & Layer & Holds & The inverse half measures & Decided from \\
\midrule
Form & Parts & pose of every body in every frame & readability and coverage of each body & one frame \\
 & Relations & joints, gears, and closed loops between the bodies & closure of the assembly; form verdict & one frame \\
 & Motion & motion of the assembled bodies & tangent velocity, normal acceleration & several frames \\
\addlinespace[2pt]
State & -- & degrees of freedom $s$ and velocities $\dot s$, with the effort $u$ attached & declared coordinates, derivatives, floors & frames and the query \\
\addlinespace[2pt]
Law & Components & inertia, potential, dissipation, input port & response to effort; fitted coefficients & frames and effort \\
 & Balance & the components in the one equation \eqref{eq:reduced} & Gauss residual $\delta u$ in \Nm{}; power residual $r_E$ & frames and effort \\
 & Law & one law behind all interventions, equal to the true one & consistency and anchor; reuse and adjustment across twins & sets of forecasts and effort \\
\bottomrule
\end{tabularx}
\end{table*}

\subsection{Setting}
Consider a robot--object system with generalized coordinates $q\in\R^n$, velocity $v=\dot q$, and dynamics
\begin{equation}
\begin{aligned}
M(q)\dot v + c(q,v) &= F + J(q)\Tr\nu,\\
F &= B(q)u + Q_d(q,v) - \nabla V(q),
\end{aligned}
\label{eq:full}
\end{equation}
where $M\succ0$ is a smooth mass matrix, $c$ collects Coriolis and centrifugal terms, $B$ maps actuator efforts $u$ to generalized forces, $Q_d$ are dissipative forces, $V$ is the potential, and $\nu$ are the reactions of holonomic constraints $\phi(q)=0$ whose Jacobian $J$ has full row rank $m$. We consider ideal, time-independent bilateral constraints; contact switches and actuator saturation require separate treatment. Admissible configurations form the constraint manifold $S=\phi^{-1}(0)$ of dimension $d=n-m$, and the columns of $N(q)\in\R^{n\times d}$ span $\ker J(q)$. The kinetic energy defines the metric $g(v,w)=v\Tr Mw$, and $P_T=N(N\Tr MN)^{-1}N\Tr M$ is the $g$-orthogonal projector onto $T_qS$. For a curve on $S$, the covariant acceleration $D_t\dot q=\ddot q+\Gamma(\dot q,\dot q)$ splits by Gauss's formula into a tangential part and a normal part fixed by the second fundamental form $\II$ of $S$:
\begin{equation}
D_t\dot q=\nabla^S_t\dot q+\II(\dot q,\dot q).
\label{eq:gaussformula}
\end{equation}
At a fixed admissible $(q,v)$, Gauss's least-constraint principle selects
\begin{equation}
a^\ast=\arg\min_{Ja+\dot Jv=0}\tfrac12\left\|a-M^{-1}(F-c)\right\|_M^2.
\label{eq:least-constraint}
\end{equation}
Its stationarity condition is $Ma^\ast+c-F=J^\top\nu$, with a reaction multiplier enforcing the acceleration constraint. Since $MD_t\dot q=F+J\Tr\nu$ and $P_TM^{-1}J\Tr=0$, tangential and normal projection give
\begin{equation}
\begin{aligned}
\nabla^S_t\dot q&=P_TM^{-1}F,\\
\II(\dot q,\dot q)&=(I-P_T)M^{-1}\big(F+J\Tr\nu\big).
\end{aligned}
\label{eq:split}
\end{equation}
In other words, the applied forces determine the tangential acceleration, the geometry of $S$ determines the normal acceleration, and the reactions adjust to enforce the latter.

\subsection{Observation Layer}\label{sec:observation-layer}
A WAM outputs images rather than states. The images that show one physical state $(q,v)\in TS$ under different lighting, texture, background, or viewpoint form a fiber over that state, and the readout $\pi:\mathcal O\to TS$ maps each image to the state it shows. A difference between two forecasts therefore has a component along the fibers, which changes appearance only, and a component across them, which changes the physics. Pixel distances weigh both together: a static background contributes pixels but no motion, which dilutes physical errors, while two law twins differ in every pixel and share their physics. The evaluation uses only the component across the fibers. This has two consequences for the readout. First, recovering the pose of every body in $Q$ rather than projecting onto $S$ keeps the normal layer measurable. Second, the readout error defines a tube around the readout of the true motion: deviations smaller than its radius cannot be resolved by any evaluator that uses this readout, which is why calibration floors accompany every score (Section~\ref{sup:state}).

\subsection{Timing Invariance of Admissibility}

\begin{lemma}[Admissibility is invariant to timing]\label{lem:timing}
Let $q:[0,T]\to S$ be $C^2$, let $\sigma:[0,T']\to[0,T]$ be a $C^2$ increasing bijection, and let $\tilde q=q\circ\sigma$. Then $\tilde q$ preserves the following holonomic admissibility conditions: $\phi(\tilde q)=0$, $J(\tilde q)\dot{\tilde q}=0$, and $(I-P_T)D_t\dot{\tilde q}=\II(\dot{\tilde q},\dot{\tilde q})$. Its tangential acceleration is $\dot\sigma^2\nabla^S_t\dot q+\ddot\sigma\,\dot q$, evaluated at $\sigma(t)$, which in general violates \eqref{eq:split}.
\end{lemma}

Choosing $\sigma(0)=0$ and $\dot\sigma(0)=1$ keeps the initial state, so one initial state admits infinitely many admissible time laws, whereas a specified input selects one dynamical trajectory under the usual local uniqueness conditions.

\subsection{Reaction-Space Force Errors}

\begin{lemma}[Reaction-space errors are invisible in motion]\label{lem:reaction}
Replacing $F$ by $F+J\Tr\mu$ for any $\mu(t)$ leaves the constrained motion unchanged and shifts the reaction from $\nu$ to $\nu-\mu$. Only the $d$ components $N\Tr F$ of a force error affect the motion.
\end{lemma}

\subsection{Observation Channels}

\begin{definition}[Observation channel and blind set]\label{def:channel}
An observation channel $\mathcal C$ comprises the observed quantities (for example poses, velocities, and actuator efforts), the action semantics (torque or position commands), the controller, the distribution of excitation over states and inputs, and the sampling and readout errors. For a model class $\Theta$ and a true law $\theta^\ast$, the blind set $\mathcal B_{\mathcal C}(\theta^\ast)$ is the set of $\theta\in\Theta$ under which $\mathcal C$ produces the same distribution of observations as under $\theta^\ast$.
\end{definition}

A model fitted through a training channel $\mathcal C_{\rm tr}$ and scored through an evaluation channel $\mathcal C_{\rm ev}$ can carry an error $\theta\neq\theta^\ast$ undetected when $\theta\in\mathcal B_{\mathcal C_{\rm tr}}(\theta^\ast)\cap\mathcal B_{\mathcal C_{\rm ev}}(\theta^\ast)$, and approximately so when the observation distributions differ by less than the data and tolerances can resolve. Each transformation in Table~\ref{tab:invariances} under which physics changes but a channel does not is a blind direction of that channel. An SI-effort evaluation channel resolves timing, feedback, and scale errors that geometric and pixel metrics leave unresolved; reaction-space errors remain invisible to it as well, but they do not change the predicted motion.

Two kinds of mismatch follow. Some channels have exact observational equivalences, such as passive scale ambiguity and reaction-space perturbations; others have weak sensitivity or finite numerical resolution, as under high-gain feedback or discrete sampling. The physics has symmetries that the pixel channel lacks, e.g., the same law realized in different forms, and models are distracted there by appearance~\cite{S:kang2025howfar}. The relevant behavioral question is whether the model preserves certified law equivalences while responding to changes that alter the physical predictions.

Robot data are usually recorded under feedback $u=\pi(q,v,r)$, for example servo tracking of a reference $r$, so a model trained on tracked motion can fit its rollouts while responding to torque far too weakly. The following theorem quantifies, in a smooth, unsaturated, high-gain regime, how much information about the plant the motion and the effort then carry. We write the law as $\tau_\theta(s,\dot s,\ddot s)=u$ with $\tau_\theta=M_S\ddot s+C_S\dot s+\nabla V_S+d_S$ and parameters $\theta\in\R^p$, and we let $\tau_\theta(t)$ denote the effort that $\theta$ requires to follow a reference $r(t)$ exactly.

\begin{theorem}[Information carried by observation channels]\label{thm:channels}
Let the effective dynamics be fully actuated with known input map, normalized to $B_S=I$. Configurations are observed at times $t_1,\dots,t_N$ with independent Gaussian noise of covariance $\sigma_q^2I$, and, when the effort port is observed, efforts with covariance $\sigma_u^2I$. Let $\mathcal F$ be the Fisher information about $\theta$ at the true law, and $Y_j=\partial\tau_\theta(t_j)/\partial\theta$. For the tracking claims, assume a compact parameter neighborhood, a uniformly positive-definite mass matrix, smooth coefficients with bounded derivatives, a $C^5$ reference, and uniformly exponentially stable fast tracking dynamics. Gains and observation-noise covariances are independent of $\theta$; no input saturation or constraint-mode switch occurs. Solutions and parameter sensitivities remain in the stated smooth domain.
\begin{itemize}
\item[(i)] \emph{Passive motion.} If the family is closed under scaling, $\tau_{c\theta}=c\,\tau_\theta$, the scale direction lies in $\ker\mathcal F$.
\item[(ii)] \emph{Open loop with known input.} $\mathcal F=\sigma_q^{-2}\sum_jS_j\Tr S_j$ with sensitivities $S_j=\partial s(t_j)/\partial\theta$. A parameter direction $z$ is informative precisely when $S_jz\ne0$ for at least one observed time. Nonzero effort can break passive scale ambiguity, but the sampling schedule must observe the resulting sensitivity.
\item[(iii)] \emph{Tracking, motion only.} Let $u=K_p(r-s)-K_d\dot s$ with $K_p=\gamma\Gamma_p$ and $K_d=\sqrt\gamma\,\Gamma_d$ for fixed $\Gamma_p,\Gamma_d\succ0$, the reference $r$, an initial state on the reference, and sampling times in $[t_0,T]$ with $t_0>0$. As $\gamma\to\infty$,
\begin{equation}
F_{\rm motion}=\frac{1}{\gamma^2\sigma_q^2}\sum_{j}Y_j\Tr\Gamma_p^{-\top}\Gamma_p^{-1}Y_j+O(\gamma^{-5/2}).
\label{eq:fisher-motion}
\end{equation}
\item[(iv)] \emph{Tracking with the effort port.} Under the same conditions, $F_{\rm effort}=\sigma_u^{-2}\sum_jY_j\Tr Y_j+O(\gamma^{-1/2})$.
\item[(v)] \emph{Constraints.} In the full model \eqref{eq:full}, force perturbations $F\mapsto F+J\Tr\mu$ carry no information in any channel that observes only motion.
\end{itemize}
The leading terms in (iii)--(iv) also hold for velocity-error feedback $u=K_p(r-s)+K_d(\dot r-\dot s)$ under the same stability conditions.
\end{theorem}

\begin{corollary}[Evidence ladder for density form twins]\label{cor:ladder}
For the common law \eqref{eq:common} with scalar gain $K_p=\gamma\gamma_p$, the information about the scale $c$ of the twin $c\lambda$ (Proposition~\ref{prop:scale}) is zero under passive observation, $\sigma_q^{-2}K_p^{-2}\sum_j\tau_\lambda(t_j)^2+O(\gamma^{-5/2})$ under tracking with motion only, and $\sigma_u^{-2}\sum_j\tau_\lambda(t_j)^2+O(\gamma^{-1/2})$ when efforts are observed.
\end{corollary}

The theorem is an asymptotic statement about the specified tracking channel, rather than a guarantee for arbitrary finite-gain controllers. The leading high-gain information has null space $\ker\sum_jY_j\Tr Y_j$, fixed by the excitation of the reference. Higher-order finite-gain sensitivities can still distinguish some of these directions. Passive scale symmetry persists across passive initial conditions, while reaction-space perturbations remain invisible to motion under ideal constraints. Classical closed-loop identification~\cite{S:ljung1999system,S:forssell1999closed}, dual control~\cite{S:feldbaum1960dual}, and excitation design~\cite{S:swevers1997optimal} contain these ingredients, and recent analyses of controlled world models~\cite{S:controlledwmid2026} and of physical information in latent models~\cite{S:pokeworld2026} study related questions for learned models; Theorem~\ref{thm:channels} makes the effects quantitative for the channels of the forward half (Section~\ref{sec:constructions} in the main paper), and Section~\ref{app:numerics} verifies the asymptotics numerically. Its constructions test whether these channel properties predict which capabilities of a WAM dissociate across mechanically equivalent realizations.

\subsection{Correspondence of Motions}
By (T4) of Definition~\ref{def:equivalence} in the main paper, input powers of law twins agree: $(B_Au_A)\Tr\dot s=(B_BKu_A)\Tr D\psi\,\dot s$.
\begin{proposition}[Motions correspond]\label{prop:correspond}
Under Definition~\ref{def:equivalence} in the main paper, $s_A(t)$ solves \eqref{eq:reduced} for $A$ with input $u_A(t)$ if and only if $\psi(s_A(t))$ solves \eqref{eq:reduced} for $B$ with input $Ku_A(t)$.
\end{proposition}

\subsection{Curvature of the Kinetic Metric}
\begin{remark}[One versus several degrees of freedom]\label{rem:curvature}
Every one-dimensional metric is flat: with arc length $\int\sqrt{M_S}\,ds$ as coordinate the inertia becomes one, so correspondence of one-degree-of-freedom mechanisms amounts to matching arc length together with (T2)--(T4). For $d\ge2$ the induced metric can be curved, and isometries preserve curvature (Gauss's Theorema Egregium~\cite{S:gauss1828disquisitiones}); matched curvature is necessary for twins and serves as a coordinate-free test quantity. For a planar two-link arm, $M_S=\left[\begin{smallmatrix}a+2b\cos s_2 & c+b\cos s_2\\ c+b\cos s_2 & c\end{smallmatrix}\right]$ has Gaussian curvature
\begin{equation}
K_g(s_2)=\frac{b\,(ac-b^2-c^2)\cos s_2}{\big(ac-c^2-b^2\cos^2 s_2\big)^2},
\label{eq:curvature}
\end{equation}
which is generically nonzero and changes sign at $s_2=\pm\pi/2$, for nondegenerate parameters with $b\ne0$. Two decoupled sliders are flat and cannot realize this nonflat metric by an isometry.
\end{remark}

\subsection{Energy Ledger}
With momentum $p=M_S\dot s$ and state $x=(s,p)$, the effective dynamics \eqref{eq:reduced} take the port-Hamiltonian form
\begin{equation}
\begin{aligned}
\dot x&=(\mathcal J-\mathcal R)\nabla H+Gu,\qquad y=G\Tr\nabla H,\\
\dot H&=-\nabla H\Tr\mathcal R\nabla H+y\Tr u,
\end{aligned}
\label{eq:ph}
\end{equation}
with $H=\tfrac12p\Tr M_S^{-1}p+V_S$, the canonical skew-symmetric interconnection $\mathcal J$, the positive-semidefinite resistive structure $\mathcal R$ (or a nonlinear dissipative relation) representing $d_S$, and $G=[0;\,B_S]$. In velocity coordinates, the Christoffel identity $\dot s^\top C_S\dot s=\tfrac12\dot s^\top\dot M_S\dot s$ cancels the metric derivative in $\dot H$. Multiplying \eqref{eq:reduced} by $\dot s^\top$ therefore gives $\dot H=u^\top B_S^\top\dot s-\dot s^\top d_S$, whose integral along a forecast is the power residual \eqref{eq:energy}. Four consequences follow. (i) $(H,\mathcal R,G)$ does not depend on the intervention: one ledger must explain all initial states, input programs, and channels. (ii) This holds only up to declared equivalences: energy offsets, coordinate changes, and the equivalence classes of port-Hamiltonian realizations~\cite{S:ortega2024learnability}. (iii) Once corresponding physical trajectories and ports are specified, the energy balance is scalar and independent of their coordinate representation. (iv) The input port fixes the energy scale: with $u$ known in SI units, power balance breaks the scale symmetry of passive motion (Proposition~\ref{prop:scale}).

\subsection{Common Structure across Physical Relations}
Table~\ref{tab:capabilities} collects the capabilities that operationalize dynamics understanding and the anchor that makes their measurement correct. The same metric couples relations that are usually measured separately. In \eqref{eq:reduced}, $M_S$ determines the instantaneous response to effort, $\partial\ddot s/\partial u=M_S^{-1}B_S$, with collocated response $\Lambda=B_S\Tr M_S^{-1}B_S$; it determines the velocity-quadratic terms through its Christoffel symbols; and it determines the kinetic energy $T=\tfrac12\dot s\Tr M_S\dot s$ entering \eqref{eq:energy}. An explanation fitted to impulse responses therefore predicts free-motion terms and energy changes without further fitting; for $d=1$ with configuration-dependent inertia $I(s)$, the response-implied $I(s)$ fixes the term $\tfrac12I'(s)\dot s^2$. We exploit this coupling as cross-relation hold-out (Section~\ref{sec:relational} in the main paper). The response also obeys order relations that need no parameter values.

\begin{proposition}[Order relations of the action response]\label{prop:order}
$\Lambda=B\Tr N(N\Tr MN)^{-1}N\Tr B$ is symmetric positive semidefinite (reciprocity). At a fixed configuration and input map, if mass is added, $M'=M+\Delta M$ with $\Delta M\succeq0$, or bilateral constraints are added, $\operatorname{span}N'\subseteq\operatorname{span}N$, then $\Lambda'\preceq\Lambda$.
\end{proposition}

At the matched state and port, positive-semidefinite mass addition cannot increase the collocated response, and reciprocity couples orthogonal inputs. Changes that also alter geometry or the actuator map require their own correspondence before this order comparison applies.

\begin{table*}[t]
\centering
\caption{Capabilities operationalizing dynamics understanding, and the anchor that makes the measurement correct.}
\label{tab:capabilities}
\footnotesize
\begin{tabularx}{\textwidth}{@{}L{2.3cm}YYY@{}}
\toprule
Capability & What the experiment changes & Expected behavior & Alternative explanation excluded \\
\midrule
(C1) Reuse & Mechanism, appearance, and visible motion, with certified equivalent effective dynamics & After the public maps, compatible and correct predictions for unseen actions & Memorized appearance and typical motions \\
(C2) Adjustment & Inertia, potential, or dissipation, with mechanism and appearance fixed and declared dynamic evidence & Responses change in the right direction and by the right amount & Ignoring evidence; applying an average law \\
(C3) Consistency & Initial state, actions, probes, and control channel, within one physical instance & One law $(H,\mathcal R,G)$ explains all predictions & Each prediction plausible alone but implying different systems \\
(A) Anchor & Simulator-held law or independent calibration & Consistent predictions also match the true instance & Consistently predicting a wrong system; collapse to static predictions \\
\bottomrule
\end{tabularx}
\end{table*}

\subsection{Quantitative Characterization}
With measurement error, fix a reference kinetic metric transported between twins and let $\|\delta a\|_{\mathcal S}$ be the maximum, over interventions in $\mathcal S$, of the RMS field error at matched query states. Let
\begin{equation}
\kappa_A=\sup_{\theta\in\Theta}\frac{\|a_\theta^\ast-a_{\lambda_0}^\ast\|_{\mathcal I}}{\|a_\theta^\ast-a_{\lambda_0}^\ast\|_{\mathcal I_A}},\qquad 0/0:=0,
\label{eq:kappa}
\end{equation}
which measures how stably the anchor queries identify the law on the full query set.
\begin{theorem}[Quantitative characterization]\label{thm:quantitative}
Let a path from $\omega_0$ to $\omega=(m,\lambda)$ in $\Omega$ use $n_1$ certified twin pairs and $n_2$ law changes. Suppose that $\|\hat a_{(B,\lambda')}-(\psi,K)_\ast\hat a_{(A,\lambda')}\|_{\mathcal I}\le\varepsilon_1$ on its twin pairs, $\|(\hat a_{(m',\lambda_2)}-\hat a_{(m',\lambda_1)})-(a^\ast_{\lambda_2}-a^\ast_{\lambda_1})\|_{\mathcal I}\le\varepsilon_2$ on its law changes, $\|\hat a_{\omega_0}-a^\ast_{\theta_0}\|_{\mathcal I}\le\varepsilon_3$ for some $\theta_0\in\Theta$, and $\|\hat a_{\omega_0}-a^\ast_{\lambda_0}\|_{\mathcal I_A}\le\varepsilon_A$. Then
\begin{equation}
\|\hat a_\omega-a^\ast_\lambda\|_{\mathcal I}\le(1+\kappa_A)\,\varepsilon_3+\kappa_A\,\varepsilon_A+n_1\varepsilon_1+n_2\varepsilon_2 .
\label{eq:quantitative}
\end{equation}
Conversely, if $\|\hat a_\omega-a^\ast_\lambda\|_{\mathcal I}\le\delta$ for every $\omega\in\Omega$, the four tolerances hold with $\varepsilon_1=\varepsilon_2=2\delta$ and $\varepsilon_3=\varepsilon_A=\delta$.
\end{theorem}
Therefore, a calibrated anchor, a shared structure, and short chains of correspondences and law changes together control the error on instances that are never anchored themselves.

\subsection{Counterexamples and Anchor Observability}
\begin{proposition}[Non-implication]\label{prop:nonimplication}
Within the one-degree-of-freedom family of Section~\ref{sec:twins} in the main paper, with law changes in (C2) that keep the inertia fixed and an explainer class that also contains scaled laws, there are predictors that
\begin{itemize}
\item[(a)] satisfy (C1) and (C3) but violate (C2) and (A): they predict with a fixed law $\bar\lambda\ne\lambda_0$ transported to each mechanism;
\item[(b)] satisfy (C2) and (C3) but violate (C1): they add to the true law a spurious gravity-like term $\delta h_m$ that differs between twins;
\item[(c)] satisfy (C1), (C2), and (A) but violate (C3): they use the true law on $\mathcal I_A$ and a scaled law $c\lambda$, $c\ne1$, on actuated interventions outside $\mathcal I_A$;
\item[(d)] satisfy (C1)--(C3) and the anchor equality on a passive set, yet are wrong because that set fails the identifying requirement in (A): they predict with $c\lambda$ everywhere.
\end{itemize}
\end{proposition}

\begin{proposition}[Anchor observability for linearly parameterized laws]\label{prop:kappa}
Suppose $a^\ast_\theta(x)=\Phi(x)\beta(\theta)$ with features $\Phi(x)\in\R^{d\times p}$ and an injective map $\theta\mapsto\beta$ onto an open set. Let $G_\iota=|\mathcal X_\iota|^{-1}\sum_{x\in\mathcal X_\iota}\Phi(x)\Tr M_{\rm ref}\Phi(x)$ and $G_A=\sum_{\iota\in\mathcal I_A}G_\iota$. If $\ker G_A\not\subseteq\ker G_\iota$ for some $\iota\in\mathcal I$, then $\kappa_A=\infty$; otherwise
\begin{equation}
\kappa_A\le\max_{\iota\in\mathcal I}\Big(|\mathcal I_A|\;\lambda_{\max}\big(G_A^{+1/2}\,G_\iota\,G_A^{+1/2}\big)\Big)^{1/2}.
\label{eq:kappa-bound}
\end{equation}
For the common law \eqref{eq:common} at fixed $\varepsilon$,
\[
\begin{aligned}
\Phi(q,\dot q,u)&=\big[\,u,\ -g_0\sin q,\ -\dot q,\ -\tanh(\dot q/\varepsilon)\,\big],\\
\beta&=\big(1/I,\ h/I,\ b/I,\ f/I\big);
\end{aligned}
\]
passive anchors leave the first coordinate unexcited, so $\kappa_A=\infty$ whenever actuated interventions are tested.
\end{proposition}

\begin{remark}\label{rem:characterization}
The proofs in Section~\ref{app:proofs} show what each relation transports, i.e., (C3) across interventions, (C2) across laws, and (C1) across mechanisms, and why the anchor must identify the scale (case (d), Proposition~\ref{prop:kappa}, and Theorem~\ref{thm:channels}(i)). Theorem~\ref{thm:quantitative} adds two design rules: errors accumulate linearly along the path from the anchor, so tested instances should lie few edges from $\omega_0$, for example in a star of twin pairs and law changes around it; and the anchor amplifies consistency and anchoring errors by $\kappa_A$, so anchors should contain actuated interventions that excite every direction the queries excite. The classical counterpart of Proposition~\ref{prop:kappa} is persistent excitation in identification~\cite{S:ljung1999system,S:swevers1997optimal}.
\end{remark}

\section{The Forward Half: From Law to Form}\label{sup:simulator}

In this section, we detail the forward half summarized in Section~\ref{sec:constructions} of the main paper. Algorithm~\ref{alg:simulator} lists both halves of FoLaSim in the order in which they run.

\begin{algorithm}[t]
\caption{FoLaSim, the form--law twin simulator}
\label{alg:simulator}
\footnotesize
\begin{algorithmic}[1]
\REQUIRE target laws $\{\lambda\}$; topology specifications $\{m\}$; densities $\mathcal C$; effort programs and controllers; cameras; for the inverse, a model with declared capabilities and an explainer class $\Theta$
\ENSURE certified twin corpus; capability profile with floors and coverage
\STATE \textbf{Forward}
\FOR{each law $\lambda$ and topology $m$}
  \STATE \textbf{synthesize}: solve $Y\pi=y$ \eqref{eq:matching}; realize $\pi$ as finite parts, or report $\lambda$ unrealizable on $m$
  \STATE \textbf{certify} body and engine terms against $\lambda$; \textbf{form twins}: scale densities by $c\in\mathcal C$
  \FOR{each density and program}
    \STATE simulate; gate closure, law residual, energy; render; record the effort publicly, the law privately
  \ENDFOR
\ENDFOR
\STATE audit law twins under shared efforts and form twins in free motion
\STATE \textbf{Inverse}
\STATE validate on designed faults; calibrate the readout on development laws and freeze it
\FOR{each instance $\omega$, evidence $e$, intervention $\iota$}
  \STATE \textbf{query} the model; receive the forecast
  \STATE \textbf{form space}: pose of every body, unprojected; closure; form verdict
  \STATE \textbf{state space}: coordinates $\hat s_t$, derivatives, effort $u_t$, floors
  \STATE \textbf{law space} (intact form): $\delta u$ \eqref{eq:gram} and $r_E$ \eqref{eq:energy}
\ENDFOR
\STATE \textbf{cross the twins}: reuse, adjustment \eqref{eq:change}, consistency, anchor; bootstrap over laws
\end{algorithmic}
\end{algorithm}

\subsection{Specifications}
A topology is a declarative specification. It declares the public coordinates (joints of the model), the law class, the working domain, the input map $T$ with motor efforts $\tau=Tu$, the ground parts, the moving bodies, the constraints, the actuators, and the cameras. Each moving body lists its parts, its tunable parts, and optionally an explicit embedding (origin, rotating offsets, slopes, and angle rates in the generalized angles), which the body certificate uses. Every moving body enters the model as a planar free body with three slides and a hinge about $-Y$; joints, gears, belts, screws, and closed loops are equality constraints (fixed joints, joint relations $q_i=r\,q_j$, and connections of two body points), and actuators are motors on a joint or on a fixed tendon. Therefore, one model writer serves every topology. Parts come from a library that carries the exact mass and inertia of each solid: boxes, cylinders, rings, involute gears, prisms, bars, helices, racks, internal ring gears, shaft supports, and rotor hardware. Parameters drawn per law, such as the payloads of P1 and P2, are declared as distributions with a seed.

\begin{table*}[t]
\centering
\caption{Tunable parts. Each kind enters the reduced dynamics linearly in its parameters and is realized as a finite solid, or reported unrealizable. Growth along the hinge axis moves neither the in-plane center of mass nor the axial inertia, which keeps ballasts and flywheels linear in their parameter.}
\label{tab:tunable}
\footnotesize
\begin{tabularx}{\textwidth}{@{}L{1.4cm}L{2.2cm}YL{4.1cm}L{3.3cm}@{}}
\toprule
Kind & Parameters & Shape family & Realization & Realizable when \\
\midrule
Ballast & mass $m$ & steel cylinder of fixed radius at a fixed in-plane position, growing along the hinge axis & length $m/(\rho\pi r^2)$ & $m>0$ \\
Payload & mass $m$ & box of fixed size and position & density $m/V$ & $500\le m/V\le19{,}300$\,kg$\cdot$m$^{-3}$ \\
Flywheel & axial inertia $J$ & balanced steel ring of fixed radii, growing along its axis & $m=2J/(r^2+r_i^2)$, length $m/(\rho\pi(r^2-r_i^2))$ & $J>0$ \\
Block & $m$, $h_z$, $J$ & steel block of fixed width on a mount, sliding along the link & $z=h_z/m$, $\ell^2=12\,(J-m(z^2+x_0^2)-I_{\rm mount})/(m-m_{\rm mount})-w^2$ & $m>m_{\rm mount}$, $\ell^2>0$ \\
\bottomrule
\end{tabularx}
\end{table*}

The condition $\ell^2>0$ on a block is the point-mass bound: its hinge inertia must exceed that of its mass concentrated at its center, the planar case of the semidefinite conditions on inertial parameters~\cite{S:wensing2018lmi}.

\subsection{Reduced Dynamics Read from the Engine}
Given public coordinates $s$, the dependent joints $q_d$ are solved from the equality residual $\phi(q)$ and its Jacobian $J=[J_s\;J_d]$, both taken from MuJoCo, by Newton steps $q_d\leftarrow q_d-J_d^{+}\phi$ with continuation in steps of 0.05\,rad until $\|\phi\|_\infty<10^{-13}$. The tangent basis is $N_s=I$, $N_d=-J_d^{+}J_s$; the pseudoinverse absorbs the redundant rows of planar closed loops. With $M$ from the engine's mass matrix, $b(q,\dot q)$ its bias forces, and $A$ its actuator moments,
\begin{equation}
\begin{aligned}
M_S&=N\Tr MN, & \nabla V_S&=N\Tr b(q,0),\\
h_S&=N\Tr\big(b(q,Nv)+M\,(\partial_vN)\,v\big), & B_S&=N\Tr A\Tr T,
\end{aligned}
\end{equation}
where $\partial_vN$ is the derivative of $N$ along $v$, taken by central differences with step $10^{-6}$.

\subsection{Matching and Solution}
The matching conditions \eqref{eq:matching} are imposed on the unique entries of $M_S$ and on $\nabla V_S$ at nine states (one degree of freedom) or a $5\times5$ grid (two) spanning 95\% of the working domain. Because the reduced terms are affine in $\pi$, the columns of $Y$ are exact differences of the engine's terms around the nominal design $\pi_0$, and one refinement step removes their round-off. The simulator returns the solution nearest the nominal design in relative terms,
\begin{equation}
\pi=\pi_0+W\,(YW)^{+}\,(y-Y\pi_0),
\end{equation}
with $W=\mathrm{diag}(\max(|\pi_0|,10^{-3}))$,
accepts it if the matching error is below $10^{-9}\max(1,\|y\|_\infty)$ and every part is realizable (Table~\ref{tab:tunable}), and otherwise reports the topology unrealizable with the failed condition. Table~\ref{tab:free} gives the rank of $YW$ for each topology.

\begin{table}[t]
\centering
\caption{Inertial directions that the law fixes and leaves free, the same for all 32 development laws.}
\label{tab:free}
\footnotesize
\setlength{\tabcolsep}{3pt}
\begin{tabularx}{\columnwidth}{@{}L{1.3cm}ccY@{}}
\toprule
Topology & Unknowns & Rank & Free directions \\
\midrule
A1, A3, B1--B3 & 2 & 2 & none \\
A2 & 3 & 2 & split between two identical ballasts \\
R2 & 6 & 4 & mass of link 1; mass of link 2, with the first moment and inertia of link 1 compensating \\
R1 & 6 & 3 & as R2, and the split of $A+4D$ \\
T2 & 8 & 6 & as R2; the two counterweights are fixed \\
\bottomrule
\end{tabularx}
\end{table}

For the coupled class below, $B=L_1h_2$ and $Q=g_0h_2$, where $h_2$ is the first moment of the second link. A serial arm therefore requires $B/Q=L_1/g_0$. The parallel sources meet this condition because their lower stage hangs from a coupler that translates with the upper link at radius $L_1$.

\subsection{Constant-Inertia Mechanisms}\label{sup:families}
\begin{figure*}[t]
\centering
\includegraphics[width=\textwidth]{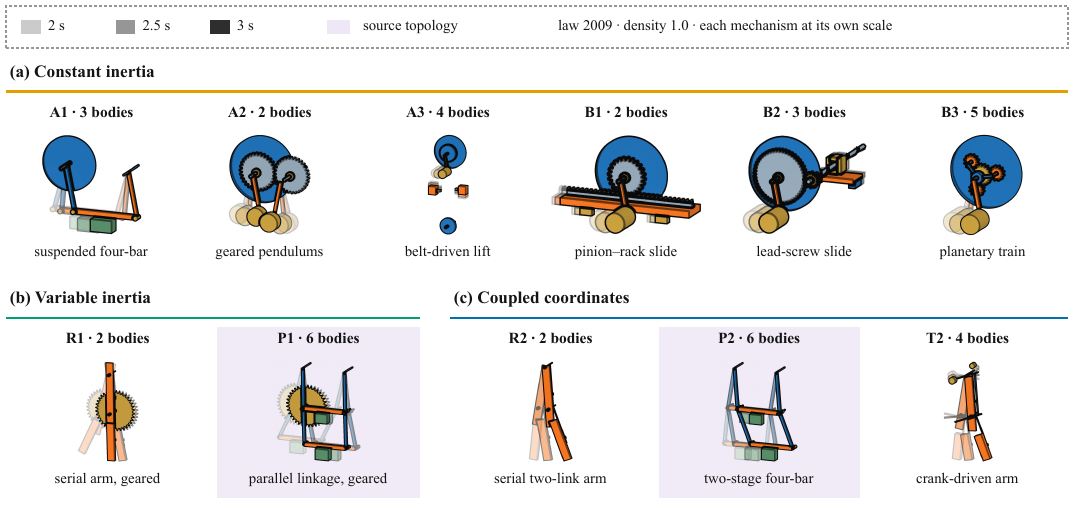}
\caption{The eleven built-in topologies, each at law 2009 of its class and density 1.0, ghosted at 2.0, 2.5, and 3.0\,s of the open-loop multisine and drawn to its own scale. The parallel sources P1 and P2 (shaded) carry the law read from their parts, which R1, R2, and T2 are synthesized to carry; the six constant-inertia topologies are synthesized from the constants of their common law.}
\label{fig:gallery}
\end{figure*}

The six constant-inertia families realize
\begin{equation}
I\ddot q+g_0h\sin q+b\dot q+f\tanh(\dot q/\varepsilon)=u,
\label{eq:common}
\end{equation}
where $I$ is inertia, $h$ the first moment, $g_0$ gravitational acceleration, $b$ viscous damping, and $f$ Coulomb friction smoothed over $\varepsilon$. The parameters $\lambda=(I,h,b,f,\varepsilon)$ are matched term by term with finite rigid bodies, and every realization has input power $u\dot q$. Table~\ref{tab:families} gives the six topologies and their public coordinate and input maps; Fig.~\ref{fig:gallery} shows all built-in mechanisms at one law. Their joints are ideal, rigid, bilateral, and free of backlash. All inertia and balancing forces come from realizable parts, including finite counterweights and flywheels, under the declared input. The construction thus changes the mechanical realization itself, beyond changes in rendering.

\begin{table*}[t]
\centering
\caption{Six mechanism families realizing the common law \eqref{eq:common}. A and B denote source and target constructions; the particular family and combination hold-outs are specified per experiment.}
\label{tab:families}
\footnotesize
\begin{tabularx}{\textwidth}{@{}L{0.6cm}L{5.2cm}Y@{}}
\toprule
& Mechanism & Public coordinate and input \\
\midrule
A1 & Suspended parallel four-bar & $q$: swing-arm angle from the downward vertical; $u$: torque on the left input shaft \\
A2 & Gear-coupled double pendulum (equal teeth) & arm angles $q$ and $-q$; $u$: torque on the first shaft \\
A3 & Timing-belt dual-slider balanced lift & slider displacements $+rq$ and $-rq$; $u$: torque on the driving pulley shaft \\
B1 & Eccentric-counterweight pinion--rack slide & $x-x_0=rq$; $u$: pinion-shaft torque \\
B2 & Eccentric input shaft with lead-screw slide & screw angle $nq$, $x-x_0=n\ell q$ ($\ell$ = lead$/2\pi$); $u$: input-shaft torque \\
B3 & Fixed-ring planetary train, offset-loaded carrier & $q$: carrier angle; sun angle $n_sq$; $u=n_s\tau_s$ \\
\bottomrule
\end{tabularx}
\end{table*}

\subsection{Variable-Inertia and Coupled Mechanisms}\label{sup:coupled}
P2 is a two-stage parallel four-bar with six moving bodies. Its law is read from those parts and matched by R2, a serial two-link mechanism with one sliding block per link. In absolute link angles $s=(\alpha,\beta)$, both have
\begin{equation}
\begin{aligned}
M_S&=\begin{bmatrix}A&B\cos(\alpha-\beta)\\B\cos(\alpha-\beta)&D\end{bmatrix},\\
V_S&=-P\cos\alpha-Q\cos\beta.
\end{aligned}
\label{eq:coupled-twins}
\end{equation}
The constants $A,B,D,P,Q$ fix four of R2's six block parameters and obey the realizability condition $B/Q=L_1/g_0$. Matching total masses, first moments, and effective axial inertias gives positive-inertia realizations of both assemblies. Physical motor efforts obey $\tau_1=u_\alpha+u_\beta$ and $\tau_2=u_\beta$, so $\tau_1\dot\alpha+\tau_2(\dot\beta-\dot\alpha)=u^\top\dot s$. The declared port map preserves work despite the relative-joint motor coordinates.

R1/P1 include the mass and inertia of the gear that imposes $\beta=2\alpha$ before moment matching. With $q=\alpha$, the reduced inertia and law are
\begin{align}
I(q)&=A+4D+4B\cos q,\\
I(q)\ddot q-2B\sin q\,\dot q^2&+P\sin q+2Q\sin(2q)+d(q,\dot q)=u.
\label{eq:variable-sup}
\end{align}

\subsection{T2: A Topology Given as a Specification}\label{sup:t2}
T2 has an upper link of 0.28\,m and a lower link of 0.25\,m, each carrying a block. A crank of radius 0.08\,m, coaxial with the shoulder and driven by the elbow motor, is pinned to a rod whose other end is pinned to an equal crank on the lower link; crank, rod, lower-link crank, and upper link form a parallelogram, so the elbow motor sets the absolute angle $\beta$ and the input map is the identity. The drive adds $\sin(\alpha-\beta)$ and $\cos\beta$ terms outside \eqref{eq:coupled-twins}. The two counterweights that cancel these terms are declared as ballasts on the rod and on the base crank; synthesis sets them to 0.381\,kg and 0.544\,kg for every law, since they cancel terms of the drive geometry, not of the law, and sets the blocks as on R2. The parallelogram is kept away from its singular configuration by a working domain of $\pm0.6$\,rad.

\subsection{Automatic Form Generation}\label{sup:forms}
\begin{figure*}[t]
\centering
\includegraphics[width=\textwidth]{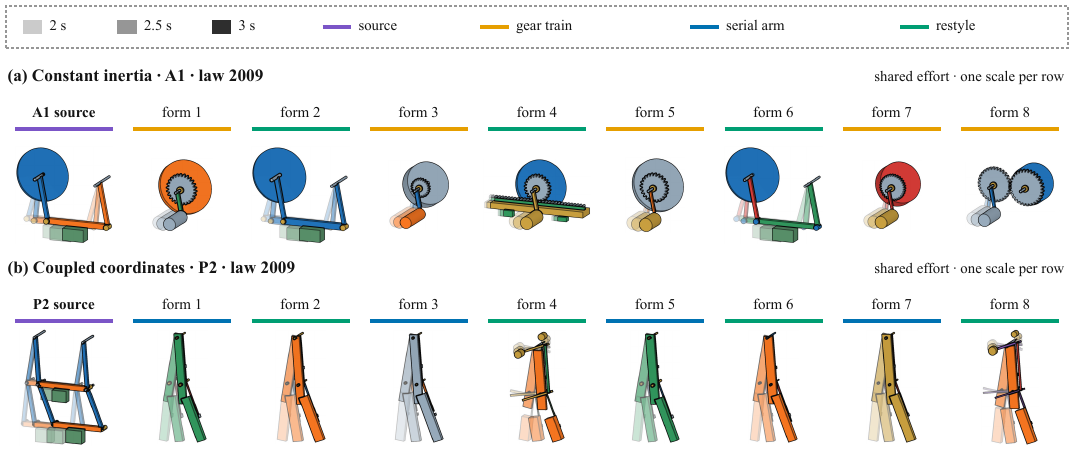}
\caption{All forms generated for law 2009: eight per class beside the source mechanism, A1 for constant inertia and P2 for the coupled class, ghosted at 2.0, 2.5, and 3.0\,s of one shared effort program and drawn to one scale per row; colored rules identify the family that drew each form. Each form is certified and moves with its source within the relation audit (Table~\ref{tab:forms-validation}).}
\label{fig:forms-all}
\end{figure*}

The form generator of Section~\ref{sec:twins} in the main paper samples a specification from a seed in one of three families (Table~\ref{tab:families-generated}). Fig.~\ref{fig:forms-all} shows every form generated for law 2009.

\begin{table}[t]
\centering
\caption{Form families of the generator. Every draw is synthesized for the target law and certified; unrealizable or uncertified draws are discarded.}
\label{tab:families-generated}
\footnotesize
\setlength{\tabcolsep}{3pt}
\begin{tabularx}{\columnwidth}{@{}L{1.45cm}L{1.35cm}Y@{}}
\toprule
Family & Class & What is drawn \\
\midrule
gear train & constant inertia & one to three meshed gears of one module, tooth counts, positions, eccentric arms with ballasts on gears turning with the input, balanced flywheels on any gear \\
serial arm & coupled & lower link length (0.18--0.34\,m), link cross-sections, sliding blocks and their mounts, materials; the upper link from the law \\
restyle & any & a built-in specification with tunable parts: colors and materials, cross-sections of structural parts, radii of ballasts and flywheels, block widths \\
\bottomrule
\end{tabularx}
\end{table}

Two constraints keep the draws inside the declared law class: eccentric ballasts sit only on gears that turn with the input at ratio $\pm1$, since on any other gear their gravity would act as $\sin(nq)$ rather than $\sin q$; and a serial arm takes its upper link from the law. Restyling changes no joint, anchor, or embedding, so the kinematics of the re-dressed mechanism is that of its template, while its finite parts, and with them the solved tunable parameters, differ. A kept form records its family, seed, and draw, which regenerate it exactly, and is written as an editable specification.

\subsection{Density Form Twins}\label{sec:scale}
The density form twin of Definition~\ref{def:formtwins} in the main paper isolates the scale of the law.

\begin{proposition}[Density form twins and the evidence ladder]\label{prop:scale}
For $c>0$ let $c\lambda=(cI,ch,cb,cf,\varepsilon)$. (i) With $u\equiv0$, $\lambda$ and $c\lambda$ produce identical motions from identical initial states. (ii) Under the same input $u(t)$, the motion under $c\lambda$ equals that under $\lambda$ with input $u/c$; the action response $\partial\ddot q/\partial u$ becomes $1/(cI)$. (iii) Under PD feedback $u=K_p(r-q)-K_d\dot q$, the closed loop of $c\lambda$ equals that of $\lambda$ with gains $K_p/c$ and $K_d/c$. (iv) If both track a common reference exactly, the required efforts differ by the factor $c$.
\end{proposition}

Table~\ref{tab:ladder} lists which observation channels distinguish the scale. Compliant control changes the sensitivity continuously and does not by itself guarantee identification. Scale ambiguity is classical, since a Lagrangian is determined at most up to a constant factor~\cite{S:douglas1941inverse}, and it has been discussed for port-Hamiltonian learning~\cite{S:phast2026}. FoLaSim uses it as a controlled instrument.

\begin{table}[t]
\centering
\caption{Evidence ladder for the density form twins of Proposition~\ref{prop:scale}.}
\label{tab:ladder}
\footnotesize
\begin{tabularx}{\columnwidth}{@{}L{2.6cm}Y@{}}
\toprule
Observation channel & Can $c$ be distinguished? \\
\midrule
Passive motion & No (Proposition~\ref{prop:scale}(i); Theorem~\ref{thm:channels}(i)) \\
Stiff servo, motion only & Barely: information decays as $K_p^{-2}$ (Corollary~\ref{cor:ladder}) \\
Compliant or impedance control & Depends on excitation, sampling, and noise \\
Open-loop torque & With informative samples: response scales as $1/c$ \\
Measured efforts & Informative if the reference excites the scale; nonzero high-gain limit (Corollary~\ref{cor:ladder}) \\
\bottomrule
\end{tabularx}
\end{table}

\subsection{Programs and Gates}\label{sup:programs-gates}
\begin{figure*}[t]
\centering
\includegraphics[width=\textwidth]{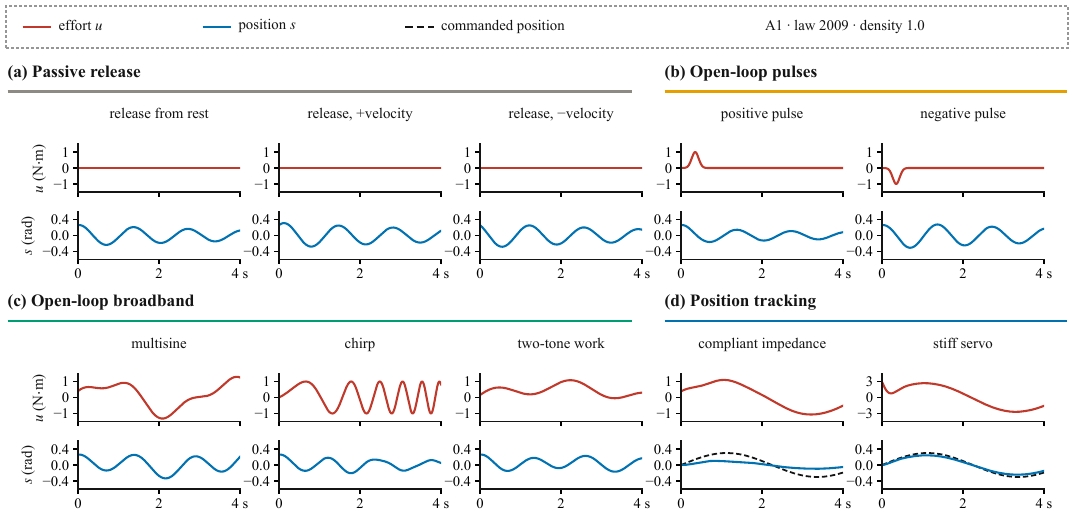}
\caption{The ten programs of A1 at law 2009 and density 1.0, grouped by the channel through which they show the law: the effort $u$ (top) and the public coordinate $s$ (bottom) over the 4\,s episode, with the commanded position (dashed) under tracking. Mechanisms with two degrees of freedom add pulses on the second coordinate.}
\label{fig:programs}
\end{figure*}

The programs expose the law through passive motion, open-loop response, or tracked effort (Table~\ref{tab:programs}; Fig.~\ref{fig:programs}). Each instance has ten programs, or twelve with the additional second-coordinate pulses for two degrees of freedom. MuJoCo integrates each 4\,s episode with the law's dissipation applied as generalized forces. Two calibrated cameras render $1024\times1024$ frames at 60\,Hz, each paired with shaft efforts and, under tracking, commanded positions.

The body certificate checks reduced inertia, velocity coupling, and gravity against the target law to $10^{-9}$ over a working-domain grid. The engine certificate independently checks the projected MuJoCo mass matrix and bias forces to $10^{-7}$. The episode gate requires closure error below 2\,mm, target-law residual below 0.08\,\Nm{} RMS (0.10 for two degrees of freedom), and energy closure within 0.08\,J. The relation audit requires agreement within 2\,mrad (3\,mrad for two degrees of freedom) for law twins under identical open-loop efforts and form twins released from rest.

\begin{table}[t]
\centering
\caption{Programs on the public coordinates, $u$ in \Nm{}, clipped to $\pm12$. Index $i$ runs over the coordinates; two-degree-of-freedom instances add pulses on the second coordinate.}
\label{tab:programs}
\footnotesize
\setlength{\tabcolsep}{3pt}
\begin{tabularx}{\columnwidth}{@{}L{1.6cm}Y@{}}
\toprule
Program & Effort \\
\midrule
release & $u=0$ from rest and from $\dot s=\pm0.8$ ($\pm0.6$) \\
pulses & $\pm\exp(-((t-0.35)/0.12)^2)$ \\
multisine & $\sin2.1t+0.4\cos4.7t$; $\sin(2.1+0.7i)t+0.3\cos(4.2+0.5i)t$ \\
chirp & $\sin(1.2t+1.8t^2)$; $0.35\sin((1.1+0.3i)t+(1.5+0.2i)t^2)$ \\
work & $0.7\sin0.8t+0.4\sin(3.3t+0.4)$; $0.8\sin(0.8+0.3i)t+0.3\sin(3.1t+0.3i)$ \\
impedance & PD tracking of $0.30\sin1.4t$ ($0.20\sin(1.4t+0.3i)$), gains $(5,0.9)$ ($(4,0.8)$) \\
servo & as impedance, gains $(45,7)$ ($(35,5)$) \\
\bottomrule
\end{tabularx}
\end{table}

\subsection{Regeneration of the Corpus}\label{sup:corpus}
The development corpus realizes 32 laws per dynamics class at densities 0.7, 1.0, and 1.4, comprising 960 assets, 9,984 physical episodes, and 19,968 two-view videos. Damping and friction scale with density. Additional same-form controls alter only one body mass, a load, friction, or damping. Table~\ref{tab:regeneration} checks the corpus by regenerating its assets from specifications.

\begin{table*}[t]
\centering
\caption{Regeneration of the corpus by the general synthesizer. Inventory differences are relative to the corpus; law differences are the largest entrywise difference of $M_S$ and $h_S$ on random states.}
\label{tab:regeneration}
\footnotesize
\begin{tabularx}{\textwidth}{@{}L{4.2cm}L{4.6cm}Y@{}}
\toprule
Check & Scope & Result \\
\midrule
One-degree-of-freedom synthesis & 32 laws $\times$ A1--B3 (192) & inventories equal to the corpus to $1.9\times10^{-10}$; body and engine certificates $\le1.1\times10^{-11}$ and $\le2.3\times10^{-10}$ \\
Parallel sources & 32 laws $\times$ P1, P2 & inventories identical to the corpus; certificates $\le3.7\times10^{-12}$ and $\le2.4\times10^{-9}$ \\
Serial twins from the parallel law & 32 laws $\times$ R1, R2 & law equal to the corpus law to $2.2\times10^{-10}$; other masses along the free directions \\
T2 & 32 laws of P2 & all realizable; certificates $\le9.8\times10^{-12}$ and $\le2.0\times10^{-8}$ \\
Law-twin audit & 4 laws, four open-loop programs & one degree of freedom $5.3\times10^{-6}$\,rad; P2, R2, T2 $4.3\times10^{-4}$\,rad \\
Form-twin audit & 4 laws, densities 0.7 and 1.4, release & $2.5\times10^{-12}$\,rad \\
Episode gate & one law, six families, three densities, ten programs & 180 of 180; closure $\le1.3\times10^{-6}$\,m, law residual $\le2.7\times10^{-3}$\,\Nm{} RMS, energy ledger $\le3.3\times10^{-3}$\,J \\
Unrealizable laws & T2 with $L_1=0.30$\,m; B1 with $h=0.01$; B1 with $I=0.05$ & reported: matching error $1.2\times10^{-2}$; negative ballast; negative flywheel inertia \\
\bottomrule
\end{tabularx}
\end{table*}

\subsection{MJX Backend}\label{sup:backend}
FoLaSim uses MuJoCo 4.0.0, with backward compatibility to 3.2. MJX, its JAX implementation, loads all eleven topologies unchanged except for the unsupported engine energy flag, used only by the C path's energy ledger. In double precision (JAX 0.12.0.1), mass matrices and bias forces equal those of the C engine to a relative $2\times10^{-17}$ at matched states of all topologies. Batched reduced terms on 36 R2 states equal the certified law to $2.3\times10^{-12}$, and their body-mass derivatives equal finite differences to $8\times10^{-10}$. Section~\ref{sup:e1-forward} reports rollout parity, throughput, and gradients.

\subsection{Software}\label{sup:software}
FoLaSim is released as a Python package, \texttt{folasim}, on MuJoCo (Apache License 2.0), with documentation, examples, a command line, and tests that run in continuous integration; the paper's laws are reproduced from their ids bit for bit. It is organized as a simulator rather than as experiment scripts: every step is a function of a specification and a law.
\begin{itemize}
\item \emph{Mechanisms}: a registry, \texttt{make(name, law\_id, density)}, of the eleven built-in specifications; \texttt{synthesize(law, spec)} on any specification, or a report that it cannot carry the law; \texttt{form\_twin(c)}; the certificates and twin audits; and the generator of Section~\ref{sup:forms}.
\item \emph{Simulation}: effort programs and controllers, episodes with energy ledger and gate, a step-by-step environment under torque, impedance, or servo control (with a Gymnasium view), rendering from calibrated cameras whose pinhole calibration is read from the engine's own scene camera, corpus generation with the public/private split of Section~\ref{sec:certify} in the main paper, and batched, differentiable rollouts on MJX.
\item \emph{Reading}: the law-space reader with its floors, the four tests, the image readout and form verdict of Section~\ref{sup:readout}, and the reports of Section~\ref{sup:reports}.
\item \emph{Robots}: open descriptions of the SO-101 arm, the LEAP hand, and the Unitree Go2, with their laws read from the model, equimomental law twins, density form twins, and effort programs (Section~\ref{sup:robots-open}); three contact tasks with YCB objects, their twins, controllers, and the reading with contact forces as unknowns, down to the law link by link (Section~\ref{sup:contact}); any MJCF or URDF description mounted as a contact scene, sixteen robots of MuJoCo Menagerie by name (Section~\ref{sup:zoo}); a robot's law read from its motion and effort, with only its kinematics trusted, and the law twin built from it (Section~\ref{sup:law}); and for every scene a step-by-step environment, episode files, rendering, and command-line tools.
\item \emph{Models}: a black-box protocol in which a model declares what it observes (states or images), whether it accepts evidence, and its action port (effort or position), and returns states or frames; the two-by-two test; and reference predictors with known faults of law and of form, rendered as video when needed, with which the reader is checked on any mechanism.
\end{itemize}
The command line exposes the same steps: listing and building mechanisms, generating forms, describing a mechanism, inspecting a forecast, simulating and rendering, generating corpora, and checking the reader against designed faults.

\section{The Inverse Half: From Form to Law}\label{sup:measurement}

In this section, we detail the inverse half summarized in Section~\ref{sec:measurement} of the main paper, in the order of the three spaces, followed by the crossing of the twins and the conditions that make the measurement reliable.

\subsection{Form Space: Readout and Form Verdict}\label{sup:readout}
\paragraph*{Readout} The readout fits the planar translation and orientation of every moving body, with depth fixed by the planar scene; its color gain and bias are the coordinate along the fibers of Section~\ref{sec:observation-layer}. Fits are initialized from the previous pose or a past-velocity estimate, at several scales and from several starts, and selected by image-fit validity and the worst-body photometric residual, without joint projection or reference states, so a body displaced by 25\,mm is read as displaced by 25\,mm. A frame is read reliably when the optimizer succeeds, every body shows at least eight pixels, and the local photometric Jacobian has condition below $10^4$. Its median angle error of 0.0024\,rad is measured on codec reconstructions of 24 development references, with a maximum of 0.0031\,rad. Tracking stride and Jacobian scaling are chosen per mechanism and probe on development laws 24 and 25, by the largest normalized pose error on reference videos and their codec reconstructions, and frozen before any confirmation forecast is scored.

\paragraph*{Form verdict} For every joint (ground pivot, pin between bodies, or coupling) and frame, the verdict records whether both bodies are read reliably, the joint gap in the native model, and the pixel cost, computed as the increase of the photometric residual around the joint's bodies when the fitted poses are projected onto the constraint manifold. A joint is broken when its gap and pixel cost both exceed their thresholds for three consecutive frames. It is intact when at least 80\% of frames are reliably read and keep the gap below threshold, and undecidable otherwise. A forecast is broken if any of its joints is. Thresholds are calibrated on intact development references (Section~\ref{sup:e2-form}).

\subsection{State Space: Derivatives, Effort, and Floors}\label{sup:state}
\paragraph*{Discrete residual} Since WAMs predict frames, velocities $\hat v_k$ are estimated from the position readout by a linear differentiator; velocities that a model reports separately are used only as a consistency check. In a smooth bilateral phase, we retain the measured body coordinates and evaluate the residual at the midpoint of each frame interval, $\bar q_k=\tfrac12(\hat q_k+\hat q_{k+1})$ and $\bar v_k=\tfrac12(\hat v_k+\hat v_{k+1})$:
\begin{equation}
\begin{aligned}
\bar\rho_k&=\bar N\Tr\big(\bar M\,(\hat v_{k+1}-\hat v_k)/\Delta t+\bar c-\bar F\big),\\
r_k&=\big(\bar\rho_k\Tr(\bar N\Tr\bar M\bar N)^{-1}\bar\rho_k\big)^{1/2},
\end{aligned}
\label{eq:discrete}
\end{equation}
where bars denote evaluation at $(\bar q_k,\bar v_k)$ with the input $u_{k+1/2}$. This midpoint form evaluates \eqref{eq:residual} with a central difference of velocity, in the same units; for samples of a $C^4$ trajectory with exact velocities or a second-order centered velocity stencil, $r_k=r(t_{k+1/2})+O(\Delta t^2)$. The corresponding reduced-coordinate discretization evaluates $M_S\Delta\dot s/\Delta t+C_S\bar{\dot s}+\nabla V_S+d_S-B_Su$ at $\bar s=(s_k+s_{k+1})/2$; Proposition~\ref{prop:discrete} concerns this reduced form. Ambient midpoints of different embeddings need not correspond, so the complete-body implementation is checked by reference replay and step refinement.

\paragraph*{Floors} Derivatives come from one nine-frame, degree-four local polynomial differentiator. The floors propagate four development error traces per target, i.e., the pose errors of the readout on reference videos and on their codec reconstructions, on the clock of the forecast. Exact reference positions processed at the same clock separate the readout error from the differentiation error.

\subsection{Law Space: Residuals of a Single Forecast}\label{sup:single}
\paragraph*{Admissibility} Closure, velocity, and normal-acceleration residuals are measured on the unprojected trajectories of all bodies; law comparisons use the declared reduced coordinates. Exact-state and image readouts are reported separately.

\paragraph*{Causation} The mass-weighted residual \eqref{eq:discrete} has units of force divided by the square root of mass (or torque divided by the square root of inertia); the effort correction \eqref{eq:gram} converts it to the units of each port, \Nm{} for rotary ports. It requires that only bilateral constraints be active and $B_S$ be invertible, and in full coordinates the response is $\Lambda=B\Tr N(N\Tr MN)^{-1}N\Tr B$.

\paragraph*{Energy and response} We evaluate the power residual \eqref{eq:energy}. Symmetric effort probes $\pm\delta u$ at matched states give an estimate $\hat\Lambda$ from collocated accelerations; comparisons use physical generalized efforts and collocated outputs, never Jacobians of normalized action tokens.

The P2 example in main-paper Fig.~\ref{fig:overview} illustrates the distinction between form and law failures at law 2009, density 1.4, under servo commands. In panel (a), a lower link of the LingBot-VA forecast leaves its pins by 21 and 24\,mm. Panels (b,c) instead use the designed time warp $s(0.8t)$, rendered with an intact assembly. Derivatives use the fixed nine-readout filter. At 2.5\,s, the effort required by this forecast and the applied motor effort turn opposite ways. Over the forecast, actuator work is 1.96\,J while stored energy increases by 0.07\,J. The two ports leave 5.9 and 2.9\,\Nm{} unexplained, or 48 and 33 times their respective floors; the exact forecast remains within its floors.

\subsection{Relational Measures}\label{sup:relational}
\paragraph*{(R1) Transfer error} For certified twins $(A,B)$, matched initial states ($s_B=\psi(s_A)$, $\dot s_B=D\psi\,\dot s_A$), and matched inputs ($u_B=Ku_A$), we average over prediction steps $k$
\begin{equation}
E_{\rm tr}=\big\langle\,\|\mathcal P_{p_k\to z_k}D\psi(\hat s_{A,k})\dot{\hat s}_{A,k}-\dot{\hat s}_{B,k}\|_{M_B(z_k)}\big\rangle_k,
\label{eq:transfer}
\end{equation}
where $p_k=\psi(\hat s_{A,k})$, $z_k=\hat s_{B,k}$, and $\mathcal P$ is parallel transport along the unique minimizing geodesic in a declared convex normal neighborhood. Positions are compared using the distance induced by $g_B$. At matched query states $p_k=z_k$, transport is the identity, as in our state-query confirmation. Implemented trajectory comparisons use the declared public charts; direct velocity subtraction at distinct states is a chart-dependent diagnostic and does not carry the intrinsic invariance guarantee of parallel transport. The transfer error is reported with the physical residuals of both predictions, because agreement between twins does not imply that either prediction is correct.

\paragraph*{(R2) Change error} Besides the change error \eqref{eq:change}, we report how the error on the true query improves with correct rather than absent evidence, as a function of the evidence budget.

\paragraph*{(R3) Shared-explanation error} For an instance with support interventions $\mathcal I_S$ and held-out interventions $\mathcal I_Q$, let $R_\iota(\theta)=\sum_k r_k(\theta)^2$ be the residual of the predicted trajectory under a candidate law $\theta\in\Theta$. Following the comparison of individual and shared fits in OneWorld~\cite{S:oneworld2026} and the anchoring of IMPLY~\cite{S:imply2026}, we report the individual-fit error $\sum_\iota\min_\theta R_\iota(\theta)$, the shared fit $\hat\theta_S=\arg\min_\theta\sum_{\iota\in\mathcal I_S}R_\iota(\theta)$ with its support error, its held-out error $\sum_{\iota\in\mathcal I_Q}R_\iota(\hat\theta_S)$, and the discrepancy between $\hat\theta_S$ and the true law. The class $\Theta$ is fixed in advance with bounded capacity, and its replay error on ground truth is reported, so that a flexible explainer cannot render any prediction consistent. In \emph{cross-relation hold-out}, $\mathcal I_S$ consists of impulse responses and $\mathcal I_Q$ of releases, velocity reversals, work programs, other speeds and configurations, and, through $\psi$, another mechanism.

\paragraph*{(R4) Reliability} We record the readout error under reference replay, the coverage, and the variance across repeated samples. Thresholds are set on an independent calibration set, and statistics are grouped by law instance and mechanism pair rather than by frame.

\paragraph*{(R5) Physical correctness} These are the single-prediction measures of Section~\ref{sec:core} in the main paper.

We report a capability profile rather than one aggregate score. For the compositional hold-out (Section~\ref{sec:crossing} in the main paper), the same measures compare cross-mechanism support, same-mechanism support, and no support, separating adaptation within a mechanism, transfer of an existing law, and transfer of a newly adapted change.

\subsection{Reading the Mass Matrix of a Model from Its Response}\label{sec:response-geometry-sup}
The local consistency test of Section~\ref{sec:relational} in the main paper asks whether the mass matrix implied by a model's response to effort predicts the velocity coupling, conservative force, and dissipated power of the same model. Hold the evidence fixed and use a known, fully actuated port, normalized to $B_S=I$. Write the predictor as $\hat a(s,v,u)$ and its force response as $A(s,v)=\partial_u\hat a(s,v,0)$. Input affinity, velocity independence of $A$, reciprocity, and positive definiteness come first. When the symmetric part of $A(s,0)$ is positive definite and well conditioned, the model's mass matrix is $\widehat M(s)=[\operatorname{sym}A(s,0)]^{-1}$, and the antisymmetric part is kept as a failure signal. With $a_0(s)=\hat a(s,0,0)$, the model's free motion separates into
\begin{align}
\widehat g(s)&=-\widehat M(s)a_0(s),\nonumber\\
\widehat c_{\rm even}(s,v)&=-\widehat M(s)\left[\frac{\hat a(s,v,0)+\hat a(s,-v,0)}{2}-a_0(s)\right],\nonumber\\
\widehat d_{\rm odd}(s,v)&=-\frac{\widehat M(s)}{2}\left[\hat a(s,v,0)-\hat a(s,-v,0)\right],
\label{eq:response-decomposition}
\end{align}
its conservative force, velocity-quadratic force, and velocity-odd dissipation. The Christoffel symbols of its own mass matrix predict the second,
\begin{equation}
c_i[\widehat M](s,v)=\sum_{j,k}\left(\partial_k\widehat M_{ij}-\tfrac12\partial_i\widehat M_{jk}\right)v_jv_k.
\label{eq:metric-christoffel}
\end{equation}
We compare $\widehat c_{\rm even}$ with $c[\widehat M]$, measure the curl $\partial_j\widehat g_i-\partial_i\widehat g_j$, and check $v^\top\widehat d_{\rm odd}\ge0$, with no fitting between the response and these tests.

\begin{proposition}[Reconstruction from response and free motion]\label{prop:response-structure}
Let a smooth acceleration field on a simply connected configuration domain be affine in $u$, with symmetric positive-definite response $A(s)$ independent of $v$. Define $\widehat M$, $\widehat g$, $\widehat c_{\rm even}$, and $\widehat d_{\rm odd}$ as above. The field is generated by a natural mechanical law with this mass matrix, a potential, and smooth velocity-odd dissipation if and only if
\begin{equation}
\widehat c_{\rm even}=c[\widehat M],\qquad
\partial_j\widehat g_i=\partial_i\widehat g_j,\qquad
v^\top\widehat d_{\rm odd}\ge0
\label{eq:structure-conditions}
\end{equation}
throughout the declared state domain. The potential is determined up to a constant. This characterizes the stated class; it excludes additional velocity-even nonconservative forces and unmodeled ports.
\end{proposition}

Signed effort differences estimate $A$ and centered coordinate differences estimate its derivatives, so the identities are tested at sampled states; invalid mass-matrix candidates remain in the coverage record. Since a structure can be valid but wrong, its distance from the true law is always reported with it. Matrix inversion and differentiation amplify numerical error (Section~\ref{sec:conditioning}); the test is therefore repeated at two coordinate steps and in double precision before a structural failure is interpreted.

\subsection{Numerical Conditioning of Response Inversion}\label{sec:conditioning}
Numerical conditioning is therefore part of this test. If a symmetric response has least eigenvalue $m_A>0$ and its perturbation obeys $\|\Delta A\|_2\le\eta<m_A$, then
\begin{equation}
\|(A+\Delta A)^{-1}-A^{-1}\|_2\le\frac{\eta}{m_A(m_A-\eta)}.
\label{eq:inverse-sensitivity}
\end{equation}
Coordinate differentiation can amplify this error further by the inverse step size. We therefore compare two coordinate steps and FP32/FP64 evaluation of the same weights before attributing a Christoffel mismatch to the learned law. This check is especially important for models whose architecture already imposes Lagrangian structure (Section~\ref{sec:e3} in the main paper).

\subsection{Invariance under Correspondence}

\begin{proposition}[Invariance of the measures]\label{prop:invariance}
Under Definition~\ref{def:equivalence} in the main paper: (i) the causal residual is invariant, $r_A[\hat s,u]=r_B[\psi\circ\hat s,Ku]$, for every trajectory $\hat s$ and input $u$; (ii) $\Lambda_A=K\Tr\Lambda_BK$; (iii) the power residual \eqref{eq:energy} is invariant; and (iv) admissibility residuals and reactions are not invariant in general.
\end{proposition}

\begin{proposition}[Discrete invariance]\label{prop:discrete}
Consider the reduced-coordinate midpoint residual described after \eqref{eq:discrete}, with uniformly positive-definite metrics and smooth bounded law coefficients. Positions on $B$ are $\psi(s_k)$ and inputs are $Ku_k$. (i) For affine $\psi$ and constant $K$, the residuals agree exactly when the same linear, constant-annihilating velocity stencil is used on both sequences. (ii) For samples of a common $C^4$ curve, a $C^4$ map $\psi$, and exact nodal velocities or centered second-order velocity stencils, $|r_k^A-r_k^B|\le C\Delta t^2$ away from boundary stencils.
\end{proposition}

The smoothness and stencil assumptions matter: bounded discrete accelerations alone do not establish second-order covariance. The native complete-body residual uses the independently embedded coordinates and is therefore compared at its measured discretization error, even when the reduced public map is linear. Section~\ref{app:numerics} checks the normalized reduced residual under a nonlinear map.

\subsection{Decidable Queries}\label{sec:decidable}
For evidence $\mathcal E$, a set of interventions with observed fields $a_{\rm obs}$, and a tolerance $\eta$, let
\begin{equation}
\Theta_\eta(\mathcal E)=\{\theta\in\Theta:\ \|a^\ast_\theta-a_{\rm obs}\|_{\mathcal E}\le\eta\}
\label{eq:consistent}
\end{equation}
be the laws consistent with the evidence, and define the unresolved width of a query set $\mathcal Q$ as $w_{\mathcal Q}(\mathcal E,\eta)=\sup_{\theta,\theta'\in\Theta_\eta(\mathcal E)}\|a^\ast_\theta-a^\ast_{\theta'}\|_{\mathcal Q}$.

\begin{proposition}[Decidable queries]\label{prop:decidable}
(i) Every predictor that receives only $\mathcal E$ has worst-case query error at least $w_{\mathcal Q}/2$ over $\Theta_\eta(\mathcal E)$, with worst case understood as a supremum. (ii) Under mechanical equivalence, $(\psi,K)_\ast$ maps the consistent set of evidence on $A$ onto the consistent set of the transported evidence on $B$, and the widths of corresponding query sets agree. (iii) For linearly parameterized laws, if $\ker G_{\mathcal E}\subseteq\ker G_\iota$ for every query intervention $\iota\in\mathcal Q$, then $w_{\mathcal Q}\le2\eta\,\kappa(\mathcal E,\mathcal Q)$, where $\kappa(\mathcal E,\mathcal Q)$ is the right-hand side of \eqref{eq:kappa-bound} with $\mathcal I_A$ replaced by $\mathcal E$ and $\mathcal I$ by $\mathcal Q$. An unexcited query-relevant direction instead leaves positive width whenever the consistent set contains a nontrivial segment in that direction. An unbounded feasible ray of this kind is sufficient for infinite width; nonidentification alone is not.
\end{proposition}

Proposition~\ref{prop:decidable} separates evidence sufficiency from model use of that evidence, in the spirit of the certificate-gated protocol of PokeWorld, which first checks that a physical parameter is recoverable from raw observations before attributing its absence from a latent world model to the training objective~\cite{S:pokeworld2026}. A large width prevents interpreting an error as failure to identify the law. Truth-relative errors for no-history and wrong-history controls remain useful measures of prior prediction and sensitivity, as in E3, but are not identification verdicts. A passive history cannot decide actuated queries (Proposition~\ref{prop:kappa}); by (ii), evidence gathered on one mechanism decides the corresponding queries on its twin. Two candidate laws are distinguishable at a stated tolerance when their query fields differ by more than the sum of the tolerances.

\subsection{Information Conditions and Probes}
Table~\ref{tab:info} lists the information conditions and Table~\ref{tab:probes} the probes, organized by the physical relation each constrains. Probes must cover the blind directions of Table~\ref{tab:invariances} and span control channels (open loop, impedance, stiff servo); otherwise the experiment covers only the excitation and feedback conditions already observed. All branches start from matched physical states, and a probe that a model's native interface cannot execute is recorded as not measurable rather than emulated through an invented interface. Static-appearance and no-history controls assess the predictive prior under unresolved physical uncertainty.

\begin{table}[t]
\centering
\caption{Information conditions.}
\label{tab:info}
\footnotesize
\begin{tabularx}{\columnwidth}{@{}L{1.75cm}YY@{}}
\toprule
Condition & The model receives & Supported conclusion \\
\midrule
Public parameters & Initial state, action interface, explicit physical parameters & Whether it can roll out a given law (diagnostic only) \\
Same-mechanism history & A finite interaction history of the instance & Whether it adapts to hidden physical change \\
Cross-mechanism support & History of $A$, initial state of $B$, public map & Whether a law obtained on $A$ is reused on $B$ \\
\bottomrule
\end{tabularx}
\end{table}

\begin{table}[t]
\centering
\caption{Probes organized by the physical relation they constrain.}
\label{tab:probes}
\footnotesize
\begin{tabularx}{\columnwidth}{@{}L{2.2cm}YY@{}}
\toprule
Probe & Main evidence & Held-out queries \\
\midrule
Signed short torque pulses at matched states & Local action response, effective inertia & Unseen amplitudes, durations, combinations \\
Releases or slow motion at several configurations & Gravity and conservative terms & Unseen configurations, the other mechanism \\
Opposite velocities at matched configurations & Dissipation and velocity-dependent terms & Other speeds, stopping \\
Finite work segments & Input work versus storage and dissipation & Action programs not used in fitting \\
Orthogonal inputs (multi-DOF) & Reciprocity, rank, and coupling of $\Lambda$ & Joint inputs, responses after added constraints \\
\bottomrule
\end{tabularx}
\end{table}

\subsection{Two-by-Two Test}\label{sup:twobytwo}
Table~\ref{tab:twobytwo} names the four cells of the test of Section~\ref{sec:crossing} in the main paper, and a fifth condition without evidence measures the prior of the model.

\begin{table}[t]
\centering
\caption{The two-by-two test for a query on mechanism $B$ under law $\lambda_2$. The rows vary the form, the columns the law; each cell names the evidence and what a model that has learned the law must do with it.}
\label{tab:twobytwo}
\footnotesize
\setlength{\tabcolsep}{3pt}
\begin{tabularx}{\columnwidth}{@{}L{1.55cm}YY@{}}
\toprule
Evidence & Same law $\lambda_2$ & Other law $\lambda_1$ \\
\midrule
Same form $B$ & own evidence: use it & same form, different law: must not be misled \\
Other form $A$ & same law, different form: must reuse it & both differ: control \\
\bottomrule
\end{tabularx}
\end{table}

Weights are frozen during the test, and the public correspondence reveals no hidden parameter. The evidence is designed to decide the queries: by Proposition~\ref{prop:decidable}(ii), evidence on $(A,\lambda_2)$ decides the queries on $(B,\lambda_2)$ exactly when it decides the corresponding queries on $A$, which requires actuated interventions (Proposition~\ref{prop:kappa}); all evidence clips therefore contain actuated segments.

\subsection{Reports of Form and Law}\label{sup:reports}
A report summarizes the readings of the three spaces as findings. For a mechanism, it lists the bodies, parts, masses, joints, and ports, and the law, read from the parts when the specification has none to tune. For a forecast it proceeds as follows.

\paragraph*{Form} With full configurations, the largest gap of every equality constraint; with frames, the readout and verdict of Section~\ref{sup:readout}; with public coordinates alone, the assembly is reconstructed and holds by construction.

\paragraph*{Implied law} The implied law is the least-squares fit of the generalized effort on the constants of the declared class (inertia, coupling, gravity, damping, and friction at a fixed width), with standard errors from the residual; the law implied by the true motion on the same clock, with which it is compared, equals the true law at fine clocks.

\paragraph*{Findings} The findings use fixed thresholds. One law explains a forecast when the fitted law explains at least 80\% of the variance of its effort; that law is the true one uniformly scaled when every constant's ratio lies within 3\% or two standard errors of their median. A time warp is reported when the true motion played at 0.5 to 1.5 times its speed fits the forecast more than three times better than the true motion itself, and the amplitude when it lies outside half to twice the truth.

\subsection{Reliability Conditions}\label{sup:reliability}
Five conditions make the measurement reliable: (1) the correspondence comes from construction; (2) the inverse half passes its validation on designed faults (Section~\ref{sec:metaval} in the main paper) before any model is scored; (3) interventions cover blind directions and span channels; (4) adjustment is tested with a declared evidence budget, that is, matched input--motion prefixes and frozen model parameters; and (5) the measured law, plant or closed loop, is declared (Section~\ref{sec:interfaces} in the main paper). In addition, the error of the measurement chain itself must be bounded, as stated below.

\begin{proposition}[Measurement-chain sensitivity]\label{prop:chain}
(a) \emph{Pose bias.} If the readout adds a slowly varying bias $\beta(t)$ to the reduced coordinates, the reduced residual changes by $\mathcal K\beta+\mathcal D\dot\beta+M_S\ddot\beta+O(\|(\beta,\dot\beta,\ddot\beta)\|^2)$ with derivatives evaluated at $(s,\dot s,\ddot s,u)$, $\mathcal K=\partial_s\rho$, and $\mathcal D=\partial_{\dot s}\rho$; a constant bias therefore enters through configuration-dependent terms such as gravity stiffness. (b) \emph{Effort estimation.} If the input is estimated as $\hat u=u+e_u$, then $|\hat r-r|\le\|e_u\|_\Lambda=(e_u\Tr\Lambda e_u)^{1/2}$; errors in $\ker\Lambda$, such as co-contraction, do not affect the verdict. (c) \emph{Structure estimation.} If the tangent projector is estimated with error $\varepsilon_T=\|\hat P_T-P_T\|_g$, then $|\hat r-r|\le\varepsilon_T\|D_t\dot{\hat q}-M^{-1}F\|_g$; for a physically correct motion this is reaction leakage, at most $\varepsilon_T\|M^{-1}J\Tr\nu\|_g$.
\end{proposition}

Pose bias matters when evaluating gravity-loaded mechanisms: the relevant first-order coefficient is the gravity stiffness $\partial_s\nabla V_S$, not the magnitude of gravity torque alone. We measure this sensitivity from the physical anchor and reference replay. For recorded robots, proprioceptive state and calibrated payload information provide the preferred reference; a nominal-model mismatch is not attributed directly to a WAM.

\section{Experimental Protocols and Results}\label{sup:experiments}
In this section, we give the protocols and the complete results of the experiments in Section~\ref{sec:experiments} of the main paper.

\subsection{Corpus, Models, and Statistics}\label{sup:protocol}
\paragraph*{Corpus} The development inventory is given in Section~\ref{sup:corpus}. Its two-view videos are observation variants of physical episodes, not additional trials. Laws 1000--1015 form the pilot set used to establish protocols. Confirmation laws 2000--2015 supply 480 assets and 4,992 episodes. The changed-density target combinations of B1, P1, and P2 are excluded from training and checkpoint selection. State, RGB, and LingBot-VA checkpoints, readout settings, and protocols were frozen before confirmation; the other video models use the same readout settings. Neither the pilot nor confirmation set selects checkpoints or thresholds.

\paragraph*{Models} The 54 state configurations comprise two sets of 27 across three dynamics classes and three seeds: the history-free MLP, context GRU, and mechanical-context model, and the published-architecture controls. RGB context models and no-context controls supply 30 checkpoints across five seeds (Section~\ref{sup:e3-rgb}). The four native video models use released interfaces, with BWM based on Wan2.2-TI2V-5B and Ctrl-World on Stable Video Diffusion (Sections~\ref{sup:e3-lingbot} and~\ref{sup:e3-models}). Their interfaces determine which relations are measurable, and each model class is evaluated separately.

\paragraph*{Statistics} Contrasts use identical query states and efforts. Probes, densities, and seeds are averaged within each law before bootstrapping laws with 4,000 resamples. State and RGB evaluations use sixteen confirmation laws, and video evaluations use six; per-seed contrasts additionally check effect direction. Frames and views do not count as independent law instances. Coverage, rejected trajectories, and numerical failures accompany the scores. Robot experiments use twenty training, four validation, and twelve test laws per family with three seeds (Section~\ref{sup:robots-open}). Table~\ref{tab:experiment-status} locates the complete settings and results.

\begin{table*}[t]
\centering
\caption{The evidence behind each experiment, and where its settings and complete results are given.}
\label{tab:experiment-status}
\footnotesize
\begin{tabularx}{\textwidth}{@{}L{0.65cm}YL{3.6cm}@{}}
\toprule
& Evidence & Sections \\
\midrule
E1 & 32 development laws $\times$ 10 families regenerated from specifications; T2; 64 look-alike and synthesized twin pairs; 96 generated forms; MJX parity, throughput, and gradients; robot twins over nine laws of each family, on fixed bases and in three contact tasks; sixteen Menagerie robots in contact & \ref{sup:e1-forward}, \ref{sup:robots-open}, \ref{sup:contact}, \ref{sup:zoo} \\
E2 & 30 functional checks; 3,888 A1/B1 traces of nine designed predictors, also scored by the reference-rollout distance; intact, degraded, and broken clips of six families; reports on R2, B1, and the three robots; six forecasts of each contact task and the law read through contact; the exact and scaled forecasts of sixteen Menagerie robots; the law of twenty robot scenes read from motion under a wrong inertial table & \ref{sup:e2-validation}--\ref{sup:e2-form}, \ref{sup:robots-open}, \ref{sup:contact}, \ref{sup:zoo}, \ref{sup:law} \\
E3 & 27 state configurations and 30 RGB checkpoints on laws 2000--2015; 27 rollout and common-fit evaluations; 2,448 LingBot-VA forecasts and 936 forecasts of BWM, Ctrl-World, and Cosmos3-Nano; state models on the three robots, three seeds each & \ref{sup:e3-state}--\ref{sup:e3-models}, \ref{sup:robots-open} \\
E4 & Nine coupled state models as planners: 64 candidates, 96 assets, three targets, replay in the forward half; four training corpora $\times$ five seeds on 4,040 generated episodes; four SO-101 corpora $\times$ three seeds on 3,360 episodes & \ref{sup:e4-decisions}, \ref{sup:e4-training}, \ref{sup:robots-open} \\
\bottomrule
\end{tabularx}
\end{table*}

\subsection{E1: Validation of the Forward Half}\label{sup:e1-forward}
\paragraph*{Regeneration} The ten families are regenerated from their specifications for all 32 development laws and compared with the inventories of the corpus (Table~\ref{tab:regeneration}). Table~\ref{tab:forward} summarizes the ranks of the matching systems, the certificate errors, and the relation audits per class.

\begin{table}[t]
\centering
\caption{Validation of the forward half. Rank of the matching system against the number of tunable parameters; largest relative body and engine certificate errors over the 32 development laws; largest gap between a twin and its reference under four shared open-loop programs (eight laws per class), with the twin at its nominal design and after synthesis.}
\label{tab:forward}
\footnotesize
\setlength{\tabcolsep}{2.5pt}
\begin{tabularx}{\columnwidth}{@{}L{1.25cm}L{.9cm}cL{1.45cm}Yl@{}}
\toprule
Class & Twins & Rank & Certificates & Nominal design & Synthesized \\
\midrule
Constant & A1--B3 & 2/2--3 & $1\!\times\!10^{-11}$, $2\!\times\!10^{-10}$ & 0.73--0.94\,rad (A2); others leave the domain & 5.3\,$\mu$rad \\
Variable & P1/R1 & 3/6 & $7\!\times\!10^{-12}$, $2\!\times\!10^{-9}$ & 0.18--0.21\,rad & 22\,$\mu$rad \\
Coupled & P2/R2 & 4/6 & $1\!\times\!10^{-11}$, $6\!\times\!10^{-10}$ & 0.33--0.53\,rad & 0.50\,mrad \\
Coupled & P2/T2 & 6/8 & $1\!\times\!10^{-11}$, $2\!\times\!10^{-8}$ & 0.26--0.40\,rad & 0.50\,mrad \\
\bottomrule
\end{tabularx}
\end{table}

\paragraph*{Synthesized versus drawn twins} For eight development laws per class (laws 0--7), the reference mechanism (A1 for the constant-inertia class, P1 and P2 for the others) and each candidate twin are driven by the same four open-loop programs: chirp, two-tone work, multisine, and a positive pulse. Each candidate is simulated twice: at its nominal design, where every tunable part keeps the value drawn in its specification and the dissipation of the target law is applied, and after synthesis. The gap is the largest absolute difference of the public coordinates over the 4\,s episode and the four programs; a candidate that leaves the admitted domain ($|s_i|>1.1$\,rad) counts as diverged. Table~\ref{tab:forward} gives the gaps per class; at the nominal design all eight-law runs of A3, B1, B2, and B3 leave the domain. The form-twin check releases A1 at densities 0.7 and 1.4 (largest gap $2.2\times10^{-12}$\,rad) and drives both with the same positive pulse (0.050--0.062\,rad).

\paragraph*{Automatically generated forms} For laws 0, 7, 2000, and 2009 of each class, the generator of Section~\ref{sup:forms} produced eight forms per law from the law of the source mechanism (A1, P1, P2). Table~\ref{tab:forms-validation} reports the draws needed, the certificates, and the law-twin audit of every generated form against its source under the chirp and the positive pulse. The variable-inertia class is generated by restyling only.

\begin{table*}[t]
\centering
\caption{Automatically generated forms: four laws per class, eight forms per law. Kept forms over draws; largest relative body and engine certificate errors; largest gap to the source mechanism under shared open-loop efforts; seconds per kept form on one CPU core.}
\label{tab:forms-validation}
\footnotesize
\begin{tabularx}{\textwidth}{@{}L{2.4cm}Yccccc@{}}
\toprule
Class & Families & Kept/draws & Body certificate & Engine certificate & Audit gap (rad) & s per form \\
\midrule
Constant inertia & gear train, restyle & 32/36 & $1.5\times10^{-11}$ & $1.5\times10^{-10}$ & $5.6\times10^{-6}$ & 0.22--0.27 \\
Variable inertia & restyle & 32/32 & $6.8\times10^{-12}$ & $1.1\times10^{-10}$ & $2.2\times10^{-5}$ & 0.41--0.42 \\
Coupled & restyle, serial arm & 32/32 & $9.7\times10^{-11}$ & $2.6\times10^{-8}$ & $5.1\times10^{-4}$ & 0.46--0.58 \\
\bottomrule
\end{tabularx}
\end{table*}

\begin{figure*}[t]
\centering
\includegraphics[width=\textwidth]{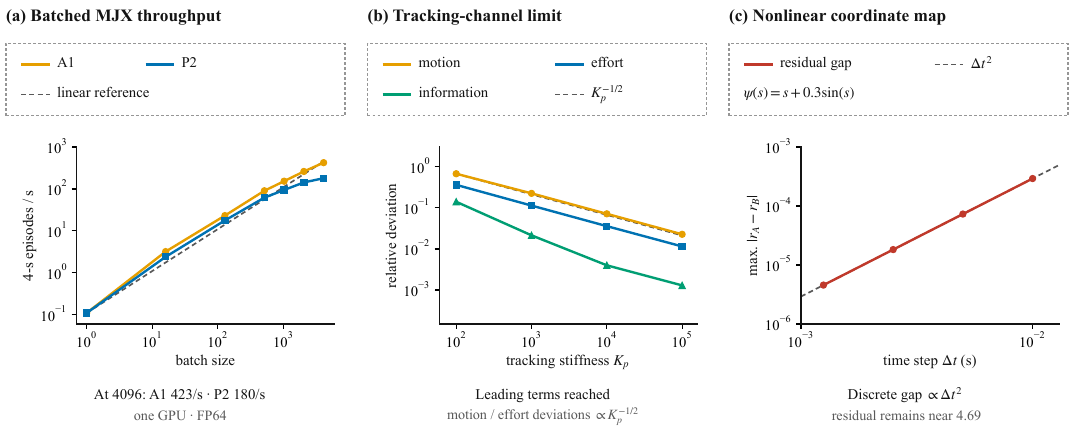}
\caption{Scaling and convergence. (a) MJX throughput against the batch for A1 and P2 on one GPU in double precision, beside linear scaling. (b) The tracking channel of Theorem~\ref{thm:channels} on the common law: the relative deviations of the motion and effort sensitivities from the leading terms of Corollary~\ref{cor:ladder}, and of the motion information from its leading term, against the stiffness. (c) Twins related by the nonlinear map $\psi(s)=s+0.3\sin s$: the largest gap between their discrete residuals against the time step (Section~\ref{app:numerics}).}
\label{fig:checks}
\end{figure*}

\paragraph*{MJX} MJX of MuJoCo 4.0.0 with JAX 0.12.0.1 in double precision on one NVIDIA RTX PRO 6000 (96\,GB) runs every topology at development law 0 under the open-loop chirp for 4\,s at 1200\,Hz. The largest difference of the public coordinates from the C engine is below $8\times10^{-14}$\,rad for A2, A3, B1, and B2, $1.1\times10^{-7}$\,rad for B3, and between $3.8\times10^{-6}$ (R1) and $1.5\times10^{-4}$\,rad (P2) for the others, all below the relation audit. Throughput grows almost linearly with the batch up to 512 episodes and reaches 423 A1 and 180 P2 episodes of 4\,s per second at batch 4,096 (Fig.~\ref{fig:checks}a). The gradient study rolls out the density-$c$ form twin of A1 for 64 values of $c\in[0.7,1.4]$ and differentiates, in forward mode through the whole rollout, the Gauss residual of the rollout under the reference law; the derivative agrees with central differences ($h=10^{-4}$) to a median relative error of $8\times10^{-9}$ (maximum $2.7\times10^{-5}$), and the batch takes 10.2\,s.

\subsection{E2: Validation of the Inverse Half}\label{sup:e2-validation}
Before the corpus validation of Section~\ref{sec:e2} in the main paper, thirty functional checks on exact and deliberately corrupted predictions for a pendulum, a variable-inertia system, and a coupled arm pass. In the corpus validation, a predictor that switches its law by intervention is rejected by the consistency test on every tested intervention. A detector need not reject an action-ignoring predictor on a release with zero input; such interventions remain in the denominator and are interpreted by their intended failure mode.

\subsection{E2: Reference-Rollout Baseline}\label{sup:e2-baseline}
The nine predictors are generated on A1 and B1 assets that the forward half synthesizes for laws 24--31 at densities 0.7, 1.0, and 1.4 (48 assets, nine probes). Every prediction, the exact one included, is integrated by the same solver, so the floors reflect numerical consistency; Table~\ref{tab:baseline} compares the scores predictor by predictor. Each is scored three ways: the RMS distance from the simulated rollout of the same query (rad), the Gauss residual under the true law (\Nm{}), and the power residual (J). Floors are per probe: twice the 99th percentile of the exact predictor on calibration laws 24--27; rates are on test laws 28--31. Two readout settings are compared: exact states at 120\,Hz, as above, and the 10\,Hz clock of the video readout with i.i.d.\ noise of 0.0024\,rad, the median readout error on codec reconstructions.

\begin{table*}[t]
\centering
\caption{Reference-rollout distance (D) against the Gauss residual (G) on the nine designed predictors, 4\,s horizon, test laws 28--31. Exact states: fraction of forecasts flagged over all nine probes, and the median value in multiples of its per-probe floor under open-loop efforts and under tracking. Video clock: fraction flagged at 10\,Hz with 0.0024\,rad readout noise.}
\label{tab:baseline}
\footnotesize
\begin{tabularx}{\textwidth}{@{}L{3.2cm}ccYYYY@{}}
\toprule
& \multicolumn{4}{c}{Exact states, 120\,Hz} & \multicolumn{2}{c}{Video clock, 10\,Hz} \\
Predictor & D flagged & G flagged & Margin D / G, open loop & Margin D / G, tracked & Flagged D / G, open loop & Flagged D / G, tracked \\
\midrule
Exact & 0 & 0 & 0.5 / 0.4 & 0.5 / 0.3 & 0 / 0 & 0 / 0 \\
Ignores actions & 0.67 & 0.67 & 92 / 882 & 1{,}666 / 723{,}234 & 1 / 0.54 & 1 / 1 \\
Mean law & 1 & 1 & 78 / 433 & 114 / 24{,}710 & 1 / 0.40 & 0.67 / 0.67 \\
Switch by mechanism & 0.33 & 0.33 & 6.8 / 47 & 58 / 7{,}847 & 0.50 / 0.12 & 0.50 / 0.50 \\
Switch by intervention & 0.67 & 0.67 & 22 / 203 & 141 / 84{,}621 & 1 / 0 & 1 / 0.92 \\
Time warp & 1 & 1 & 288 / 1{,}580 & 871 / 384{,}104 & 1 / 1 & 1 / 1 \\
Scale-blind & 0.44 & 0.44 & 22 / 153 & 116 / 16{,}704 & 0.67 / 0.17 & 0.67 / 0.67 \\
Servo only & 0.56 & 0.56 & 92 / 882 & 464 / 596{,}324 & 1 / 0.54 & 0.50 / 0.50 \\
Stationary & 1 & 1 & 462 / 7{,}004 & 1{,}666 / 723{,}234 & 1 / 1 & 1 / 1 \\
\bottomrule
\end{tabularx}
\end{table*}

A fault that leaves the motion of a probe unchanged (an action-ignoring predictor on a release, a scale-blind predictor in free motion) escapes both scores. Over the horizon, the median rollout distance of the mean-law predictor grows from 0.033 to 0.054\,rad under open-loop efforts while its residual stays at 0.22\,\Nm{}; for the time warp the distance grows from 0.075 to 0.198\,rad.

\subsection{E2: Faults of Form}\label{sup:e2-form}
Known-pose clips of the six twin families A1, B1, R1, P1, R2, and P2 (development laws 24 and 25, density 1.0, servo program, 12 frames at 10\,Hz) are rendered at the readout resolution. Intact clips are clean or degraded by Gaussian blur ($\sigma=1.2$\,px), JPEG compression (quality 25), the color transform of the appearance control, downscaling to 128\,px and back, or pixel noise ($\sigma=6/255$). Broken clips displace one link (A1: arm1; B1: rack carriage; P1, P2: lower0; R1, R2: link2) from frame 1 on by 1, 2, 3, 5, 8, 12, or 25\,mm along $x$ or $z$. The frozen readout fits every body without joint projection, and the verdict of Section~\ref{sup:readout} is applied to every joint and frame. Thresholds are twice the 99th percentile of the evidence on intact law-24 clips; law 25 is the test. Table~\ref{tab:formfaults} gives the false alarms and detections per family under both calibrations.

\begin{table*}[t]
\centering
\caption{Faults of form on law 25. False alarms among the six intact clips; detections out of two directions at 1, 2, 3, 5, 8, 12, and 25\,mm; smallest displacement detected in both directions at it and every larger size; largest pin-gap threshold. Left: thresholds calibrated on all intact law-24 clips, the protocol; right: on clean law-24 clips only.}
\label{tab:formfaults}
\footnotesize
\begin{tabularx}{\textwidth}{@{}lcYcc|cYc@{}}
\toprule
& \multicolumn{4}{c|}{All degradations (protocol)} & \multicolumn{3}{c}{Clean references only} \\
Family & False alarms & Detected per size & Smallest (mm) & Gap threshold (mm) & False alarms & Detected per size & Smallest (mm) \\
\midrule
A1 & 0/6 & 0 0 0 2 2 2 2 & 5 & 9.5 & 0/6 & 2 2 2 2 2 2 2 & 1 \\
B1 & 0/6 & 0 2 2 2 2 2 2 & 2 & 1.0 & 0/6 & 1 2 2 2 2 2 2 & 2 \\
P1 & 0/6 & 0 0 0 0 0 0 2 & 25 & 13.7 & 1/6 & 1 2 2 2 2 2 2 & 2 \\
P2 & 0/6 & 0 0 0 0 0 0 2 & 25 & 23.0 & 1/6 & 1 2 1 2 2 2 2 & 5 \\
R1 & 0/6 & 0 0 0 0 2 2 2 & 8 & 7.5 & 1/6 & 1 2 2 2 2 2 2 & 2 \\
R2 & 0/6 & 0 0 0 0 0 0 1 & >25 & 12.1 & 2/6 & 1 2 2 2 2 2 2 & 2 \\
\bottomrule
\end{tabularx}
\end{table*}

\paragraph*{Reports} The reports of Section~\ref{sup:reports} were checked on R2 (law 2009) and B1 (law 2000) with the reference predictors of known faults, read as state forecasts on a 60\,Hz clock and as videos on a 10\,Hz clock through the image readout, with thresholds calibrated on clean, blurred, and JPEG-compressed references of each mechanism (Table~\ref{tab:reports-validation}). Another law is law 2012 for R2 (its ratios to the true law are 1.02, 0.94, 0.94, 1.01, 0.94 at the state clock) and law 2003 for B1.

\begin{table}[t]
\centering
\caption{Reports on forecasts with known faults, as states (60\,Hz) and as videos (10\,Hz).}
\label{tab:reports-validation}
\footnotesize
\setlength{\tabcolsep}{3pt}
\begin{tabularx}{\columnwidth}{@{}llL{1.6cm}Y@{}}
\toprule
 & Clock & Forecast & Finding \\
\midrule
R2 & state & exact & follows the true law within the floor \\
R2 & state & law $\times1.4$ & true law scaled by 1.40 \\
R2 & state & time warp 0.8 & no single law; true motion at 0.80 times its speed \\
R2 & state & stationary & no single law; motion 0.00 of the truth \\
R2 & state & another law & not the true law: upper inertia $\times$1.02, coupling $\times$0.94, lower inertia $\times$0.94, upper gravity $\times$1.01, lower gravity $\times$0.94 \\
R2 & video & exact & follows the true law within the floor \\
R2 & video & law $\times1.4$ & true law scaled by 1.38 \\
R2 & video & time warp 0.8 & no single law; true motion at 0.80 times its speed \\
R2 & video & stationary & no single law; motion 0.00 of the truth \\
R2 & video & another law & true law scaled by 0.93 (a near-uniform change) \\
R2 & video & body displaced 30\,mm & form broken at pin link1-link2 \\
B1 & state & exact & follows the true law within the floor \\
B1 & state & law $\times1.4$ & true law scaled by 1.40 \\
B1 & state & time warp 0.8 & no single law; true motion at 0.80 times its speed \\
B1 & state & stationary & no single law; motion 0.00 of the truth \\
B1 & state & another law & not the true law: inertia $\times$1.18, gravity moment $\times$0.91 \\
B1 & video & exact & follows the true law within the floor \\
B1 & video & law $\times1.4$ & true law scaled by 1.43 \\
B1 & video & time warp 0.8 & no single law; true motion at 0.80 times its speed \\
B1 & video & stationary & no single law; motion 0.00 of the truth \\
B1 & video & another law & not the true law: inertia $\times$1.23, gravity moment $\times$1.01 \\
B1 & video & body displaced 30\,mm & form broken at coupling rack\_carriage\_x \\
\bottomrule
\end{tabularx}
\end{table}

\subsection{E3: Models Given Canonical Coordinates}\label{sup:e3-state}
Support and query episodes are disjoint. The state models are scored by the trajectory analogue of \eqref{eq:change}: the RMS difference between the predicted and the true change of the rollout trajectory when the target density changes from nominal (Table~\ref{tab:state-transfer}).

\begin{table*}[t]
\centering
\caption{Form against law for models given canonical coordinates: target mechanisms, sixteen confirmation laws, both changed densities, and three training seeds. RMS of the trajectory change caused by the density change (true), and RMS error of the predicted change, in mrad with 95\% law-cluster intervals, under evidence from the twin or the target itself, under the same or another law. Without evidence the predicted change is zero, so its error equals the true change.}
\label{tab:state-transfer}
\footnotesize
\begin{tabularx}{\textwidth}{@{}L{2.1cm}L{2.5cm}L{1.1cm}YYYY@{}}
\toprule
Dynamics & Model & True change & Twin, same law & Own, same law & Own, other law & Twin, other law \\
\midrule
Constant inertia & Context GRU & 38.7 & 2.7 [2.4, 3.1] & 4.6 [4.1, 5.1] & 74.7 [72.3, 77.0] & 76.8 [74.3, 79.3] \\
Constant inertia & Mechanical context & 38.7 & 4.2 [3.8, 4.8] & 6.6 [5.8, 7.5] & 73.3 [70.4, 76.0] & 74.5 [72.0, 76.9] \\
Variable inertia & Context GRU & 9.9 & 2.2 [2.2, 2.3] & 5.2 [5.1, 5.3] & 14.7 [14.4, 14.9] & 21.1 [20.9, 21.4] \\
Variable inertia & Mechanical context & 9.9 & 2.8 [2.7, 2.8] & 4.4 [4.3, 4.5] & 16.3 [16.0, 16.6] & 21.6 [21.2, 21.9] \\
Coupled 2-DOF & Context GRU & 37.3 & 5.7 [5.3, 6.1] & 15.7 [15.0, 16.5] & 63.0 [61.9, 64.1] & 74.0 [72.7, 75.4] \\
Coupled 2-DOF & Mechanical context & 37.3 & 3.3 [3.1, 3.5] & 16.5 [16.0, 17.0] & 62.3 [61.3, 63.3] & 72.8 [71.5, 74.0] \\
\bottomrule
\end{tabularx}
\end{table*}

\subsection{E3: RGB Context Models}\label{sup:e3-rgb}
\paragraph*{Settings} RGB context and no-context models are trained in three dynamics classes with five seeds (11, 23, 47, 67, 89; 30 checkpoints) and also receive the timing of every frame. Training uses certified same-law pairs, so the experiment assesses transfer learned from paired supervision, not unsupervised discovery of correspondences. The 96-pixel models observe a three-frame prefix and predict ten frames at 10\,Hz; recorded efforts are playback inputs. Each run uses 12,000 steps, a four-frame training horizon, and a ten-frame evaluation horizon, and selects its checkpoint by validation RGB loss. Evidence contains the first eight frames at 10\,Hz of two separate signed-pulse episodes; training pairs use correct same-mechanism evidence with probability 0.45, correct cross-mechanism evidence with probability 0.45, and none with probability 0.10, and the no-context model ignores evidence. For reference future $Y_{tpc}$ and last-prefix image $X_{pc}$, with color intensities in $[0,1]$, the evaluation mask and loss are
\begin{align}
D_{tp}&=\mathbf 1\!\left\{\tfrac13\sum_c|Y_{tpc}-X_{pc}|>0.02\right\},\nonumber\\
E_{\rm motion}&=\frac{\sum_{t,p,c}D_{tp}(\hat Y_{tpc}-Y_{tpc})^2}{\max(3\sum_{t,p}D_{tp},1)}.
\label{eq:rgb-motion-score}
\end{align} Full-frame MSE averages the same squared errors over all pixels and channels; the reference-defined mask is common to all predictions of a query and never enters model input. Neither score is an SI dynamics measure; the 96-pixel predictions are scored in image space only.

\begin{table*}[t]
\centering
\caption{RGB transfer on changed-density target mechanisms, five training seeds. Paired motion-region MSE differences ($\times10^{-3}$), correct cross-mechanism support minus each comparator, with 95\% law-bootstrap intervals over sixteen laws. The first three comparisons hold the checkpoint fixed; the last compares with a separately trained no-context model. Negative values favor correct support.}
\label{tab:rgb-main}
\footnotesize
\begin{tabularx}{\textwidth}{@{}L{1.4cm}L{.55cm}YYYY@{}}
\toprule
Dynamics & $T$ (s) & Wrong physics & Empty history & Nominal physics & Separate no-context model \\
\midrule
Constant & 1 & $-0.198$ [-0.223, -0.179] & $+0.054$ [+0.023, +0.084] & $+0.046$ [+0.017, +0.073] & $+0.977$ [+0.857, +1.097] \\
Constant & 2 & $-0.225$ [-0.263, -0.188] & $+0.315$ [+0.258, +0.377] & $+0.311$ [+0.263, +0.363] & $+0.620$ [+0.508, +0.730] \\
Constant & 3 & $-0.131$ [-0.163, -0.105] & $+0.348$ [+0.265, +0.427] & $+0.295$ [+0.257, +0.329] & $-0.036$ [-0.137, +0.072] \\
Variable & 1 & $-0.402$ [-0.419, -0.385] & $-0.593$ [-0.674, -0.514] & $-0.424$ [-0.442, -0.407] & $+1.156$ [+1.068, +1.240] \\
Variable & 2 & $-0.370$ [-0.386, -0.355] & $-0.558$ [-0.596, -0.521] & $-0.316$ [-0.327, -0.305] & $-9.963$ [-10.035, -9.895] \\
Variable & 3 & $-0.190$ [-0.198, -0.181] & $-0.527$ [-0.538, -0.516] & $-0.086$ [-0.098, -0.074] & $-6.929$ [-7.016, -6.848] \\
Coupled & 1 & $-9.819$ [-9.978, -9.673] & $-5.220$ [-5.926, -4.589] & $-4.361$ [-4.462, -4.246] & $-1.592$ [-2.573, -0.711] \\
Coupled & 2 & $-6.545$ [-6.631, -6.460] & $-2.624$ [-3.139, -2.163] & $-2.130$ [-2.188, -2.072] & $+4.074$ [+2.784, +5.284] \\
Coupled & 3 & $-5.120$ [-5.232, -5.019] & $-2.109$ [-2.464, -1.801] & $-1.603$ [-1.651, -1.564] & $+1.138$ [+0.387, +1.858] \\
\bottomrule
\end{tabularx}
\end{table*}

\paragraph*{Evidence use} Beyond the sensitivity to evidence and the adjustment of Section~\ref{sec:rgb-result} in the main paper, Table~\ref{tab:rgb-main} shows a third level, i.e., benefit. The coupled model improves on its own empty-history forecast in every seed at every horizon, and the variable-inertia model improves on average at every horizon, with one seed reversing at 1 and 3\,s. The constant-inertia model only reaches the level of sensitivity, since its changed-physics history is slightly worse than nominal or empty history. Moreover, transfer depends on its direction: the correspondence between A1 and B1 holds both ways, but for A1 queries with B1 support, correct history raises motion-region error by $16.4\times10^{-3}$ relative to empty history in every seed (Table~\ref{tab:rgb-crossed}), whereas the variable-inertia and coupled pairs transfer in both directions. However, the context pathway does not improve the forecast uniformly: compared with the separately trained no-context model, the coupled context model is better at 1\,s but worse at 2 and 3\,s (Table~\ref{tab:rgb-seeds}). Without history the prediction does not change, so its error equals the true change; wrong-law support comes from the opposite density change.

\paragraph*{Crossed evidence and horizon}\label{sec:rgb-protocol}
Source and target families, changed and nominal densities, are reported together; Table~\ref{tab:rgb-ablation} gives the effect of each intervention on the evidence.

The visual protocol crosses two families, three densities, two query probes, twelve support conditions, and three prefix horizons for every law and model. It yields 192 queries per model, each evaluated under twelve conditions at 1, 2, and 3\,s. Horizons share the same thirty-frame rollout and are dependent observations. The twelve conditions are correct, wrong-physics, and nominal-physics histories from the same or other mechanism; empty history; and cross-mechanism static, action-shuffled, frame-shuffled, jointly shuffled, and alternate-probe histories. All support episodes are disjoint from query episodes.

Cross support always uses the other mechanism, including reverse-direction queries. In contrast, the state and planning protocols retain a fixed source convention: their source-family ``cross'' conditions use source self-history. Only their target-family results establish cross-mechanism evidence use. Target B1/P1/P2 changed-density combinations remain excluded from training and selection. Source/nominal conditions test the other cells of the composition rather than increasing the number of unseen target combinations.

\begin{table*}[t]
\centering
\caption{Full crossed RGB comparison at 1\,s: correct cross history minus each comparator, motion-region MSE $\times10^{-3}$ with 95\% law-bootstrap intervals. Source and target families, nominal and changed densities, share five trained seeds and sixteen fresh laws. At nominal density the correct and nominal histories coincide.}
\label{tab:rgb-crossed}
\footnotesize
\begin{tabularx}{\textwidth}{@{}L{1.3cm}L{1cm}L{1.3cm}YYY@{}}
\toprule
Dynamics & Family & Physics & Wrong physics & Empty history & Nominal physics \\
\midrule
Constant & A1 & nominal & $-0.036$ [-0.054, -0.018] & $+18.891$ [+17.823, +19.941] & $+0.000$ [-0.000, +0.000] \\
Constant & A1 & changed & $+0.059$ [+0.043, +0.074] & $+16.399$ [+15.467, +17.329] & $+0.030$ [+0.019, +0.040] \\
Constant & B1 & nominal & $-0.083$ [-0.103, -0.062] & $+0.001$ [-0.029, +0.028] & $-0.000$ [-0.000, +0.000] \\
Constant & B1 & changed & $-0.198$ [-0.223, -0.179] & $+0.054$ [+0.023, +0.084] & $+0.046$ [+0.017, +0.073] \\
Variable & R1 & nominal & $-0.338$ [-0.369, -0.308] & $-0.112$ [-0.175, -0.053] & $-0.000$ [-0.000, +0.000] \\
Variable & R1 & changed & $-0.346$ [-0.374, -0.321] & $-0.405$ [-0.476, -0.339] & $-0.299$ [-0.310, -0.285] \\
Variable & P1 & nominal & $-0.620$ [-0.664, -0.578] & $-0.175$ [-0.261, -0.093] & $+0.000$ [+0.000, +0.000] \\
Variable & P1 & changed & $-0.402$ [-0.419, -0.385] & $-0.593$ [-0.674, -0.514] & $-0.424$ [-0.442, -0.407] \\
Coupled & R2 & nominal & $-7.338$ [-7.757, -6.926] & $-1.042$ [-1.917, -0.232] & $-0.000$ [-0.000, +0.000] \\
Coupled & R2 & changed & $-16.138$ [-16.486, -15.794] & $-7.254$ [-8.041, -6.481] & $-6.360$ [-6.560, -6.164] \\
Coupled & P2 & nominal & $-1.860$ [-2.056, -1.676] & $-1.002$ [-1.933, -0.193] & $+0.000$ [-0.000, +0.000] \\
Coupled & P2 & changed & $-9.819$ [-9.978, -9.673] & $-5.220$ [-5.926, -4.589] & $-4.361$ [-4.462, -4.246] \\
\bottomrule
\end{tabularx}
\end{table*}

Shuffling permutes entries within each eight-frame support episode with a fixed seed, retaining time and episode labels. Action shuffling changes the effort entries; frame shuffling changes images; joint shuffling preserves image/action pairing while breaking their time alignment. The static condition repeats the initial image of each episode while retaining its action record. The alternate condition substitutes two independent signed-velocity probes for the signed-pulse support. These interventions alter the evidence but not the query. They are not certified alternative laws, and their effect depends on the variation and informativeness of the chosen probes.

\begin{table*}[t]
\centering
\caption{Evidence interventions on changed-density target RGB queries: mean motion-region MSE difference ($\times10^{-3}$), correct cross support minus the altered support. Negative values favor intact support.}
\label{tab:rgb-ablation}
\footnotesize
\begin{tabularx}{\textwidth}{@{}L{1.5cm}L{.7cm}YYYYY@{}}
\toprule
Dynamics & $T$ (s) & Static & Action shuffle & Frame shuffle & Joint shuffle & Alternate probes \\
\midrule
Constant & 1 & -2.624 & -0.010 & -0.493 & -0.507 & -0.585 \\
Constant & 2 & -2.191 & -0.019 & +0.054 & +0.044 & -0.622 \\
Constant & 3 & -1.468 & -0.028 & +0.793 & +0.791 & -0.777 \\
Variable & 1 & +0.645 & -0.005 & +0.136 & +0.145 & -0.315 \\
Variable & 2 & +0.570 & -0.001 & +0.114 & +0.116 & -0.170 \\
Variable & 3 & +1.461 & -0.002 & -0.930 & -0.931 & -1.441 \\
Coupled & 1 & -18.057 & -0.028 & -11.677 & -11.658 & -16.269 \\
Coupled & 2 & -12.177 & -0.012 & -7.724 & -7.703 & -17.087 \\
Coupled & 3 & -12.856 & -0.014 & -8.640 & -8.619 & -17.592 \\
\bottomrule
\end{tabularx}
\end{table*}

The motion-region mask remains \eqref{eq:rgb-motion-score}; empty masks are retained as unmeasurable rather than scored as perfect. Full-frame and persistence scores are saved alongside it. Bootstrap units are sixteen laws, with repeated probes, densities, and five seeds averaged inside each law. This confirms the tested fits across new laws, while Table~\ref{tab:rgb-seeds} shows uncertainty due to fitting. For example, the variable-inertia model's mean separate-model advantage at 2\,s includes a $-47.427\times10^{-3}$ effect at seed 23, while three of five seeds favor the no-context model. The mean effect is therefore not consistent across training runs.

\begin{table*}[t]
\centering
\caption{Per-training-seed effects for changed-density target RGB queries, correct cross support minus empty history in the same model or the separately trained no-context model. Motion-region MSE $\times10^{-3}$; seeds are not replaced by law-bootstrap confidence intervals.}
\label{tab:rgb-seeds}
\footnotesize
\begin{tabularx}{\textwidth}{@{}L{1.4cm}L{.6cm}L{2.4cm}YYYYY@{}}
\toprule
Dynamics & $T$ (s) & Comparator & 11 & 23 & 47 & 67 & 89 \\
\midrule
Constant & 1 & Empty history & +0.074 & -0.025 & +0.092 & -0.037 & +0.164 \\
Constant & 1 & Separate model & +0.217 & -0.227 & +0.218 & -0.583 & +5.257 \\
Constant & 2 & Empty history & +1.224 & +0.112 & +0.126 & -0.010 & +0.122 \\
Constant & 2 & Separate model & +1.363 & -0.749 & -1.674 & -0.670 & +4.832 \\
Constant & 3 & Empty history & +0.854 & +0.404 & +0.134 & +0.107 & +0.240 \\
Constant & 3 & Separate model & +4.749 & -1.097 & -5.152 & -2.418 & +3.736 \\
Variable & 1 & Empty history & -0.415 & -1.205 & -0.583 & -0.811 & +0.050 \\
Variable & 1 & Separate model & -3.581 & +5.387 & +0.708 & +5.093 & -1.829 \\
Variable & 2 & Empty history & -0.344 & -1.087 & -0.441 & -0.722 & -0.195 \\
Variable & 2 & Separate model & -3.977 & -47.427 & +0.998 & +0.518 & +0.074 \\
Variable & 3 & Empty history & -0.096 & -0.863 & -0.454 & -1.342 & +0.121 \\
Variable & 3 & Separate model & -3.867 & -39.196 & +2.335 & +6.137 & -0.051 \\
Coupled & 1 & Empty history & -6.411 & -2.994 & -5.569 & -4.697 & -6.427 \\
Coupled & 1 & Separate model & -5.742 & -6.217 & +0.036 & -4.083 & +8.045 \\
Coupled & 2 & Empty history & -3.778 & -0.988 & -2.564 & -1.071 & -4.719 \\
Coupled & 2 & Separate model & -1.625 & -0.956 & +6.241 & +4.707 & +12.003 \\
Coupled & 3 & Empty history & -2.643 & -1.647 & -2.303 & -0.729 & -3.225 \\
Coupled & 3 & Separate model & -7.364 & +1.034 & +4.135 & -0.375 & +8.259 \\
\bottomrule
\end{tabularx}
\end{table*}

\subsection{E3: Accuracy versus Structure}\label{sup:e3-structure}
Each of the 27 state predictors is rolled out for 4\,s. Besides the shared fit of Section~\ref{sec:e3} in the main paper, separate fits to each query give an in-sample reference, and the same procedure on simulated trajectories measures the explainer's own error.

\begin{table*}[t]
\centering
\caption{Common explanations for target mechanisms, all three densities, correct source history, and three training seeds. Mean effort RMSE (\Nm). ``Separate'' fits each query in sample; ``shared'' fits impedance/servo forecasts and tests chirp/work forecasts. The last two columns apply a law fitted to native support to model forecasts and to native queries. A low shared error alone does not establish physical correctness.}
\label{tab:common-structure}
\footnotesize
\begin{tabularx}{\textwidth}{@{}L{1.6cm}L{3cm}YYYY@{}}
\toprule
Dynamics & Predictor & Separate & Shared & Native law on prediction & Native law on native query \\
\midrule
Constant & MLP & 0.0958 & 0.3422 & 0.2603 & 0.0070 \\
Constant & Context GRU & 0.0112 & 0.0480 & 0.0355 & 0.0070 \\
Constant & Mechanical context & $7.30\!\times\!10^{-4}$ & 0.0042 & 0.0235 & 0.0070 \\
Variable & MLP & 0.1225 & 0.7959 & 0.2396 & 0.0083 \\
Variable & Context GRU & 0.0261 & 0.3825 & 0.0750 & 0.0083 \\
Variable & Mechanical context & $5.01\!\times\!10^{-5}$ & $6.14\!\times\!10^{-5}$ & 0.0353 & 0.0083 \\
Coupled & MLP & 0.1531 & 0.6957 & 0.3446 & 0.0161 \\
Coupled & Context GRU & 0.0267 & 0.0974 & 0.0377 & 0.0161 \\
Coupled & Mechanical context & $3.52\!\times\!10^{-4}$ & 0.0082 & 0.0193 & 0.0161 \\
\bottomrule
\end{tabularx}
\end{table*}

Table~\ref{tab:common-structure} gives the values behind Section~\ref{sec:e3} in the main paper. For the variable-inertia GRU, the shared structure's 0.3825\,\Nm{} on held-out interventions compares with 0.0261\,\Nm{} for separate in-sample fits. For the mechanical-context model, whose single structure explains its forecasts to $6.1\times10^{-5}$\,\Nm{}, the law fitted to simulated support leaves 0.0353\,\Nm{} on its forecasts, four times its 0.0083\,\Nm{} error on simulated queries.

In the response diagnostics (Table~\ref{tab:response-geometry}), the mechanical-context model meets the structure relations to $2.1\times10^{-8}$\,\Nm{} and a curl of $3.5\times10^{-12}$; DeLaN and its dissipative extension meet them to $3.3\times10^{-7}$ and $4.3\times10^{-7}$\,\Nm{}, and the Gauss anchor exposes their two poorly fitted coupled seeds with mean residuals above $3\times10^{3}$.

\paragraph*{Settings}
The 54 frozen state configurations are queried on the confirmation laws. The response-geometry diagnostic uses 64 matched states in the declared local domain, configuration coordinates in $[-0.3,0.3]$, velocities in $[-0.6,0.6]$, signed effort steps of $0.02$\,\Nm{}, and coordinate steps of 0.005 and 0.01\,rad. Model context is fixed across each set of local probes. Reciprocity, input affinity, velocity dependence, metric validity, Christoffel compatibility, storage curl, and dissipated power are retained as separate outputs.

\paragraph*{Numerical control before structural interpretation} A coupled dissipative-DeLaN checkpoint illustrates the need for \eqref{eq:inverse-sensitivity}. At the 0.005\,rad coordinate step, its mean geometry-force mismatch is 9.761\,\Nm{} in FP32, but $9.52\times10^{-7}$\,\Nm{} when the same weights and queries are evaluated in FP64 (Table~\ref{tab:precision}). The inferred mass condition number is approximately 1,828. Thus the apparent structural failure in this case is numerical amplification, not a violation of the architecture's mechanical identity. Other coupled runs have large validation prediction error and learned masses near the numerical floor; higher precision does not repair that fit. Structural compatibility and agreement with the true law answer different questions, and both remain in the final comparison.

\begin{table}[t]
\centering
\caption{Precision control on the same coupled dissipative-DeLaN checkpoint (seed 11). The coordinate difference is 0.005\,rad.}
\label{tab:precision}
\footnotesize
\begin{tabular}{@{}lrr@{}}
\toprule
Precision & Max.\ mass condition & Geometry mismatch (\Nm) \\
\midrule
FP32 & 1827.7 & $9.761$ \\
FP64 & 1827.6 & $9.52\times10^{-7}$ \\
\bottomrule
\end{tabular}
\end{table}

\paragraph*{Rollouts and explainer}\label{sec:common-structure}
Rollouts use RK4 at 120\,Hz for 4\,s and cease reporting after leaving $|q_i|\leq1.1$ or $|v_i|\leq20$. They are not projected back into the domain. Actual feedback efforts for impedance/servo are reconstructed on predicted states. The common explainer uses nonnegative mechanical coefficients, positive-inertia constraints, and a shared coefficient for variable inertia and its required geometric force; query predictions never select its coefficients. Friction width in the constant-inertia explainer is a declared physical prior, not model input.

The native-fitted reference is evaluated both on native trajectories and on predictions. This controls the error of the explainer, differentiation, and cross-intervention extrapolation. In particular, a learned smooth mechanical trajectory can have lower shared-fit residual than the native-reference residual without being a more accurate law. The GRU/mechanical-context comparisons preserve this distinction.

\begin{table*}[t]
\centering
\caption{FP64 response diagnostics on target mechanisms, all densities, and three seeds, at coordinate step 0.005\,rad. Context models receive correct source evidence; published controls have no context interface. Coverage is the fraction of valid candidate metrics. Geometry disagreement is in \Nm; storage curl uses the declared rotational coordinates; negative dissipation is a fraction of valid points. The final column is the independent mass-weighted Gauss norm, not an effort error.}
\label{tab:response-geometry}
\footnotesize
\begin{tabularx}{\textwidth}{@{}L{1.3cm}L{2.8cm}YYYYY@{}}
\toprule
Dynamics & Predictor & Metric coverage & Geometry mismatch & Storage curl & Negative dissipation & Gauss anchor \\
\midrule
Constant & MLP & 1.0000 & 0.1051 & 0 & 0 & 0.2618 \\
Constant & Context GRU & 1.0000 & 0.0063 & 0 & 0 & 0.0345 \\
Constant & Mechanical context & 1.0000 & $1.06\!\times\!10^{-14}$ & 0 & 0 & 0.0218 \\
Constant & Canonical MLP & 1.0000 & 0.1174 & 0 & 0 & 0.2762 \\
Constant & DeLaN & 1.0000 & $9.15\!\times\!10^{-7}$ & 0 & 0 & 0.3125 \\
Constant & Dissipative DeLaN & 1.0000 & $7.78\!\times\!10^{-7}$ & 0 & 0 & 0.2663 \\
Variable & MLP & 1.0000 & 0.2723 & 0 & 0.3854 & 0.0957 \\
Variable & Context GRU & 1.0000 & 0.0442 & 0 & 0 & 0.0317 \\
Variable & Mechanical context & 1.0000 & $3.73\!\times\!10^{-8}$ & 0 & 0 & 0.0243 \\
Variable & Canonical MLP & 1.0000 & 0.2245 & 0 & 0.0260 & 0.0497 \\
Variable & DeLaN & 1.0000 & $4.62\!\times\!10^{-6}$ & 0 & 0 & 0.0620 \\
Variable & Dissipative DeLaN & 1.0000 & $4.85\!\times\!10^{-6}$ & 0 & 0 & 0.0666 \\
Coupled & MLP & 1.0000 & 0.0590 & 1.9087 & 0.0938 & 0.2261 \\
Coupled & Context GRU & 1.0000 & 0.0088 & 0.3739 & 0.0130 & 0.0657 \\
Coupled & Mechanical context & 1.0000 & $2.07\!\times\!10^{-8}$ & $3.54\!\times\!10^{-12}$ & 0 & 0.0252 \\
Coupled & Canonical MLP & 1.0000 & 0.1086 & 3.9046 & 0.1719 & 0.2084 \\
Coupled & DeLaN & 1.0000 & $3.34\!\times\!10^{-7}$ & $3.43\!\times\!10^{-6}$ & 0 & $3.36\!\times\!10^{3}$ \\
Coupled & Dissipative DeLaN & 1.0000 & $4.33\!\times\!10^{-7}$ & $6.59\!\times\!10^{-6}$ & 0 & $3.34\!\times\!10^{3}$ \\
\bottomrule
\end{tabularx}
\end{table*}

The one-dimensional storage curl is identically zero and therefore uninformative about a global potential. Context-free predictors can return the same structural diagnostic for every law at the fixed query grid; their zero-width law intervals do not imply certainty over training seeds. The native anchor still changes with the physical law. In the coupled DeLaN runs, two poorly fitted seeds produce large physical errors despite small geometry disagreement. This demonstrates the need to report structure and anchoring together, with the precision control preventing a floating-point artifact from being labeled a structural violation.

\subsection{E3: LingBot-VA}\label{sup:e3-lingbot}
\paragraph*{Protocol}\label{sec:lingbot-protocol}
The confirmation protocol is frozen before generation. LingBot-VA (released weights) receives a three-frame prefix, commanded joint positions at 60\,Hz on its two action channels, and, under the history conditions, a separate support episode of the other closed-loop probe through its interleaved video--action cache; the video codec is reset at episode boundaries while the transformer history is kept. It generates 120 frames at 30\,Hz with five inference steps. Changing the commands under a fixed seed first changes the generated pixels at frame 15 (0.5\,s) in all six mechanisms, so cross-command fits start after 0.5\,s. The matrix covers laws 2000, 2003, 2006, 2009, 2012, and 2015; the twin pairs A1/B1, R1/P1, and R2/P2; densities 0.7, 1.0, and 1.4; servo and impedance queries; five histories (none, correct same-mechanism, correct cross-mechanism, wrong-physics cross-mechanism, and a static image of the twin); and generation seeds 11 and 23, for 2,160 forecasts. Laws 2000 and 2012 add zeroed and negated commands (288 forecasts). All 2,448 forecasts are analyzed without errors, giving 1,080 twin pairs, 720 density pairs, 1,728 history contrasts, 144 cross-command structure fits, and 72 cross-mechanism structure fits.

\begin{table}[t]
\centering
\caption{Capability profile of LingBot-VA: six confirmation laws, three twin pairs, three densities, servo and impedance queries, and two generation seeds, scored as a closed-loop law with the known controller. Brackets are 95\% law-cluster intervals; the last column gives the value of an exact model.}
\label{tab:lingbot}
\footnotesize
\setlength{\tabcolsep}{3pt}
\begin{tabularx}{\columnwidth}{@{}L{1.6cm}YL{1.3cm}@{}}
\toprule
Relation & LingBot-VA & Exact \\
\midrule
Projection & 2,361 of 2,448 forecasts pass every check; coverage 99.9\% & all \\
Motion, no history & 0.19 of the true amplitude & 1 \\
Commands zeroed or negated & 0.021 / 0.022 rad (nominal 0.024) & 0.007 / 0.084 \\
Reuse, no history & twin gap 0.022 rad, anchor error 0.094 rad & 0, 0 \\
Adjustment & response gain $-0.14$ [$-0.25$, $-0.04$] & 1 \\
Consistency & held-out command branch 1.1\,\Nm{} & $9.1\times10^{-5}$ \\
Anchor & effort residual 13--107$\times$ floor & $\approx1\times$ \\
Evidence vs none & same $+0.010$, cross $+0.059$, wrong $+0.075$, static $+0.031$ rad & $\le0$ \\
Form verdict & 2,057 broken, 122 intact, 269 undecidable & all intact \\
\bottomrule
\end{tabularx}
\end{table}

\paragraph*{Profile} Table~\ref{tab:lingbot} gives the capability profile behind Section~\ref{sec:e3} in the main paper, and Table~\ref{tab:lingbot-detail} its measurement record by mechanism. Without history the median amplitude is 0.11--0.18 of the true one for all but the variable-inertia twins, and zeroing or negating the commands leaves the forecasts nearly unchanged although the true motion vanishes or reverses. Cross-mechanism histories raise the median amplitude from 0.19 to 0.60 of the true one while increasing the error: the model takes up motion from the evidence, but not the motion that the law implies. In form space, no forecast reaches the closure of the reference videos: the smallest joint gap is 5.6\,mm on B1 and 21.7\,mm on P2, against 2--3.5\,mm for the references (Fig.~\ref{fig:overview}a in the main paper); Section~\ref{sup:e3-form} gives the form verdict on all 2,448 forecasts.

\begin{table}[t]
\centering
\caption{LingBot-VA measurement record by mechanism, 408 forecasts each. Full coverage counts forecasts that pass every solver and visibility check. Floors are medians of the propagated development errors; the residual ratio is the median actuator correction over its floor; the amplitude is the median ratio of predicted to true motion without history.}
\label{tab:lingbot-detail}
\footnotesize
\setlength{\tabcolsep}{3.5pt}
\begin{tabular}{@{}lrrrrr@{}}
\toprule
Family & \makecell[r]{Full\\coverage} & \makecell[r]{Pose floor\\(rad)} & \makecell[r]{Effort floor\\(\Nm)} & \makecell[r]{Residual\\ratio} & Amplitude \\
\midrule
A1 & 408/408 & 0.0020 & 0.110 & 48.8 & 0.13 \\
B1 & 404/408 & 0.0031 & 0.112 & 42.6 & 0.18 \\
P1 & 405/408 & 0.0041 & 0.108 & 20.7 & 1.09 \\
P2 & 372/408 & 0.0015 & 0.066 & 80.2 & 0.11 \\
R1 & 404/408 & 0.0124 & 0.183 & 13.2 & 0.62 \\
R2 & 368/408 & 0.0013 & 0.053 & 106.5 & 0.16 \\
\bottomrule
\end{tabular}
\end{table}

Relative to no history, trajectory error rises by 0.010 [0.009, 0.012]\,rad with same-mechanism history of the same law, 0.014 [0.011, 0.019]\,rad with same-mechanism history of the other law (432 further forecasts under the same protocol), 0.059 [0.054, 0.063]\,rad with correct cross-mechanism history, 0.075 [0.065, 0.083]\,rad with wrong-physics history, and 0.031 [0.026, 0.036]\,rad with a static image of the twin (95\% intervals over six laws; 432 pairs each). Correct cross-mechanism history beats wrong-physics history by 0.0165 [0.0065, 0.0265]\,rad, positively in five of six laws, but trails same-mechanism history by 0.048 [0.044, 0.052]\,rad and static history by 0.028 [0.018, 0.036]\,rad in all six. Of the 720 density pairs, 540 have reference changes at least three times the combined pose floor; their response gain is $-0.142$ [$-0.248$, $-0.042$], and 5.4\% of them fall within [0.5, 1.5]. With histories, the 0.022\,rad gap between twin forecasts grows to 0.061--0.183\,rad. Structures fitted to two of the three command branches predict the third with a median effort error of 1.14\,\Nm{} over 423 held-out folds (9 folds and 6 shared fits do not converge), while the same explainer reproduces native data to $9.1\times10^{-5}$\,\Nm{}. The median relative deviation of the shared parameters from the true law is 1.00, and for A1 and B1 the shared fit sets the gravity coefficient to zero in 67\% and 71\% of cases. The 71 cross-mechanism shared fits with a residual have a median effort residual of 0.99\,\Nm{}.

\paragraph*{Form verdict}\label{sup:e3-form}
The verdict is applied without re-fitting to the frozen confirmation readout of all 2,448 LingBot-VA forecasts (40 frames at 10\,Hz, the given first frame excluded), with the thresholds of Table~\ref{tab:formfaults} (left). Results: A1 408 of 408 broken; B1 408 of 408; P1 227 broken, 2 intact, 179 undecidable; P2 407 broken, 1 undecidable; R1 339 broken, 50 intact, 19 undecidable; R2 301 broken, 76 intact, 31 undecidable. The most frequent broken joints are the ground pivots of the B1 pinion and rack carriage (407 forecasts each), the ground pivot of A1's arm0 (402), and the upper pins of P2 (338--363). Of the 128 intact forecasts, 127 have no history (67 nominal, 30 zeroed, 30 negated commands) and one has the target's own history. Median effort residual over its floor: 11.5 for intact forecasts (R1 7.0, R2 52.1, P1 4.6), 48.3 for broken, 21.5 for undecidable; 9 of the 128 intact forecasts lie within three times their floor.

\subsection{E3: Video World Models}\label{sup:e3-models}
BWM~\cite{S:bwm2026}, Ctrl-World~\cite{S:ctrlworld2025}, and Cosmos3-Nano~\cite{S:cosmos3_2026} receive the confirmation queries of LingBot-VA (laws 2000--2015 at three-law spacing, the six twin families, densities 0.7 and 1.4, servo and impedance commands) zero-shot through their released rollouts. Each receives the query's first frame and the commanded joint positions of the query, through an explicit adapter: the commands, normalized with the frozen quantiles of LingBot-VA, drive one or two rotation channels of the model's end-effector action, and every other channel stays at the center of its range; no model receives efforts, masses, or states.
\emph{BWM} (step 12,000 on Wan2.2-TI2V-5B) predicts 57-frame windows at 30\,Hz with the first frame and the latest eight frames as history (50 steps, guidance 1); the frame is padded white to $3{:}4$, the color of the scene background, and resized to $672\times896$; the commands drive the right-arm roll and pitch of its 14-D action, both grippers open. One window gives 1.6\,s; the forecasts of Fig.~\ref{fig:video}a in the main paper run two windows (3.2\,s). Evidence replaces the eight history frames of the first window by frames of the evidence clip over its first two seconds, with its commands.
\emph{Ctrl-World} (checkpoint 10,000 on Stable Video Diffusion) predicts five frames at 5\,Hz per interaction from the current frame and six history latents (50 steps, guidance 1, empty instruction); four interactions give 3.2\,s. The frame, padded to $5{:}3$ and resized to $192\times320$, fills its three camera slots, and the commands drive the roll and pitch of its 7-D state. Evidence fills the history buffer with the evidence clip at 5\,Hz over two seconds.
\emph{Cosmos3-Nano} runs forward dynamics in Diffusers with the single-arm Franka embodiment (10-D normalized actions, resolution tier 480, third-person view, 30 steps, guidance 1, flow shift 10, one neutral prompt for every query); two chunks of 16 actions at 10\,Hz give 3.2\,s. It conditions on one frame and has no history interface.
Every forecast is read by the frozen readout of LingBot-VA with the same per-family settings, on every saved frame for Ctrl-World and Cosmos3-Nano and at 10\,Hz for BWM; floors are the development pose-error traces propagated through each target on the forecast's own clock, and the form verdict uses the thresholds of Table~\ref{tab:formfaults}. Table~\ref{tab:models} summarizes the four models.

\begin{table*}[t]
\centering
\caption{Native video models evaluated by FoLaSim. Forecasts read; form verdict (intact/broken/undecidable); median motion amplitude relative to the truth under nominal commands without history; median effort residual over its floor; response gain to the density change (resolved pairs, mean of law means, law-cluster bootstrap interval; ideal 1); error under evidence relative to no evidence, own form same law / own other / twin same / twin other.}
\label{tab:models}
\footnotesize
\begin{tabular}{@{}lrcccc c@{}}
\toprule
Model & Forecasts & Form & Amplitude & Effort/floor & Gain & Evidence \\
\midrule
LingBot-VA & 2{,}448 & 128/2090/230 & 0.19 & 45 & $-0.14$ [$-0.25$, $-0.04$] & 1.10 / 1.15 / 1.63 / 1.79 \\
BWM & 504 & 439/53/12 & 0.012 & 56 & $-0.009$ [$-0.019$, $-0.002$] & 1.11 / 1.11 / 3.10 / 2.79 \\
Ctrl-World & 216 & 0/202/14 & 2.26 & 50 & $-0.11$ [$-0.58$, $0.35$] & 0.96 / 1.09 / 1.52 / 1.59 \\
Cosmos3-Nano & 216 & 9/207/0 & 1.45 & 46 & $-0.26$ [$-1.04$, $0.47$] & -- / -- / -- / -- \\
\bottomrule
\end{tabular}
\end{table*}

\subsection{E4: Decision Making}\label{sup:e4-decisions}
The planning task uses the coupled class: sixteen laws, R2/P2, and three densities give 96 assets. A fixed seed draws 64 constant two-dimensional efforts uniformly in $[-1.5,1.5]^2$, replacing the first by zero effort. Every one of the 6,144 asset--candidate pairs is replayed in the forward half at horizon $T=1.6$\,s. Table~\ref{tab:planning} gives the coverage and costs of the study for R2 and P2 separately. The targets are $(0.2,-0.15)$, $(-0.2,0.15)$, and $(0.1,0.25)$ in public rotational coordinates.

The declared task cost and regret are
\begin{align}
J(u;g)&=\|q_T(u)-g\|_2^2+0.01T\|u\|_2^2,\\
\mathcal R&=J(u_{\rm selected};g)-\min_{u\in\mathcal U}J(u;g).
\end{align}
The model selects the lowest predicted cost among finite, in-domain candidate predictions. If every prediction is invalid, the specified fallback is the zero-effort candidate. Actual selected cost and oracle cost always use native replay, not the model's forecast. Every model sees the same action array and targets. Nine coupled state models (three architectures, three seeds) use correct source, wrong source, and empty history, producing 7,776 decisions. R2 source histories support both R2 and P2 queries; only P2 is a cross-mechanism transfer test.

\begin{table*}[t]
\centering
\caption{Expanded planning coverage and costs, separately for source R2 and target P2. Each row averages sixteen laws, three densities, three targets, and three training seeds (432 decisions). History refers to the fixed R2 source. Invalid candidates are excluded from model selection; all-invalid cases would use the declared zero-input fallback.}
\label{tab:planning}
\footnotesize
\begin{tabularx}{\textwidth}{@{}L{1cm}L{2.9cm}L{1.1cm}YYYY@{}}
\toprule
Family & Predictor & History & Regret & Actual cost & Usable / 64 & All-invalid fallbacks \\
\midrule
R2 & MLP & Correct & 0.0971 & 0.1262 & 62.0000 & 0/432 \\
R2 & MLP & Wrong & 0.0971 & 0.1262 & 62.0000 & 0/432 \\
R2 & MLP & Empty & 0.0971 & 0.1262 & 62.0000 & 0/432 \\
R2 & Context GRU & Correct & $3.20\!\times\!10^{-4}$ & 0.0294 & 63.7778 & 0/432 \\
R2 & Context GRU & Wrong & 0.0182 & 0.0473 & 63.7778 & 0/432 \\
R2 & Context GRU & Empty & 0.0048 & 0.0338 & 64.0000 & 0/432 \\
R2 & Mechanical context & Correct & $2.29\!\times\!10^{-5}$ & 0.0291 & 64.0000 & 0/432 \\
R2 & Mechanical context & Wrong & 0.0173 & 0.0464 & 64.0000 & 0/432 \\
R2 & Mechanical context & Empty & 0.0045 & 0.0336 & 64.0000 & 0/432 \\
P2 & MLP & Correct & 0.0941 & 0.1231 & 63.3333 & 0/432 \\
P2 & MLP & Wrong & 0.0941 & 0.1231 & 63.3333 & 0/432 \\
P2 & MLP & Empty & 0.0941 & 0.1231 & 63.3333 & 0/432 \\
P2 & Context GRU & Correct & $3.21\!\times\!10^{-4}$ & 0.0294 & 63.8333 & 0/432 \\
P2 & Context GRU & Wrong & 0.0193 & 0.0483 & 63.8333 & 0/432 \\
P2 & Context GRU & Empty & 0.0053 & 0.0343 & 64.0000 & 0/432 \\
P2 & Mechanical context & Correct & $4.07\!\times\!10^{-5}$ & 0.0291 & 64.0000 & 0/432 \\
P2 & Mechanical context & Wrong & 0.0174 & 0.0464 & 64.0000 & 0/432 \\
P2 & Mechanical context & Empty & 0.0070 & 0.0361 & 64.0000 & 0/432 \\
\bottomrule
\end{tabularx}
\end{table*}

On P2, mean oracle and zero-effort costs are 0.02903 and 0.06918; mean candidate cost range is 0.35860. Thus the mean regret of 0.09410 for the history-free MLP corresponds to an actual cost worse than the zero-effort baseline. Correct-history GRU and mechanical-context costs are 0.02935 and 0.02907. The history-free MLP gives identical choices across histories, providing an interface control. No all-invalid fallback occurs in any of the 7,776 decisions, although individual candidates can be invalid.

The Gauss/regret association first averages rollout residual over its four probes and regret over three targets within each model/seed, law, mechanism, density, and context. It then computes a Spearman correlation within each law over 162 such cases. The mean is 0.7396, with a 16-law bootstrap interval [0.7298, 0.7497]. The residual is computed on the four-second diagnostic trajectories rather than on the candidate planner trajectories, and the correlation ranks the cases within each law by physical error and decision quality.

\subsection{E4: Training on Simulated Corpora}\label{sup:e4-training}
\paragraph*{Corpora} The forward half generates the coupled class for training laws 0--23, validation laws 24--27, and test laws 2000--2015: P2 is read from its inventory and R2 and T2 are synthesized from it. Every episode runs one of ten programs for 4\,s and is rendered at 96\,px and 10\,Hz with the effort on both actuators; all 4,040 episodes pass the episode gate. Training and validation laws carry P2 at density 1.0 and R2 and T2 at 0.7, 1.0, and 1.4; the changed densities of P2 never appear. The no-twin corpus replaces the density changes of R2 and T2 by appearance variants of the density-1.0 assets (random part colors and floor), so every corpus has seven assets and seventy episodes per law and 1,008 training clips.

\paragraph*{Training} All four corpora train the RGB context model of E3 (2.1\,M parameters) for 12,000 steps at batch 32 with five seeds (11, 23, 47, 67, 89) and select the checkpoint by validation loss. Evidence is drawn as in E3: 10\% none, otherwise the same asset; in the form--law twin corpora, half of the remaining draws take a law twin at the same density. The no-effort corpus zeroes the efforts of the evidence episodes in training and test. The query efforts are always given.

\paragraph*{Test} Queries are the chirp and work programs of P2 at densities 0.7 and 1.4 on the sixteen test laws, with evidence from P2 or R2 at the same density, the other changed density, or density 1.0, or none. The score is the motion-region error over the first second, averaged within each law over probes and densities and divided by the error without evidence; intervals bootstrap the sixteen laws after averaging the five seeds. Table~\ref{tab:e4main} in the main paper gives the results.

\subsection{E5: Arm Response Prediction and Online Cost}\label{sup:so101-efficiency}

\textbf{The corpus exposes inertial and actuator variation through recorded position response.} We use FoLaSim's six-joint SO-101 rigid-body model with a fixed base and disabled contacts. The HX-30HM actuator model combines the manufacturer's voltage, torque, speed, and encoder specifications~\cite{S:hiwonderhx30hm} with randomized response priors. Its nominal stall torque is 2.942\,Nm and its no-load speed is 5.512\,rad/s at 11.1\,V. A DC motor approximation includes back-EMF, duty saturation, current-limited braking, command filtering, communication delay, deadband, rotor armature, and viscous and smooth dry friction. Position observations and transmitted goals are quantized to 4,096 counts per revolution. The E5 corpus and evaluations use simulated trajectories, with specification-based servo gains and response priors. Section~\ref{sup:so101-real} evaluates the trained predictor on physical HX-30HM hardware.

Training comprises 96 configurations, each with 24 six-second motion programs and six disjoint support programs. Eight further configurations supply validation queries for checkpoint selection. Link density factors range from 0.85 to 1.18, payload mass from 10 to 150\,g, and payload center-of-mass offsets from $-10$ to 10\,mm per axis. The nominal gains are $K_p=84.28$\,Nm/rad and $K_v=0.8$\,Nm\,s/rad, varied by factors 0.5--2.0 and 0.5--1.5, respectively. Other variations include supply voltage 9.0--12.6\,V, command-filter time 3--30\,ms, delay 1--8\,ms, and deadband 0--3 counts. The torque-speed relation and these response ranges define the simulation family.

The test set contains 16 unseen configurations from this family. Eight OOD configurations use payloads of 170--190\,g, beyond the training range. Sixteen Mass-test configurations reuse the Test inertials with the nominal servo, isolating inertial generalization. Each configuration provides eight chirp queries and six decision tasks with four candidate command durations per task. Query chirps differ from the stationary sinusoidal training and support programs. The same query programs are used across configurations. Each predictor receives four initial position samples and the known command sequence at 30\,Hz. The evaluator reads future measured positions after prediction. Three six-second support episodes give the 18\,s adaptation budget.

\textbf{A compact context model turns the generated trajectories into a deployable predictor.} The response model contains 242,822 parameters. A GRU with 64 hidden units encodes each support episode at 15\,Hz using positions, finite-difference velocities, and command errors. Averaging the episode contexts yields one representation of the support set. A learned initialization combines that context with the four observed position-command pairs. A two-layer GRU with 128 hidden units per layer then reads the future commands, their increments, the initial position, and the context. A small readout predicts the residual to the future position command. The entire 90-frame forecast uses the known command sequence and the initial observations.

We train for 8,000 AdamW updates at batch size 256 with learning rate $10^{-3}$, cosine decay, and gradient clipping at 1. The loss combines position error and a temporal-difference term. Support context is dropped with probability 0.2 during training, and the checkpoint with minimum validation RMSE is selected independently for seeds 11, 23, and 47. At deployment, weights are frozen and adaptation consists of encoding the supplied support episodes. Table~\ref{tab:so101-response-ablation} compares the full corpus with nominal dynamics, inertial variation alone, and a separately trained no-history model. The architecture, programs, seeds, and update budget are shared across these learned variants.

\begin{table}[t]
\centering
\caption{Response-model training ablations. All learned variants use the same architecture, update budget, and three seeds. Nominal training fixes all simulated dynamics; inertial variation fixes the nominal servo; full response variation changes both. The no-history variant receives no support context. Errors are 3\,s joint RMSE in degrees.}
\label{tab:so101-response-ablation}
\footnotesize
\setlength{\tabcolsep}{4pt}
\begin{tabular}{@{}lrrr@{}}
\toprule
Training / input & Test & OOD & Mass-test \\
\midrule
Command as prediction & 1.7856 & 1.8376 & 1.7745 \\
Nominal training & 0.3552 & 0.2830 & 0.0583 \\
Inertial variation & 0.3339 & 0.2558 & \textbf{0.0342} \\
Without support history & 0.2886 & 0.3182 & 0.1620 \\
Full response variation & \textbf{0.0633} & \textbf{0.0850} & 0.0425 \\
\bottomrule
\end{tabular}
\end{table}

\textbf{Evidence from another law tests whether the predictor follows the law.} Within each split, the query and support commands are identical across configurations, so the configurations share the form of the arm and differ only in its law. We therefore repeat the evaluation with the support episodes of the next configuration of the split and without support. We also compute the adjustment gain $g=\sum\langle\hat q_{k+1}-\hat q_k,\,q_{k+1}-q_k\rangle/\sum\lVert q_{k+1}-q_k\rVert^2$ over programs, frames, and joints 1--5, where each forecast $\hat q_k$ uses the support of its own configuration (Table~\ref{tab:so101-formlaw}). Full response variation yields a model that reads the law from its support: other-law support is worse than none on Test and OOD, and the gain approaches one. On Mass-test, all configurations share the nominal servo, so other-law support still carries the common servo response and helps. Nominal training yields a model whose forecast does not depend on its support, with a gain of zero. Inertial variation lies between the two and comes closest to full variation on Mass-test, where only the inertia changes. The no-history model receives no support, and its gain comes from the four initial samples alone.

\begin{table}[t]
\centering
\caption{Does the response model follow the law? Within each split, the configurations share the arm and the commands and differ only in the law. Columns 3--5: 3\,s joint RMSE in degrees with 18\,s of support from the same configuration (same law), from the next configuration of the split (other law), or without support. Gain: the predicted change of motion between consecutive configurations projected on the true change (1: the forecast follows the law; 0: it ignores it); 95\% bootstrap intervals over Test configurations: full variation 0.85--0.94, inertial 0.08--0.30, nominal 0.00--0.00, no history $-$0.04-$-$0.50. Means over three seeds and the configurations of a split; Mass: Mass-test.}
\label{tab:so101-formlaw}
\footnotesize
\setlength{\tabcolsep}{4pt}
\begin{tabular}{@{}llrrrr@{}}
\toprule
Training & Split & Same law & Other law & None & Gain \\
\midrule
Full var. & Test & 0.063 & 0.477 & 0.273 & 0.90 \\
 & OOD & 0.085 & 0.382 & 0.320 & 0.82 \\
 & Mass & 0.042 & 0.085 & 0.141 & 0.64 \\
\addlinespace[2pt]
Inertial & Test & 0.334 & 0.378 & 0.355 & 0.18 \\
 & OOD & 0.256 & 0.280 & 0.282 & 0.17 \\
 & Mass & 0.034 & 0.082 & 0.057 & 0.65 \\
\addlinespace[2pt]
Nominal & Test & 0.355 & 0.355 & 0.355 & 0.00 \\
 & OOD & 0.283 & 0.283 & 0.283 & 0.00 \\
 & Mass & 0.058 & 0.058 & 0.059 & 0.00 \\
\addlinespace[2pt]
No history & Test & 0.289 & 0.289 & 0.289 & 0.23 \\
 & OOD & 0.318 & 0.318 & 0.318 & 0.14 \\
 & Mass & 0.162 & 0.162 & 0.162 & 0.46 \\
\bottomrule
\end{tabular}
\end{table}

\textbf{The physics baselines pair native solvers with support-set identification.} Newton 1.8.0~\cite{S:newton2026} uses its Featherstone solver with semi-implicit Euler integration, Genesis 1.4.6~\cite{S:genesis2026} uses its native rigid solver with implicitfast integration, and Isaac Sim 6.2.0.0~\cite{S:isaacsim2026} uses native GPU PhysX. All three import the nominal SO-101 model and use the same explicit HX-30HM actuator parameterization. A 23-parameter support-only calibration adjusts gains, damping, friction, armature, filtering, delay, deadband, supply, torque limits, and mass scale. Each fit runs 160 CMA-ES generations with 24 candidates. Every held-out configuration is calibrated independently for each of the three seeds. True actuator efforts, private inertial parameters, and held-out model descriptions are excluded from the input.

Isaac Sim anchors the articulation through a world-fixed joint. Native joint drives are disabled, and the explicit actuator supplies the applied effort. Each rollout restores the supplied joint positions and velocities, zeroes root velocity and residual effort, and refreshes kinematics before integration. Body damping and sleeping are disabled. The 120 calibrated cases produce 3,840 query and candidate-action forecasts and 720 decisions. For evaluation, we group their independent environments into one batch of 3,840. Comparison with separate 32-trajectory rollouts gives a maximum pointwise difference of $3.82\times10^{-6}$\,rad and a maximum per-trajectory RMS difference of $1.09\times10^{-6}$\,rad, respectively 0.249\% and 0.071\% of one encoder count. Online timing uses the B1 and B32 workloads independently.

TrajID-Param~\cite{S:kovalev2026trajid} uses the released differentiable MJX rollout and PD parameterization, extended to the coupled six-joint plant. Its 18 learned scalars are the six position gains, velocity gains, and armatures. We fit a position-only trajectory objective on three-observation support windows using Adam at learning rate 0.01 and a maximum of 3,000 updates, with the original support-loss checkpoint selection. All 120 fits reached the update cap. The robot adaptation uses nominal HX-30HM torque limits and velocities estimated from encoder samples. The main table reports FP32 deployment of these FP64-identified parameters. A full precision comparison over all 120 cases gives a maximum FP32--FP64 prediction difference of $4.08\times10^{-6}$\,rad and identical choices in all 720 decision tasks. The three split RMSEs change by less than $2\times10^{-8}$ degrees. FP64 B1/B32 latencies are 609.873/666.794\,ms, compared with 550.284/638.755\,ms for FP32, on the same A800 in separate measurement windows.

\textbf{Accuracy and online cost are evaluated at their respective workload scales.} Three-second prediction RMSE scores the five moving arm joints. The gripper remains part of the six-joint dynamics and prediction, with a fixed command. Errors are averaged over eight queries and three seeds within each configuration, then over configurations within a split. The main table's overall mean weights Test, OOD, and Mass-test by 16, eight, and 16 configurations. Confidence intervals in Table~\ref{tab:so101-uncertainty} bootstrap configurations 2,000 times within each split. TrajID-Param appears once in its FP32 deployment.

\begin{table*}[t]
\centering
\caption{Uncertainty and resource use for the SO-101 response study. Intervals resample configurations 2,000 times within each split after averaging query programs and seeds. Power is total GPU board energy divided by measurement time across the three B32 workloads. Memory is the largest observed board allocation, sampled every 50\,ms, including runtime contexts and caches.}
\label{tab:so101-uncertainty}
\footnotesize
\setlength{\tabcolsep}{3.5pt}
\renewcommand{\arraystretch}{1.2}
\begin{tabular*}{\textwidth}{@{\extracolsep{\fill}}lcccrr@{}}
\toprule
 & \multicolumn{3}{c}{RMSE (${}^\circ$), 95\% confidence interval} & \multicolumn{2}{c}{B32 GPU resources} \\
\cmidrule(lr){2-4}\cmidrule(l){5-6}
Method & Test & OOD & Mass-test & Power (W) & Memory (MiB) \\
\midrule
TrajID-Param~\cite{S:kovalev2026trajid} & 0.1361 [0.1080, 0.1711] & 0.1316 [0.0926, 0.1746] & 0.0886 [0.0845, 0.0930] & 93.90 & 1367.1 \\
Newton 1.8 + calibration~\cite{S:newton2026} & 0.0307 [0.0286, 0.0331] & 0.0386 [0.0338, 0.0437] & 0.0291 [0.0281, 0.0300] & 92.90 & 1289.1 \\
Genesis 1.4.6 + calibration~\cite{S:genesis2026} & 0.0293 [0.0278, 0.0309] & 0.0378 [0.0333, 0.0429] & 0.0289 [0.0280, 0.0298] & 93.60 & 2344.3 \\
Isaac Sim 6.2 + calibration~\cite{S:isaacsim2026} & 0.0294 [0.0280, 0.0309] & 0.0379 [0.0333, 0.0432] & 0.0289 [0.0280, 0.0298] & 88.00 & 2061.3 \\
\textbf{FoLaSim response model} & 0.0633 [0.0550, 0.0714] & 0.0850 [0.0714, 0.0986] & 0.0425 [0.0414, 0.0435] & 106.07 & 1331.1 \\
\bottomrule
\end{tabular*}
\end{table*}

\textbf{Online measurements isolate the cost of a complete forecast call.} We measure every method on NVIDIA A800-SXM4-80GB GPUs with CPU thread settings fixed at four. The workloads are the first configuration in each split, seed 23, with one query trajectory (B1) or all eight queries and 24 decision candidates (B32). Two calls warm each workload before measurement over at least three seconds and at least three complete calls. We report the median of the three per-workload latency medians. Every call starts from CPU arrays and ends with a CPU NumPy forecast, including transfer and synchronization. Our support history is encoded on every call and shared among trajectories within that batch. Model loading, offline training, parameter fitting, and JIT compilation precede timing. Table~\ref{tab:so101-efficiency} normalizes B32 costs by FoLaSim.

All physics baselines integrate at 1,200\,Hz, giving 40 substeps per recorded frame and 3,600 substeps per forecast. Newton captures each observation interval as a CUDA Graph, and TrajID uses JIT-compiled batched MJX prediction. Genesis and Isaac Sim apply the explicit servo and advance their native scenes at each substep with visualization disabled. The shared integration rate exposes the cost of these prediction interfaces at a fixed temporal resolution. The learned model's accuracy uses its original three-seed FP32 evaluation. Deployment timing additionally disables TF32. TrajID's FP32 deployment is reevaluated on all 40 configurations and three seeds.

GPU energy is the difference in NVML cumulative board-energy counters over the measurement window, divided by the number of calls. It includes idle board draw and excludes CPU, host, and offline-training energy. Table~\ref{tab:so101-uncertainty} separately reports average board power and sampled peak memory. Offline learning and per-configuration identification have different roles: the response model learns across the training family, whereas the physical baselines fit the current support set. The online comparison begins after these stages and includes the support encoding needed by our predictor for each new call.

\subsection{E6: Sim-to-Real Prediction and Action Selection}\label{sup:so101-real}

\subsubsection{Hardware, calibration, and frozen deployment}
The physical platform is an SO-ARM101 with six HX-30HM servos connected to a local RTX~4060 laptop. Manual reference-pose and direction checks map the 4,096-count encoder coordinates to the simulation joint coordinates, including the wrist-roll offset. Position commands and encoder reads are scheduled at 30\,Hz. Voltage and temperature are monitored during execution. Prediction uses joint positions and position commands; two RGB-D cameras record the external motion. The evaluation measures joint-space prediction and reaching, using the first five joints while the gripper command remains fixed.

We deploy the full-response-variation checkpoint with seed 23, selected by the lowest simulation validation RMSE among its three training seeds. The weights remain fixed throughout all 33 physical executions. Before each program, the arm holds its starting target for 0.5\,s. The last four observed position-command pairs initialize the predictor, which receives the next 90 commands and generates the complete 3\,s forecast before movement. The support set is empty, so the context input is zero. This setting uses the support dropout already present during simulation training (Section~\ref{sup:so101-efficiency}). Subsequent encoder readings score the forecast without entering its inputs. Coordinate calibration establishes the interface; the experiment performs no real-data weight updates or servo-parameter fitting.

\subsubsection{Prediction across attachments and motion scales}
\textbf{The same programs test transfer across three physical attachments.} We retain one fixture and evaluate it empty, with a 6\,g block, and with a 47\,g wrench. These operator-reported object masses exclude the fixture; its total mass and the mounted centers of mass were not independently measured. Each condition executes the same eight six-second programs once. The first five reference coordinates follow chirps with a $\sin^2(\pi t/6)$ envelope, amplitudes drawn from 0.55--1 times 36 encoder counts, frequencies from 0.16--0.30\,Hz, and phases from $[-\pi,\pi]$. Their phase evolves as $2\pi(f_jt+0.015t^2)+\phi_j$. Rounding gives integer commands within $\pm3.164^\circ$ of the starting target, with the first and last commands at that target. The archived generator fixes the eight programs before execution.

Each joint's encoder samples are linearly interpolated using its own host read-completion times onto the forecast clock. For a horizon of $h$ seconds, the per-execution error is
\begin{equation}
E_h=\frac{180}{\pi}\sqrt{\frac{1}{5N_h}\sum_{k=0}^{N_h-1}\sum_{j=1}^{5}
\bigl(\hat q_{k,j}-q_{k,j}\bigr)^2},\qquad N_h=30h.
\label{eq:so101-real-error}
\end{equation}
Table~\ref{tab:so101-real-horizons} averages these errors over the eight programs in each attachment condition. The horizons use prefixes of one forecast, rather than repeated predictions as new observations arrive.

\begin{table}[t]
\centering
\caption{Frozen deployment across attachments. Joint RMSE in degrees is averaged over eight physical programs per condition. All entries use the deployed full-variation model, seed 23. Each column scores the corresponding prefix of the same 3\,s forecast; mass labels exclude the common fixture.}
\label{tab:so101-real-horizons}
\footnotesize
\setlength{\tabcolsep}{4pt}
\begin{tabular}{@{}lrrrr@{}}
\toprule
Attachment & 0.5\,s & 1\,s & 2\,s & 3\,s \\
\midrule
Empty fixture & 0.334 & 0.370 & 0.540 & 0.690 \\
6\,g block & 0.325 & 0.373 & 0.563 & 0.687 \\
47\,g wrench & 0.321 & 0.377 & 0.555 & 0.711 \\
\bottomrule
\end{tabular}
\end{table}

\textbf{Variation of actuator response strengthens transfer beyond inertial variation alone.} We replay the nominal, inertial-variation, full-response-variation, and no-history models from Section~\ref{sup:so101-efficiency} on identical archived initial observations and future commands. Each variant uses its three simulation-trained seeds, with zero support episodes and fixed weights. Across the 24 small-motion executions, full response variation lowers mean error relative to nominal training, while inertial variation alone changes it little. The same ordering holds across the nine larger reaches (Table~\ref{tab:so101-real-transfer}). This comparison links transfer to the range of actuator responses represented during pretraining. The separately trained no-history variant has greater variation across seeds, although individual seeds remain competitive.

The command-as-prediction reference returns the future position command, while the constant-state reference repeats the last observed position. Small physical displacements make the latter competitive with the deployed model. Larger reaches distinguish them more strongly because the initial position ceases to approximate the subsequent trajectory. Table~\ref{tab:so101-real-transfer} reports both references alongside the single deployed checkpoint and the three-seed pretraining comparisons. These comparisons use the same recorded executions; their seed variation measures model-training variation rather than repeated hardware trials. The Newton, Genesis, Isaac Sim, and TrajID-Param comparisons in E5 use the simulation corpus.

\begin{table}[t]
\centering
\caption{Simulation pretraining evaluated on real motion. Three-second RMSE in degrees uses joints 1--5 and identical recorded inputs for every model. Upper block: mean $\pm$ sample SD across three training-seed means, each averaged over the indicated executions. Lower block: the single deployed model and two deterministic references. Bold marks the lowest mean within each block. Training seeds reuse the same physical recordings.}
\label{tab:so101-real-transfer}
\footnotesize
\setlength{\tabcolsep}{4pt}
\begin{tabular}{@{}lrr@{}}
\toprule
Model / reference & Small motions & Reaches \\
 & (24 runs) & (9 runs) \\
\midrule
Nominal & $1.081\pm0.003$ & $0.716\pm0.006$ \\
Inertial variation & $1.077\pm0.003$ & $0.713\pm0.004$ \\
Without support history & $0.727\pm0.059$ & $0.296\pm0.095$ \\
Full response variation & $\mathbf{0.708}\pm0.015$ & $\mathbf{0.241}\pm0.012$ \\
\midrule
Deployed model (seed 23) & 0.696 & \textbf{0.232} \\
Command as prediction & 1.362 & 1.049 \\
Constant-state forecast & \textbf{0.674} & 3.055 \\
\bottomrule
\end{tabular}
\end{table}

\subsubsection{Where the payload's law appears}\label{sup:so101-real-effort}
\textbf{The servos record the effort that the encoders do not see.} Every encoder read also returns the present load of the servo (register 0x3C), i.e., the output torque as a fraction of rated torque in sign-magnitude form with bit 10 as the sign, as in the manufacturer's table~\cite{S:hiwonderhx30hm}. For each program and joint, we average the signed load over the 6\,s program, and we compare the attachments program by program, since all three run the same commands. The joint angles are aligned by command frame, and their RMS difference over the program measures the change of motion. The 6\,g block changes the law little, so its differences from the empty fixture give the repeatability of both channels. Table~\ref{tab:so101-real-effort} reports the paired differences with 95\% $t$-intervals over the eight programs.

\begin{table}[t]
\centering
\caption{Where the payload's law appears on the real arm. Each entry is the mean over the eight programs of the paired difference from the empty fixture under identical commands, with a 95\% $t$-interval. Motion: RMS difference of the measured angles of joints 1--5 over the program. Effort: change of the mean present load over the program; the loads are negative while the joints hold the arm against gravity, so a negative change is a larger supporting effort. Forecast: change of the frozen model's 3\,s RMSE.}
\label{tab:so101-real-effort}
\footnotesize
\setlength{\tabcolsep}{3pt}
\begin{tabular}{@{}lll@{}}
\toprule
Channel & 6\,g block & 47\,g wrench \\
\midrule
Motion (deg RMS) & 0.11 [0.08, 0.14] & 0.33 [0.27, 0.40] \\
\addlinespace[2pt]
\multicolumn{3}{@{}l}{Effort (\% of rated torque)} \\
\quad base yaw & $+$0.24 [$-$0.06, $+$0.53] & $-$0.08 [$-$0.31, $+$0.16] \\
\quad shoulder & $-$0.20 [$-$0.42, $+$0.03] & $-$1.95 [$-$2.40, $-$1.49] \\
\quad elbow & $-$0.25 [$-$0.50, $-$0.01] & $-$0.93 [$-$2.03, $+$0.17] \\
\quad wrist flex & $-$0.02 [$-$0.30, $+$0.26] & $-$0.82 [$-$1.02, $-$0.61] \\
\quad wrist roll & $+$0.06 [$-$0.15, $+$0.28] & $-$0.04 [$-$0.31, $+$0.23] \\
\addlinespace[2pt]
Forecast (deg) & 0.00 [$-$0.03, $+$0.02] & $+$0.02 [$-$0.03, $+$0.07] \\
\bottomrule
\end{tabular}
\end{table}

\begin{figure*}[t]
\centering
\includegraphics[width=\textwidth]{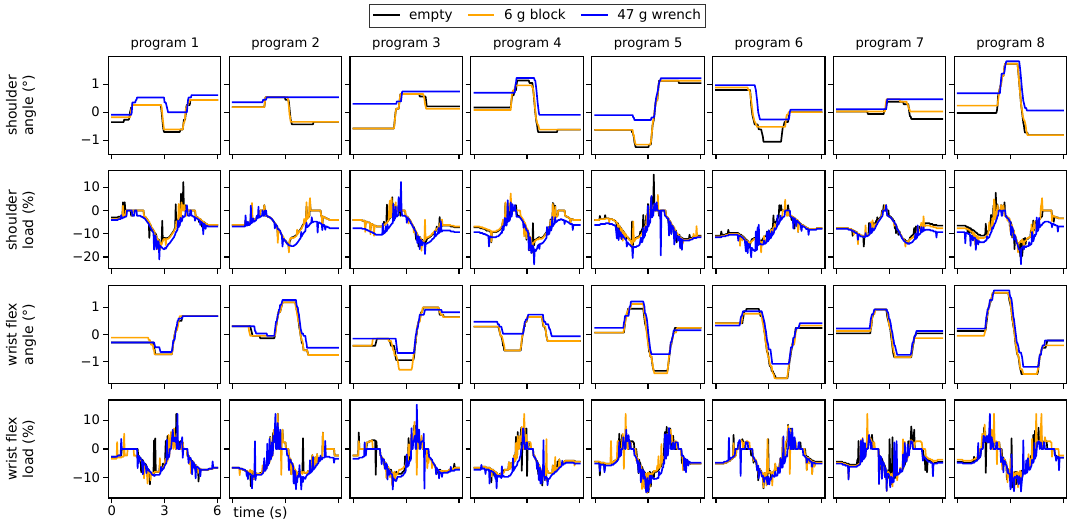}
\caption{Raw servo reads of the eight zero-shot programs on the real SO-ARM101 under identical position commands. Rows: shoulder and wrist-flex angle, relative to the empty arm's mean over the program, and present load in percent of rated torque, at 30\,Hz without filtering. The 6\,g block overlaps the empty arm; the 47\,g wrench shifts the loads and leaves a small static offset in the angles (Table~\ref{tab:so101-real-effort}).}
\label{fig:so101-real-traces}
\end{figure*}

The wrench changes the motion by three times the difference of the block, consistent with a slightly larger static error of the position loop under the added gravity load, yet the change remains below half a degree. In contrast, it changes the loads of the shoulder and the wrist by ten and forty times the difference of the block. The frozen model reads four initial samples that already contain the static offset, so its error does not change. Fig.~\ref{fig:so101-real-traces} shows the raw shoulder and wrist-flex reads of all eight programs. A calibration by the arm's own gravity torque in the FoLaSim model reads the wrench at the shoulder as 44\,g [31, 57], whereas the elbow and the wrist give 23 and 19\,g. Since the fixture and the mounted center of mass were not measured, we do not report a payload identification.

\subsubsection{Forecast-based action selection}\label{sup:so101-real-decisions}
\textbf{The predictor ranks candidate commands before the arm executes them.} With the 47\,g wrench attached, we test base-yaw offsets of $+114$, $-114$, and $+91$ encoder counts, approximately $+10.02^\circ$, $-10.02^\circ$, and $+8.00^\circ$. Other joint targets remain fixed. Each task offers durations $d\in\{1.0,1.4,1.8\}$\,s. The reference starts at 0.3\,s and follows the quintic profile $10z^3-15z^4+6z^5$, where $z=\operatorname{clip}((t-0.3)/d,0,1)$. The arm holds the target after arrival and returns to its starting target from 4.2 to 6.0\,s, beyond the forecast interval.

The selector forecasts all three commands from the same four initial observations and minimizes
\begin{equation}
C(d)=d+20e_\infty+10\mathbf{1}_{\rm fail},
\qquad e_\infty=\max_k|q_{k,1}-r_{k,1}|,
\label{eq:so101-real-cost}
\end{equation}
where $d$ is in seconds and $e_\infty$ is the peak base-yaw tracking error in radians over the 90 forecast frames. Success requires three consecutive frames within 0.03\,rad of the target during the interval from command arrival, $0.3+d$, to 0.5\,s later. Predicted positions determine the selection; interpolated encoder positions determine the measured score. We execute the selected candidate first, then the other two once each. The base-angle spans among the three candidate initial states are $0.352^\circ$, $0.527^\circ$, and $0.527^\circ$ for the three targets, respectively.

\begin{table*}[t]
\centering
\caption{Candidate selection and subsequent real execution with the 47\,g wrench. Predicted costs are those used at the initial selection, before any candidate for that target was executed. Measured costs come from one execution per candidate. Bold marks the selected duration. All nine executions meet the joint-space success criterion in Section~\ref{sup:so101-real-decisions}.}
\label{tab:so101-real-decisions}
\footnotesize
\setlength{\tabcolsep}{4pt}
\begin{tabular}{@{}lrrrrrr@{}}
\toprule
 & \multicolumn{2}{c}{1.0\,s} & \multicolumn{2}{c}{1.4\,s} & \multicolumn{2}{c}{1.8\,s} \\
\cmidrule(lr){2-3}\cmidrule(lr){4-5}\cmidrule(l){6-7}
Base-yaw target & Predicted & Measured & Predicted & Measured & Predicted & Measured \\
\midrule
$+10.02^\circ$ & \textbf{1.5692} & \textbf{1.7911} & 1.8319 & 1.9730 & 2.1615 & 2.2462 \\
$-10.02^\circ$ & \textbf{1.6221} & \textbf{1.7644} & 1.8833 & 1.9731 & 2.2123 & 2.2768 \\
$+8.00^\circ$ & \textbf{1.4880} & \textbf{1.6083} & 1.7889 & 1.8727 & 2.1207 & 2.1807 \\
\bottomrule
\end{tabular}
\end{table*}

All nine executions meet the success criterion. The selector chooses 1.0\,s for every target, which also minimizes measured cost in each three-candidate set (Table~\ref{tab:so101-real-decisions}). The faster trajectories have larger peak tracking errors, but their shorter duration more than offsets this increase under \eqref{eq:so101-real-cost}. Mean measured cost falls from 2.2346 for the fixed 1.8\,s reference to 1.7213 for the selected commands, a reduction of 22.97\%. Regret relative to the measured candidate minimum is zero for each target. Each candidate has one physical execution, and the reported minimum is over that task's three tested commands.

\subsubsection{RGB-D records and reproducible evaluation}
An Astra Pro and a RealSense D415 record RGB and native 16-bit depth at $640\times480$ resolution, with per-frame host timestamps. Both views cover all 33 formal executions. The empty-fixture and 6\,g conditions use separate recording segments. Continuous recording then covers installation of the 47\,g wrench, its prediction tests, all reaching candidates, and unloading. The earlier waits and attachment changes between segments are outside the recordings. Cameras use host-time matching without hardware exposure synchronization or color-depth registration.

Figure~\ref{fig:real}a uses the nearest received frame to each displayed time in the selected positive reach. Depth uses the acquisition preview's Turbo palette, with red for near and blue for far, on a common clipped linear 0.35--1.10\,m scale. Zero depth is displayed as missing. The D415 view is rotated by $180^\circ$ for presentation. Original frames, timestamps, program commands, encoder telemetry, initial observations, and forecasts are retained. These records support recomputation of \eqref{eq:so101-real-error} and \eqref{eq:so101-real-cost}; the raw encoder coordinates provide the numerical measurements.

\section{Extensions}\label{sup:extensions}
In this section, we describe how the construction extends from mechanisms with fixed bilateral constraints to robots on fixed bases and robots in frictional contact, from the three shipped robots to any MJCF or URDF description, wherever the law is known exactly and every twin can be certified. Section~\ref{sup:so101-real} evaluates physical deployment of a learned response model; Section~\ref{sup:robots} discusses the additional measurements needed to anchor an SI-valued law reading on recorded robots.

\subsection{Robots}\label{sup:robots-open}
\begin{figure*}[t]
\centering
\includegraphics[width=\textwidth]{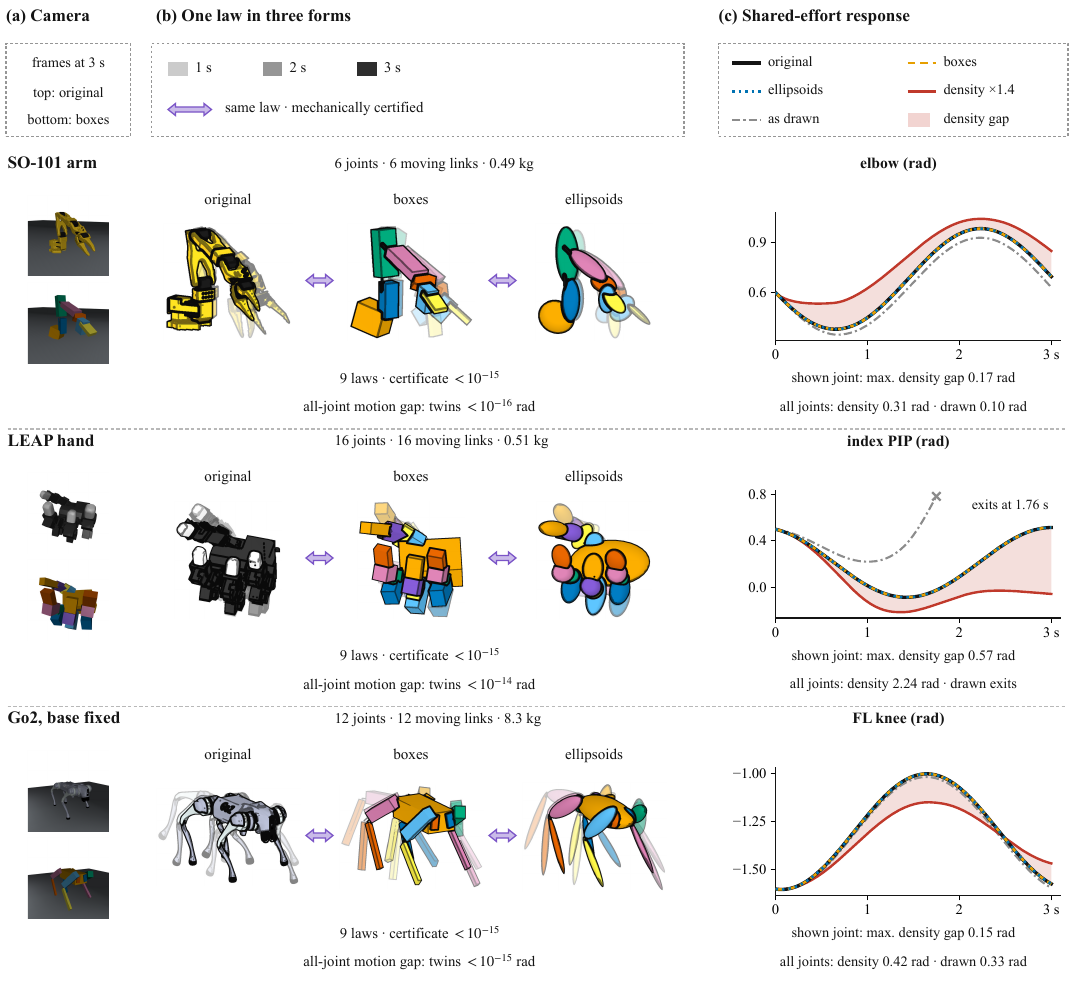}
\caption{The robots on fixed bases: one law in three forms, one form under two laws. Rows: the SO-101 arm, the LEAP hand, and the Unitree Go2 with its base held fixed. From left: the frames a camera records of the original and of its equimomental boxes at 3\,s of one shared open-loop effort program; the original, its boxes, and its ellipsoids, ghosted at 1, 2, and 3\,s, with the certificates of the twins over nine laws of each family; and one joint under the program for the original, both law twins, the density-1.4 form twin (red shading: its difference from the original on the displayed joint), and the links as drawn (cross: the drawn hand leaves its joint domain). The annotations distinguish the maximum gap on the displayed joint from the maximum over all joints.}
\label{fig:robots-fixed}
\end{figure*}

In this subsection, we give the settings of the robot experiments on fixed bases in Sections~\ref{sec:robots} and~\ref{sec:experiments} of the main paper (Fig.~\ref{fig:robots-fixed}); Section~\ref{sup:contact} gives those of the contact tasks.

\paragraph*{Descriptions and laws} The SO-101 arm, LEAP hand, and Unitree Go2 have six, sixteen, and twelve joints, respectively. Their descriptions are vendored without meshes, so laws, twins, certificates, and readings run offline; the meshes of the original appearance are fetched from pinned upstream commits on first render. FoLaSim removes keyframes, actuators, sensors, and contacts, disables joint limits, and fixes a free base at its standing height (0.6\,m for the Go2). On a fixed base the robot thus moves in free space, without contact between its own links or with anything else, so that an effort program exercises its law alone; contact enters with the scenes of Sections~\ref{sup:contact} and~\ref{sup:zoo}. Of the dissipation, the viscous part is MuJoCo's own joint damping, integrated implicitly, and the dry part $f\tanh(v/\varepsilon)$ with $\varepsilon=0.05$\,rad/s is applied by the simulator as generalized forces. Integration steps are 1/1200\,s for the arm and the quadruped and 1/6000\,s for the hand, whose fingers carry no rotor armature. The per-link factors of a robot's law family are drawn log-uniformly within $[1/1.3,\,1.3]$, while armature and dissipation keep their nominal values.

\paragraph*{Twins and certificates} MuJoCo computes the inertia of the equimomental boxes and ellipsoids \eqref{eq:equimomental} from their geometry, not from the target values. The body certificate compares mass, center of mass (relative to the link's radius of gyration), and inertia tensor link by link; the engine certificate compares the mass matrix and bias forces of the twin's engine with those of the law at eight random states and velocities. The bounding box of a link as drawn is axis-aligned in the link frame.

\paragraph*{Programs} \emph{Hold} applies the gravity torque of the home pose; \emph{positive} and \emph{negative} add a Gaussian pulse centered at 0.35\,s with width 0.12\,s, sized to change each joint's velocity by 0.6\,rad/s at the home inertia, with alternating signs along the joints; \emph{reference}~$k$ applies the inverse dynamics of the law along one sinusoid per joint about the home pose, starting at the home pose, with an amplitude of 0.3\,rad (at most 0.8 of the room to the joint range), a frequency of 0.3--0.7\,Hz, and a phase drawn with seed $k$. E1 and E2 run these programs open-loop for 3\,s; the corpora of E3 and E4 run them under impedance tracking of the same reference (the home pose for hold and pulses), whose stiffness and damping on each joint are those of a second-order system with that joint's home inertia, a damping ratio of 0.7, and natural frequencies of 6, 10, and 30\,rad/s for SO-101, Go2, and LEAP, respectively. Gains and feedforward are computed once from the nominal robot and shared across laws and twins.

\paragraph*{Reading} Forecasts are read on 60\,Hz and 10\,Hz clocks with the fixed differentiator of Section~\ref{sec:state} in the main paper. The floor is the 95\% quantile of the effort residual of the exact forecast under 32 draws of i.i.d.\ pose noise, $10^{-5}$\,rad on the state clock and 0.0024\,rad, the readout error of the image readout, on the video clock. Each link's column of the implied law is the change of MuJoCo's inverse dynamics when that link's inertial parameters are doubled, and the report applies the thresholds of Section~\ref{sup:reports} unchanged. The other law of E2 is the first draw of the family, from seed 99 on, that stays within the joint domain under every query program.

\paragraph*{Corpora of E3 and E4} Each robot's family supplies 36 laws (seed 2026): laws 0--19 train, 20--23 validate, and 24--35 test. Every episode lasts 4\,s under one of ten tracked programs computed from the nominal robot: hold, the positive and negative pulses (evidence), references 1--5 (training), and references 101 and 102 (queries). The RGB corpus of the SO-101 arm follows the E4 corpus of P2 asset for asset (Section~\ref{sup:e4-training}), with the original in place of P2 and the boxes and ellipsoids in place of R2 and T2; test laws carry the original and the boxes at all three densities (3,360 episodes, none rejected). A fixed camera renders 96-pixel frames at 10\,Hz, with the realized effort on all six joints. The state corpora of the three robots carry the original and the boxes at the three densities under the same ten programs and, on training and validation laws, thirty further references on the assets that state models train on.

\paragraph*{Models of E3 and E4} The RGB context model, its training (12,000 steps at batch 32, checkpoint by validation loss), its evidence draws, and its evaluation are those of E4, with six effort channels instead of two, under seeds 11, 23, and 47. The state model reads the two evidence episodes (the first 0.8\,s at 30\,Hz: states, velocities, efforts, and the form of the evidence asset) with a GRU of 64 units, and a three-layer network of 256 units predicts the joint accelerations from the current state, the effort, the form of the queried asset, and the context; velocities and positions follow by the trapezoidal rule at 120\,Hz. As in E3, the known impedance controller of each program is closed around the model, so its effort is reconstructed on the predicted states. Training minimizes the state and velocity error of 40-step unrolls (8,000 steps at batch 256, seeds 11, 23, and 47); test rollouts start from the true initial state of each query and are scored by their joint-angle RMS error over 1\,s.

\paragraph*{Results} Tables~\ref{tab:robot-state} and~\ref{tab:robot-rgb} give every seed of the robot experiments of Sections~\ref{sec:e3} and~\ref{sec:e4} in the main paper. Law twins are indistinguishable to a model given canonical coordinates, so the two same-law columns of Table~\ref{tab:robot-state} differ only through the form flag of the evidence.

\begin{table}[t]
\centering
\caption{State models on robots: joint-angle RMS error at 1\,s (mrad) on the original at densities 0.7 and 1.4, twelve held-out laws, without evidence and with evidence from the original (own) or its boxes (twin) under the same or the other density, and the ratio of own other-law to twin same-law error.}
\label{tab:robot-state}
\footnotesize
\setlength{\tabcolsep}{2.2pt}
\begin{tabular}{@{}lccccccc@{}}
\toprule
 & & & \multicolumn{2}{c}{Same law} & \multicolumn{2}{c}{Other law} & \\
\cmidrule(lr){4-5}\cmidrule(lr){6-7}
Robot & Seed & None & Own & Twin & Own & Twin & Ratio \\
\midrule
SO-101 & 11 & 73.5 & 11.4 & 11.6 & 130.2 & 130.0 & 11.2 \\
SO-101 & 23 & 73.9 & 13.9 & 13.6 & 128.4 & 128.6 & 9.4 \\
SO-101 & 47 & 74.4 & 14.2 & 12.3 & 127.1 & 128.8 & 10.3 \\
Go2 & 11 & 93.4 & 36.7 & 36.8 & 144.1 & 144.1 & 3.9 \\
Go2 & 23 & 92.4 & 27.3 & 27.4 & 150.8 & 150.8 & 5.5 \\
Go2 & 47 & 95.1 & 30.3 & 30.2 & 145.9 & 146.0 & 4.8 \\
LEAP & 11 & 65.6 & 19.8 & 21.4 & 108.4 & 107.2 & 5.1 \\
LEAP & 23 & 65.7 & 19.2 & 19.1 & 107.5 & 107.6 & 5.6 \\
LEAP & 47 & 70.1 & 19.5 & 20.2 & 109.5 & 108.4 & 5.4 \\
\bottomrule
\end{tabular}
\end{table}

\begin{table}[t]
\centering
\caption{RGB context models on the SO-101 arm: law-following index per training seed, and its mean over seeds with the 95\% law-bootstrap interval (Table~\ref{tab:e4main} in the main paper).}
\label{tab:robot-rgb}
\footnotesize
\setlength{\tabcolsep}{3pt}
\begin{tabular}{@{}lccccc@{}}
\toprule
Corpus & Seed 11 & Seed 23 & Seed 47 & Mean & 95\% interval \\
\midrule
No twins & 0.073 & 0.066 & 0.045 & 0.061 & [0.047, 0.075] \\
Form twins & 0.086 & 0.033 & -0.035 & 0.028 & [0.022, 0.033] \\
Form--law twins & 0.077 & 0.039 & -0.003 & 0.037 & [0.036, 0.039] \\
\quad no effort & -0.011 & -0.068 & -0.022 & -0.034 & [-0.036, -0.031] \\
\bottomrule
\end{tabular}
\end{table}

\subsection{Contact Tasks}\label{sup:contact}
In this subsection, we give the settings and the complete results of the contact tasks in Sections~\ref{sec:robots}, \ref{sec:core}, and~\ref{sec:experiments} and Fig.~\ref{fig:robots} of the main paper.

\paragraph*{Objects} The apple is the YCB scan 013\_apple~\cite{S:calli2015ycb} scaled by 0.65 to fit the SO-101's jaws (4.9\,cm across, 4.7\,cm tall), uniform at 850\,kg/m$^3$ (57.6\,g), with the moments of the scaled scan; it collides through a convex hull of 128 vertices of the scan, which ships with the package. The ball is the YCB racquetball 057\_racquetball, a sphere of radius 27.9\,mm with its catalogued 41\,g as a thin shell, $I=\tfrac23mr^2$. The textured scans are fetched from the YCB archive on first render and checked against their checksums.

\paragraph*{Contact geometry and model} Each robot collides through its description's primitives where it has them (the Go2's trunk, legs, and feet; the LEAP hand's palm and phalanges), through boxes fitted to its meshes where it collides through meshes (the SO-101's two jaws and gripper housing; the LEAP fingertips). Collision of the robot with itself is an option of every scene, off in the runs reported here; when it is on, every pair of the robot's bodies collides except a body and its parent and the parts a description draws touching at the start pose, and the twins, which share one collision geometry, stay certified. Every contact has three dimensions and a pyramidal friction cone, with friction 0.8 on the robot and the ground, 0.9 on the apple, and 1.0 on the ball (MuJoCo takes the larger of the two), and the soft-contact parameters solref $(0.005, 1)$ and solimp $(0.95, 0.99, 0.001)$. The time constant, rather than a stiffness, makes the contact mass-normalized, so density form twins keep their relations. Every task integrates at 1/6000\,s, fine enough that the integrator's own error stays below every floor of the reading; the quadruped's trunk and the objects are free bodies, and convex collision runs to a tolerance of $10^{-10}$\,m.

\paragraph*{Tasks and controllers} Every task lasts 4\,s and is recorded at 120\,Hz. Each actuated joint follows its reference under impedance, with the nominal robot's gravity as feedforward on fixed bases. Stiffness and damping follow from the nominal diagonal inertia on the arm (natural frequency 15\,rad/s, damping ratio 0.7) and on the hand's wrist (25\,rad/s, ratio 1.0); they are fixed on the quadruped (60\,\Nm/rad and 1.5\,\Nm\,s/rad) and on the fingers, which keep the LEAP hand's own position gains (3\,\Nm/rad and 0.03\,\Nm\,s/rad). \emph{Apple}: the arm starts above the apple with its jaw open by 1.0\,rad, descends by 6\,cm to the grasp pose between 0.1 and 0.9\,s, closes from 0.9\,s, and lifts by 10\,cm between 2.0 and 3.2\,s, along minimum-jerk blends between waypoints found by inverse kinematics on the nominal arm; efforts are limited to 2.94\,\Nm{}, the servo's stall torque, and to 0.6\,\Nm{} on the gripper, which then squeezes the apple with about 12\,N. \emph{Ball}: the hand is mounted on a wrist hinge along its long axis; the ball settles on the open palm for 1\,s before the task starts, the fingers close to a power grasp between 0.2 and 0.9\,s, and the wrist turns the hand palm down between 1.3 and 2.8\,s. \emph{Trot}: from a stance settled for 1\,s, diagonal legs alternate at 3\,Hz, each thigh swinging by $\pm0.3$\,rad about 0.9\,rad and each knee bending by a further $0.9\sin^2$ of the swing phase, with the amplitude ramped in over 0.6\,s and efforts limited to 23.7\,\Nm{}. The swing is a smooth bump, so the recorded effort stays continuous between samples. The other law of the family is the one the fixed-base reading uses (Section~\ref{sup:robots-open}), so both readings recover one truth; the links as drawn replace each moving link's inertial by the uniform bounding box of its meshes with the link's mass.

\paragraph*{Reading} The reading balances generalized momentum window by window, \eqref{eq:momentum} in the main paper. Positions are interpolated by cubic splines onto a grid eight times finer than the clock, with free-body orientations as normalized splined quaternions and the body-frame angular velocity $\omega=2\,\bar q\dot q$, and every window spans four samples on either side under the weight $w=\cos^2(\pi s/2T)$, integrated by the trapezoidal rule. $F_0$ is MuJoCo's inverse dynamics at zero acceleration with constraints off, together with the known smooth dry friction, and $\dot M_S$ is differentiated along the grid. Equality-constraint impulses may have either sign, while joint-limit impulses are one-sided. The effort is the servo law's response to each forecast's own states, as Section~\ref{sec:state} in the main paper prescribes under position commands. The impulse directions are the four edges $n\pm\mu t_i$ of the friction pyramid, as MuJoCo's pyramidal cone uses them, of every contact within 2\,mm at every fourth grid point inside the window, each with the Jacobian of its own configuration. Their best nonnegative combination is removed in the metric $M^{-1}$ at the window's center, and the report is the RMS over samples of the kinetic norm of what remains: the mean generalized force of the window left unexplained. The floor is the 95\% quantile over 32 draws of the exact forecast under $10^{-5}$\,rad and m of pose noise, including small rotations of the free bodies. The floating forecasts add 2\,cm to the apple's height or to the trunk's height, which lifts the whole quadruped, or move the ball 2\,cm along the palm's normal, away from the palm.

\paragraph*{Law read through contact} The need of every window is affine in the density factor of each body and of the joints' rotors, viscous and dry dissipation, and springs, as at a fixed base (Section~\ref{sup:robots-open}), and the contact impulses enter each window as nonnegative unknowns. The factors minimize the sum over windows of what the best impulses leave in the kinetic metric. The problem is convex, and variable projection~\cite{S:golub2003separable} solves it by limited-memory BFGS: each evaluation solves the windows' nonnegative least squares, and the exact gradient follows from their residuals. Another law of the family is read over the links of the family; the form twin over every moving body, the objects and the quadruped's trunk included, and the joint terms. Where contact can supply what a link's mass would, the objective is flat along that link and the factor read wanders with the readout, so the error of each factor is its spread over six readouts with $10^{-5}$ noise; a link whose spread exceeds 0.25 is one the task does not excite and is not reported.

\paragraph*{Results} Tables~\ref{tab:contact-forward}, \ref{tab:contact-reading}, and~\ref{tab:contact-law} give every result. The law twins move with the original to machine precision in every task, and the form twin moves as the original again under scaled efforts and without effort. The exact forecasts read at 0.2--1.0 of their floors, which the readout noise sets: in momentum, an impact is an impulse rather than an acceleration that the differentiator spreads over neighboring samples, and the contact Jacobian follows a stance leg as it turns within the window. At 60\,Hz every fault stays flagged; another law on the trot, which changes its contacts every 1/6\,s, keeps 2.2 times its floor. The floating forecasts keep every joint angle, so their joint-space distance from the rollout is zero. Through contact the reading recovers another law link by link wherever the robot's own efforts reach the link: the quadruped's leg links, which carry the trunk in stance and swing free in flight, within 0.04; the hand's finger links within 0.005, while the base of the thumb, which presses on the ball, stays undetermined ($1.00\pm0.17$ against 1.04); and on the arm the upper arm, forearm, and jaw within 0.04, while the wrist and the gripper, which hold the apple against the table until the lift, trade mass through the apple's support. The form twin reads 1.40 on every term of the quadruped, 1.39--1.40 on the links of the hand, and 1.34--1.44 on those of the arm. An object reads its density while the robot alone carries it: the apple, lifted for more than a second, 1.40; the ball, whose 41\,g are a small share of what the grip exerts on it, 1.11.

\begin{table*}[t]
\centering
\caption{Contact tasks, forward (E1), over nine laws of each family under the task's controller: the largest engine certificate error of the boxes and ellipsoids; the largest gap of either law twin to the original (joint angles in rad, body positions in m); the density-1.4 form twin's gap in joint angles (rad) and body positions (m); its gap under 1.4 times the efforts and without effort; and the task's outcome for the original and the form twin.}
\label{tab:contact-forward}
\footnotesize
\setlength{\tabcolsep}{4pt}
\begin{tabular}{@{}lccccccl@{}}
\toprule
Task & Certificate & Law twins & $\times1.4$, joints & $\times1.4$, bodies & $\times1.4$, efforts $\times1.4$ & No effort & Outcome, original / $\times1.4$ \\
\midrule
SO-101, apple & $1\times10^{-16}$ & $4\times10^{-6}$ & 0.097--0.102 & 0.007--0.008 & $7\times10^{-6}$ & $4\times10^{-5}$ & lift 6.9--7.3 / 6.7--7.0\,cm \\
Go2, trot & $1\times10^{-17}$ & $7\times10^{-15}$ & 0.172--0.189 & 0.712--0.766 & $6\times10^{-15}$ & $7\times10^{-15}$ & walks 1.64--1.75 / 0.92--1.00\,m \\
LEAP, ball & $2\times10^{-17}$ & $5\times10^{-15}$ & 0.032--0.035 & 0.002--0.005 & $9\times10^{-15}$ & $3\times10^{-15}$ & holds the ball, 9 / 9 of 9 \\
\bottomrule
\end{tabular}
\end{table*}

\begin{table*}[t]
\centering
\caption{Contact tasks, inverse (E2): the effort each forecast leaves unexplained over the floor, with the contact forces as unknowns, as states at 120 and 60\,Hz, and its joint-space RMS distance from the rollout (mrad, 120\,Hz). Floors in kg$^{1/2}$\,m/s$^2$, the unit of the kinetic norm.}
\label{tab:contact-reading}
\footnotesize
\setlength{\tabcolsep}{4pt}
\begin{tabular}{@{}lccccccccc@{}}
\toprule
 & \multicolumn{2}{c}{SO-101, apple} & \multicolumn{2}{c}{Go2, trot} & \multicolumn{2}{c}{LEAP, ball} & \multicolumn{3}{c}{Distance (mrad)} \\
\cmidrule(lr){2-3}\cmidrule(lr){4-5}\cmidrule(lr){6-7}\cmidrule(lr){8-10}
Forecast & 120\,Hz & 60\,Hz & 120\,Hz & 60\,Hz & 120\,Hz & 60\,Hz & Apple & Trot & Ball \\
\midrule
Exact & 0.18 & 0.44 & 0.96 & 1.00 & 0.28 & 0.26 & 0 & 0 & 0 \\
Law scaled by 1.4 & 220 & 645 & 309 & 43.2 & 284 & 389 & 21 & 50 & 7 \\
Re-timed by 0.8 & 1960 & 6170 & 1941 & 314 & 8988 & 12470 & 148 & 173 & 193 \\
Another law of the family & 24.2 & 69.0 & 12.7 & 2.16 & 39.2 & 52.6 & 3 & 6 & 3 \\
Frozen & 3502 & 11036 & 1408 & 274 & 92949 & 129504 & 278 & 182 & 968 \\
Floating 2\,cm & 221 & 642 & 1213 & 248 & 2133 & 2767 & 0 & 0 & 0 \\
\midrule
Floor (kinetic metric) & 0.00623 & 0.00198 & 0.147 & 0.681 & 0.0159 & 0.0114 & & & \\
\bottomrule
\end{tabular}
\end{table*}

\begin{table*}[t]
\centering
\caption{Contact tasks, the law read through contact (E2, 120\,Hz). Another law of the family: the links read within 0.05 of their true factors with a spread below 0.05, the largest departure and the largest spread among them, and every other link the task excites (mean $\pm$ standard deviation over six readouts, true factor in parentheses). The density-1.4 form twin: the factors of its links, of its free bodies, and of the joints' rotors, viscous, and dry dissipation. The exact forecast: the range of every term.}
\label{tab:contact-law}
\footnotesize
\setlength{\tabcolsep}{3.5pt}
\begin{tabular}{@{}lcccp{3.6cm}cccc@{}}
\toprule
 & \multicolumn{4}{c}{Another law of the family} & \multicolumn{3}{c}{Form twin, density 1.4} & Exact \\
\cmidrule(lr){2-5}\cmidrule(lr){6-8}\cmidrule(lr){9-9}
Task & Within 0.05 & Largest & Spread & Other links & Links & Free bodies & Joint terms & All terms \\
\midrule
SO-101, apple & 3 of 6 & 0.035 & 0.018 & wrist 0.88$\pm$0.08 (1.28), gripper 1.43$\pm$0.06 (1.06) & 1.34--1.44 & apple 1.40 & 1.40, 1.40, 1.40 & 0.84--1.11 \\
Go2, trot & 12 of 12 & 0.040 & 0.001 & -- & 1.40--1.41 & trunk 1.40 & 1.41, 1.40, 1.39 & 0.99--1.03 \\
LEAP, ball & 15 of 16 & 0.005 & 0.007 & thumb base 1.00$\pm$0.17 (1.04) & 1.39--1.40 & palm 1.40, ball 1.11 & 1.07, 1.40, 1.40 & 0.87--1.01 \\
\bottomrule
\end{tabular}
\end{table*}

\paragraph*{Rigid contact} The reading admits any impulse within the friction cones over a window, which suits the soft contacts of the tasks. With rigid contact, the law depends on the contact mode, since separation, sticking, and sliding obey different impulse conditions, and cone feasibility alone would accept frictionless sliding~\cite{S:moreau1988unilateral,S:desaxce1991bipotential}. In this case, the Gauss residual becomes a velocity-level projection~\cite{S:moreau1988unilateral,S:jean1999nonsmooth}; for frictionless convex velocity constraints,
\begin{equation}
\begin{aligned}
v_{\rm free}&=v_k+\Delta t\,M^{-1}(F_k-c_k),\\
v^\ast_{k+1}&=\arg\min_{v\in\mathcal A_k}\tfrac12\|v-v_{\rm free}\|_M^2,
\end{aligned}
\label{eq:moreau}
\end{equation}
where $\mathcal A_k$ is the admissible velocity set; Coulomb sliding adds a mode-dependent friction law, and twins of a rigid-contact law are certified mode by mode.

\subsection{Any Robot}\label{sup:zoo}
FoLaSim mounts any MJCF or URDF description as a contact scene without per-robot code. A URDF is compiled by MuJoCo first: its meshes are resolved beside the file, links joined by fixed joints are fused into their parents, visual elements become the robot's look, its effort limits are dropped, and a legged robot's root link is given a free joint. FoLaSim compiles the description and writes the law out: every body's inertial as MuJoCo computes it, the joints' armature, damping, and spring stiffness, their dry friction as the smooth dissipation of Section~\ref{sup:robots-open}, and one torque motor per joint with unlimited effort. Kinematics, joint limits, and equality constraints stay as described; the reading takes a connect or weld constraint's impulse at one point common to both bodies, so that the joints upstream of both carry none of it. The robot's own collision geometry, primitives or meshes, collides with a ground and with the YCB racquetball, and optionally with itself, under the contact model of Section~\ref{sup:contact}. A free base stands at its home keyframe and a fixed base on the ground, each raised until its lowest collision point clears the ground, and the ball is dropped 15\,cm onto the robot's highest part. One joint impedance, computed from the nominal robot, tracks the home pose plus a multisine of 0.15\,rad on every joint (0.4--1.2\,Hz, ramped in over 0.5\,s), with a natural frequency of 15\,rad/s and gravity feedforward on a fixed base and of 40\,rad/s on a free one, at a damping ratio of 1. A fixed base runs for 3\,s and a free base for 1\,s, as long as joint impedance alone keeps a legged robot standing, at a step of 1/6000\,s and a record rate of 120\,Hz; the reading is that of Section~\ref{sup:contact}, with its floor from eight noisy readouts. Sixteen robots of MuJoCo Menagerie~\cite{S:menagerie2022}, at the commit of the shipped robots, are known by name and fetched on first use, each with its upstream license. Table~\ref{tab:zoo} gives E1 and the reading for every one: the twins hold in contact on arms, hands, quadrupeds, and humanoids alike, with mesh and primitive collision, closed kinematic chains, and joint springs, and the reading flags each density-1.4 twin 35--306 times over its floor.

\begin{table*}[t]
\centering
\caption{Any robot: sixteen robots of MuJoCo Menagerie in contact (Section~\ref{sup:zoo}). Actuated joints, moving mass, and the most contacts at once (with the ball); the largest engine certificate error of the boxes and ellipsoids; the largest gap of either law twin to the original in contact (rad or m); the density-1.4 form twin's joint-angle gap under the same controller (rad), and its largest gap under 1.4 times the efforts and without effort; and the effort the exact forecast and the form twin's leave unexplained over the floor, as states at 120\,Hz with the contact forces unknown.}
\label{tab:zoo}
\footnotesize
\setlength{\tabcolsep}{4pt}
\begin{tabular}{@{}lccccccccc@{}}
\toprule
 & & & & \multicolumn{2}{c}{Law twins} & \multicolumn{2}{c}{Form twin, density 1.4} & \multicolumn{2}{c}{Read $\div$ floor} \\
\cmidrule(lr){5-6}\cmidrule(lr){7-8}\cmidrule(lr){9-10}
Robot & Joints & Mass (kg) & Contacts & Certificate & Gap & Gap (rad) & Scaled, none & Exact & $\times1.4$ \\
\midrule
Kinova Gen3 & 7 & 6.5 & 2 (2) & $9\times10^{-17}$ & $2\times10^{-13}$ & 0.04 & $4\times10^{-13}$ & 0.29 & 166 \\
KUKA LBR iiwa 14 & 7 & 25.6 & 1 (1) & $3\times10^{-16}$ & $9\times10^{-15}$ & 0.08 & $6\times10^{-11}$ & 0.43 & 185 \\
Franka Emika Panda & 9 & 16.8 & 1 (1) & $2\times10^{-16}$ & $1\times10^{-14}$ & 0.04 & $3\times10^{-13}$ & 0.44 & 123 \\
Universal Robots UR5e & 6 & 17.0 & 2 (2) & $2\times10^{-16}$ & $1\times10^{-15}$ & 0.04 & $2\times10^{-13}$ & 0.43 & 149 \\
Trossen ViperX 300 S & 8 & 2.8 & 1 (1) & $2\times10^{-17}$ & $1\times10^{-14}$ & 0.13 & $9\times10^{-15}$ & 0.18 & 151 \\
UFACTORY xArm7 & 13 & 11.2 & 1 (1) & $1\times10^{-17}$ & $2\times10^{-15}$ & 0.04 & $4\times10^{-11}$ & 0.23 & 196 \\
\midrule
Wonik Allegro Hand & 16 & 0.3 & 2 (2) & $2\times10^{-19}$ & $1\times10^{-16}$ & 0.13 & $1\times10^{-16}$ & 0.32 & 35 \\
Shadow Dexterous Hand & 24 & 0.8 & 1 (1) & $3\times10^{-18}$ & $1\times10^{-17}$ & 0.15 & $3\times10^{-11}$ & 0.43 & 177 \\
\midrule
Unitree A1 & 12 & 12.5 & 6 (2) & $9\times10^{-18}$ & $2\times10^{-15}$ & 0.12 & $2\times10^{-14}$ & 0.55 & 306 \\
ANYbotics ANYmal C & 12 & 45.0 & 5 (1) & $4\times10^{-17}$ & $2\times10^{-14}$ & 0.12 & $8\times10^{-14}$ & 0.38 & 125 \\
Unitree Go1 & 12 & 12.7 & 6 (2) & $9\times10^{-18}$ & $2\times10^{-14}$ & 0.15 & $1\times10^{-14}$ & 0.70 & 260 \\
Boston Dynamics Spot & 12 & 50.3 & 9 (1) & $9\times10^{-18}$ & $1\times10^{-15}$ & 0.25 & $6\times10^{-13}$ & 0.22 & 233 \\
\midrule
Unitree G1 & 29 & 33.3 & 8 (1) & $7\times10^{-18}$ & $1\times10^{-13}$ & 0.14 & $3\times10^{-13}$ & 0.39 & 85 \\
Unitree H1 & 19 & 51.4 & 8 (1) & $2\times10^{-17}$ & $2\times10^{-14}$ & 0.09 & $3\times10^{-14}$ & 0.46 & 84 \\
ROBOTIS OP3 & 20 & 3.1 & 16 (1) & $2\times10^{-18}$ & $2\times10^{-12}$ & 0.02 & $2\times10^{-12}$ & 0.53 & 138 \\
Booster T1 & 23 & 31.6 & 9 (1) & $1\times10^{-17}$ & $2\times10^{-14}$ & 0.09 & $3\times10^{-13}$ & 0.61 & 83 \\
\bottomrule
\end{tabular}
\end{table*}

\subsection{Reading the Law from Motion}\label{sup:law}
\paragraph*{Terms} A description is trusted for its kinematics, joint limits, equality constraints, and collision geometry; its law serves only as the starting point of the reading. The terms read are the ten inertial parameters $\pi_\ell=(m,\,mc,\,I_o)$ of every moving link about its body frame and, for every actuated joint, its rotor armature, viscous damping, and dry friction, with the spring stiffness where the description declares a spring; objects keep their own law. Every term enters the momentum balance \eqref{eq:momentum} of the main paper linearly, and its column is MuJoCo's own: the need of a model that carries that term alone, every other mass and joint term set to zero. A link's ten columns come from ten bodies whose parameters span $\R^{10}$, through the inverse of their parameter matrix, with the compiler's shortcuts for coinciding body and inertial frames cleared. The reading stops unless the columns reproduce the need of the description to $10^{-6}$.

\paragraph*{Reading} The law minimizes the sum over windows of what the best contact impulses leave, in the kinetic metric, plus a Gaussian prior centered on the description: each term has a spread of five times its natural size ($m$, $m\ell$, or $m\ell^2$, with $\ell$ the link's radius of gyration in the description), and the prior is weighed by the residual variance of a nearly unconstrained fit. The law is constrained to be physically consistent: each link's pseudo-inertia~\cite{S:wensing2018lmi} is at least 0.05 times the description's, a linear matrix inequality that also keeps the link away from degeneracy, and each joint term is nonnegative. For fixed active contact edges the problem is a least-squares problem under linear matrix inequalities, convex, and ADMM solves it, projecting each pseudo-inertia onto its cone by an eigenvalue cut; the active sets are then updated, and a line search accepts the step, so every iterate is consistent. The reading is repeated once in the kinetic metric of the law just read. A combination of terms counts as determined when its standard error, i.e., the standard deviation of the residual per row divided by the singular value of the projected columns in natural units, is below a tenth of its natural size; contact edges active at the law read are projected out first.

\paragraph*{Wrong descriptions} Each robot's inertial table is replaced once by a wrong one (seed 0). Each link's mass is scaled log-uniformly within $[1/1.6,\,1.6]$ and its center of mass moved by a Gaussian of a quarter of its radius of gyration. Its principal moments are scaled within the same range, under the triangle inequality, and its principal axes turned by up to $20^\circ$. Each joint's armature, damping, and dry friction are scaled within $[1/2,\,2]$. The true robot runs its scene of Section~\ref{sup:zoo}, or its task, under its controller, and the reading takes the wrong description, the states at 120\,Hz, and the servo's effort, with the contact forces unknown. The law error is the RMS difference of the generalized forces two laws need along a motion (inertial, bias, and dissipation forces; efforts and contact left out), relative to the true law's, over the actuated joints and over the free bodies separately; law twins score zero. Held-out programs change the frequencies and phases of the multisine (seed 7); a scripted task is run again from its start.

\paragraph*{Results} Table~\ref{tab:law} gives every reading. Every law read leaves no more effort unexplained than the true law itself, and the motion determines 24 to 202 combinations of the 78 to 387 terms. Wherever the motion excites the robot, the law read follows the true dynamics. On the six open-chain arms, the SO-101 mounted from its URDF among them, and on the Shadow hand, the generalized forces of a held-out program agree to 0.01--0.08\% (the wrong table: 1.5--46\%), and the robot under the law read runs that program within $3\times10^{-5}$--$4\times10^{-3}$\,rad of the truth. On the xArm7, whose gripper is a closed linkage held by equality impulses, the law read cuts the error from 22 to 3\% and the gap from 0.05 to $6\times10^{-3}$\,rad. In the tasks the arm's law is read to 0.24\% while it lifts the apple, and the trotting quadruped's to 0.35\% in its legs and 0.22\% in its trunk. Where contact holds a link, part of its law is absorbed by the contact. The LEAP hand that grips the ball and the Allegro hand on which the ball rests are read to 10.5\% and 15\%, and legged robots that keep every foot down show their trunk to within 0.2--4\% (H1 9\%) but their legs only as far as stance excites them: OP3, G1, ANYmal C, and T1 are read to 1.5--5.6\% and A1 and Go1 to 8--10\%, while on Spot and H1 the ground forces absorb the law of the legs and the leg rows stay undetermined. The equimomental boxes of every law read certify against it to $3\times10^{-16}$ and move as it does.

\begin{table*}[t]
\centering
\caption{The law read from motion with a wrong inertial table (Section~\ref{sup:law}). Links and terms read, and the combinations the motion determines to a tenth of their natural size; law error of the wrong table and of the law read on the rows of the actuated joints and of the free bodies (\%); joint-angle gap to the truth under a held-out program, for the wrong table and the law read (rad); engine certificate of the equimomental boxes of the law read. $^\dagger$A scripted task: the law error is on its episode, and the gap is its run repeated from the start.}
\label{tab:law}
\footnotesize
\setlength{\tabcolsep}{4pt}
\begin{tabular}{@{}lccccccccccc@{}}
\toprule
 & & & & \multicolumn{2}{c}{Joints (\%)} & \multicolumn{2}{c}{Free bodies (\%)} & \multicolumn{2}{c}{Gap (rad)} & \\
\cmidrule(lr){5-6}\cmidrule(lr){7-8}\cmidrule(lr){9-10}
Robot & Links & Terms & Determined & Wrong & Read & Wrong & Read & Wrong & Read & Certificate \\
\midrule
SO-101, apple$^\dagger$ & 6 & 78 & 31 & 54.9 & 0.24 & -- & -- & 0.13 & 0.002 & $1\times10^{-16}$ \\
LEAP hand, ball$^\dagger$ & 16 & 211 & 123 & 28.1 & 10.5 & -- & -- & 0.05 & 0.016 & $5\times10^{-18}$ \\
Go2, trot$^\dagger$ & 12 & 156 & 111 & 49.6 & 0.35 & 10.6 & 0.22 & 0.14 & 0.0023 & $9\times10^{-19}$ \\
\midrule
SO-101 (URDF) & 6 & 78 & 49 & 1.49 & 0.06 & -- & -- & 0.73 & 0.0043 & $3\times10^{-18}$ \\
Franka Emika Panda & 10 & 127 & 64 & 6.29 & 0.07 & -- & -- & 0.034 & $5\times10^{-5}$ & $2\times10^{-16}$ \\
Universal Robots UR5e & 6 & 78 & 47 & 24.0 & 0.08 & -- & -- & 0.039 & $7\times10^{-5}$ & $2\times10^{-16}$ \\
Kinova Gen3 & 7 & 91 & 62 & 26.7 & 0.07 & -- & -- & 0.033 & $6\times10^{-5}$ & $4\times10^{-18}$ \\
KUKA LBR iiwa 14 & 7 & 91 & 58 & 12.4 & 0.06 & -- & -- & 0.67 & $3\times10^{-4}$ & $1\times10^{-16}$ \\
UFACTORY xArm7 & 14 & 181 & 61 & 22.4 & 2.98 & -- & -- & 0.05 & 0.0064 & $3\times10^{-16}$ \\
Trossen ViperX 300 S & 9 & 114 & 56 & 45.7 & 0.03 & -- & -- & 0.1 & $8\times10^{-5}$ & $5\times10^{-17}$ \\
\midrule
Shadow Dexterous Hand & 24 & 312 & 135 & 9.06 & 0.01 & -- & -- & 0.071 & $3\times10^{-5}$ & $3\times10^{-18}$ \\
Wonik Allegro Hand & 20 & 248 & 78 & 22.5 & 14.8 & -- & -- & 0.13 & 0.0033 & $3\times10^{-18}$ \\
\midrule
Unitree A1 & 13 & 166 & 43 & 26.0 & 9.81 & 4.14 & 1.20 & 0.037 & 0.0099 & $5\times10^{-18}$ \\
Unitree Go1 & 13 & 166 & 41 & 24.2 & 8.04 & 8.48 & 0.44 & 0.15 & 0.0057 & $4\times10^{-18}$ \\
ANYbotics ANYmal C & 13 & 166 & 95 & 31.9 & 2.67 & 14.3 & 0.81 & 0.3 & 0.02 & $2\times10^{-17}$ \\
Boston Dynamics Spot & 13 & 166 & 24 & 20.4 & 67.9 & 12.7 & 4.10 & 0.074 & 0.094 & $9\times10^{-18}$ \\
\midrule
Unitree G1 & 30 & 387 & 202 & 19.6 & 3.32 & 1.87 & 0.24 & 0.067 & 0.006 & $3\times10^{-18}$ \\
Unitree H1 & 20 & 257 & 130 & 35.0 & 106 & 10.0 & 8.99 & 0.077 & 0.079 & $6\times10^{-18}$ \\
Booster T1 & 24 & 309 & 124 & 34.2 & 5.58 & 6.08 & 0.29 & 0.078 & 0.0088 & $1\times10^{-17}$ \\
ROBOTIS OP3 & 21 & 270 & 134 & 36.0 & 1.46 & 5.02 & 0.79 & 0.018 & 0.0022 & $1\times10^{-18}$ \\
\bottomrule
\end{tabular}
\end{table*}

\subsection{Recorded Robots}\label{sup:robots}
An SI anchor on recorded robots needs the same ingredients as the simulator: synchronized states, the semantics of the effort port, and a trusted dynamics model of everything that moves. The reading of Section~\ref{sup:law} supplies the last from recorded motion and effort, given the kinematics, for every combination the recording determines. An audit of 16 DROID~\cite{S:khazatsky2024droid} trajectories for timestamp consistency, joint-state availability, actuator-field semantics, and compatibility with a nominal robot dynamics model shows what is missing: the gripper, camera, payload, and external contacts contribute forces that the nominal model omits. A calibrated anchor therefore uses verified free-motion segments and calibrated loads, with the pose, effort, and structure sensitivities of Proposition~\ref{prop:chain}; a residual against the nominal model is a diagnostic of the reference, not a prediction error of a model.

\section{Proofs}\label{app:proofs}
This section gives the proofs of the statements in Sections~\ref{sup:why} and~\ref{sup:measurement} and in Section~\ref{sec:why} of the main paper.

\begin{proof}[Proof of Lemma~\ref{lem:timing}]
$\phi(\tilde q(t))=\phi(q(\sigma(t)))=0$, and $J(\tilde q)\dot{\tilde q}=\dot\sigma\,J(q)\dot q=0$. Since $\tilde q$ is itself a curve on $S$, Gauss's formula \eqref{eq:gaussformula} applies to it, so its normal acceleration is $\II(\dot{\tilde q},\dot{\tilde q})$. With $\dot{\tilde q}=\dot\sigma\,\dot q\circ\sigma$, the covariant derivative is $D_t\dot{\tilde q}=\ddot\sigma\,\dot q+\dot\sigma^2D_t\dot q$, whose tangential part is $\ddot\sigma\,\dot q+\dot\sigma^2\nabla^S_t\dot q$. If $q$ satisfies \eqref{eq:split}, $\tilde q$ does so only if $\ddot\sigma\,\dot q+\dot\sigma^2P_TM^{-1}F(q,\dot q)=P_TM^{-1}F(q,\dot\sigma\dot q)$, which fails in general.
\end{proof}

\begin{proof}[Proof of Lemma~\ref{lem:reaction}]
The constrained motion is determined by $\phi(q)=0$ and $N\Tr(MD_t\dot q-F)=0$. Since $N\Tr J\Tr=0$, replacing $F$ by $F+J\Tr\mu$ leaves both unchanged, and the normal component of \eqref{eq:full} then holds with reaction $\nu-\mu$. Only $N\Tr F\in\R^d$ enters the motion.
\end{proof}

\begin{proposition}[Layers of a forecast]\label{prop:layers}
Let $\hat q(t)$ be the readout of a forecast that stays on $S$ under input $u(t)$. (i) Its admissibility, i.e., closure, tangent velocity, and normal acceleration $\II(\dot{\hat q},\dot{\hat q})$, does not depend on the applied forces and holds for every time reparametrization $\hat q(\sigma(t))$. (ii) A force perturbation $J^\top\mu$ leaves its motion unchanged. (iii) It obeys the dynamics if and only if the tangential mismatch $\epsilon_T=P_T(D_t\dot{\hat q}-M^{-1}F)$ vanishes, and the residual $r=\|\epsilon_T\|_g$ depends neither on the coordinates nor on the choice of $N$.
\end{proposition}

\begin{proof}[Proof of Proposition~\ref{prop:layers}]
(i) Closure $\phi(\hat q)=0$ and tangency $J\dot{\hat q}=0$ involve no force. For any curve on $S$, Gauss's formula \eqref{eq:gaussformula} gives the normal part of $D_t\dot{\hat q}$ as $\II(\dot{\hat q},\dot{\hat q})$, a function of position and velocity alone. By Lemma~\ref{lem:timing}, $\hat q(\sigma(t))$ is again a curve on $S$ with tangent velocity, so it satisfies the same three conditions. (ii) is Lemma~\ref{lem:reaction}. (iii) Put $f=MD_t\dot{\hat q}-F$, where $MD_t\dot{\hat q}=M\ddot{\hat q}+c(\hat q,\dot{\hat q})$ in coordinates. The curve solves \eqref{eq:full} for some reaction $\nu$ if and only if $f\in\operatorname{range}J\Tr=\ker N\Tr$, that is, $N\Tr f=0$. Since $\epsilon_T=P_TM^{-1}f=N(N\Tr MN)^{-1}N\Tr f$ and $N(N\Tr MN)^{-1}$ has full column rank, $\epsilon_T=0$ exactly when $N\Tr f=0$. Moreover,
\[
\begin{aligned}
\|\epsilon_T\|_g^2&=f\Tr N(N\Tr MN)^{-1}N\Tr MN(N\Tr MN)^{-1}N\Tr f\\
&=f\Tr N(N\Tr MN)^{-1}N\Tr f .
\end{aligned}
\]
Replacing $N$ by $NT$ with invertible $T$ leaves $N(N\Tr MN)^{-1}N\Tr$ unchanged, and $\|\cdot\|_g$ is defined without coordinates. In local coordinates $s$ with $N=\partial\iota/\partial s$, d'Alembert's reduction gives $N\Tr f=\rho$ and $N\Tr MN=M_S$, so $\|\epsilon_T\|_g^2=\rho\Tr M_S^{-1}\rho$, which is \eqref{eq:residual}.
\end{proof}

\begin{proof}[Proof of Theorem~\ref{thm:channels}]
(i) Scaling a passive law multiplies its equation by $c$ without changing the motion or observation distribution. Thus its scale sensitivity and Fisher information vanish. (ii) Differentiating the Gaussian observation likelihood gives $\mathcal F=\sigma_q^{-2}\sum_jS_j^\top S_j$; hence $z^\top\mathcal Fz=\sigma_q^{-2}\sum_j\|S_jz\|^2$. Different continuous trajectories need not differ at the sampled times, which is why excitation alone is insufficient for the sampled statement.

(iii) Put $e=r-s$, $h_\theta=C_S\dot s+\nabla V_S+d_S$, and $\epsilon=\gamma^{-1/2}$. The error equation is
\[
M_S(s)\ddot e+\epsilon^{-1}\Gamma_d\dot e+\epsilon^{-2}\Gamma_pe=M_S(s)\ddot r+h_\theta(s,\dot s)+\epsilon^{-1}\Gamma_d\dot r.
\]
Under the stated uniform stability and smoothness assumptions, expansion about the tracking solution is uniform on $[t_0,T]$, together with the parameter derivatives needed below; the stable initial layer and its derivatives decay exponentially in $t_0/\epsilon$~\cite{S:kokotovic1986singular}. Matching orders gives
\[
\begin{aligned}
e={}&\epsilon\Gamma_p^{-1}\Gamma_d\dot r\\
&+\epsilon^2\left[\Gamma_p^{-1}\tau_\theta(r,\dot r,\ddot r)-(\Gamma_p^{-1}\Gamma_d)^2\ddot r\right]+O(\epsilon^3).
\end{aligned}
\]
The reference regularity supports this expansion through the state derivatives used by the inverse dynamics. The first coefficient and the second term in brackets are independent of $\theta$. Differentiation therefore gives $S_j=-\epsilon^2\Gamma_p^{-1}Y_j+O(\epsilon^3)$. Substitution into (ii) yields \eqref{eq:fisher-motion}. For velocity-error feedback, the $\epsilon^{-1}\Gamma_d\dot r$ forcing is absent; then $e=\epsilon^2\Gamma_p^{-1}\tau_\theta(r,\dot r,\ddot r)+O(\epsilon^3)$ and the same leading sensitivity follows.

(iv) The actual effort equals $\tau_\theta(s,\dot s,\ddot s)$. The state and its first two derivatives approach the reference with error $O(\epsilon)$, while their parameter sensitivities are $O(\epsilon^2)$. The chain rule consequently gives $\partial_\theta u=Y_j+O(\epsilon)$. The effort-noise likelihood gives the claimed information. (v) follows directly from Lemma~\ref{lem:reaction}.
\end{proof}

\begin{proof}[Proof of Corollary~\ref{cor:ladder}]
For \eqref{eq:common}, $\tau_{c\lambda}=c\,\tau_\lambda$, so the derivative along the scale at $c=1$ is $Y_j=\tau_\lambda(t_j)$; substitute into Theorem~\ref{thm:channels} (i), (iii), and (iv) with scalar gains.
\end{proof}

\begin{proof}[Proof of Proposition~\ref{prop:correspond}]
By (T1) and (T2), $L_A(s,\dot s)=\tfrac12\dot s\Tr M_A\dot s-V_A=L_B(\psi(s),D\psi\,\dot s)+\text{const}$. The Euler--Lagrange expression is covariant under point transformations, $\mathcal E_A[s]=D\psi\Tr\,\mathcal E_B[\psi\circ s]$ for every curve $s$. By (T3) and (T4), the nonconservative forces satisfy $B_Au_A-d_A=D\psi\Tr(B_BKu_A-d_B)$. Because $D\psi$ is invertible, $\mathcal E_A[s]=B_Au_A-d_A$ holds if and only if $\mathcal E_B[\psi\circ s]=B_BKu_A-d_B$.
\end{proof}

\begin{proof}[Proof of Proposition~\ref{prop:order}]
Write $A=N\Tr MN\succ0$ and $b=N\Tr B$, so that $\Lambda=b\Tr A^{-1}b$, which is symmetric positive semidefinite. Added mass gives $A'=A+N\Tr\Delta MN\succeq A$, hence $A'^{-1}\preceq A^{-1}$ and $\Lambda'\preceq\Lambda$. Added constraints give $N'=NW$ with $W$ of full column rank, and $\Lambda'=b\Tr W(W\Tr AW)^{-1}W\Tr b$. The matrix $A^{1/2}W(W\Tr AW)^{-1}W\Tr A^{1/2}$ is an orthogonal projector, hence $\preceq I$, so $W(W\Tr AW)^{-1}W\Tr\preceq A^{-1}$ and $\Lambda'\preceq\Lambda$.
\end{proof}

\begin{proof}[Proof of Theorem~\ref{thm:characterization} in the main paper]
Necessity: if the predictor is correct, (C1) follows from Proposition~\ref{prop:correspond}, (C2) is immediate, (C3) holds with $\theta=\lambda\in\Theta$, and (A) holds on any set. Sufficiency: at $\omega_0$, (C3) gives $\theta_0\in\Theta$ with $\hat a_{\omega_0}=a^\ast_{\theta_0}$ for all $\iota$, and (A) gives $a^\ast_{\theta_0}=a^\ast_{\lambda_0}$ on $\mathcal I_A$; since $\mathcal I_A$ is identifying, $\theta_0$ is equivalent to $\lambda_0$, so the predictor is correct at $\omega_0$ for all $\iota$. If it is correct at $(m,\lambda_1)$, (C2) gives $\hat a_{(m,\lambda_2)}=\hat a_{(m,\lambda_1)}+a^\ast_{\lambda_2}-a^\ast_{\lambda_1}=a^\ast_{\lambda_2}$. If it is correct at $(A,\lambda)$, (C1) and Proposition~\ref{prop:correspond} give $\hat a_{(B,\lambda)}=(\psi,K)_\ast a^\ast_{\lambda,A}=a^\ast_{\lambda,B}$. Connectivity of $\Omega$ to $\omega_0$ completes the proof.
\end{proof}

\begin{proof}[Proof of Proposition~\ref{prop:nonimplication}]
For \eqref{eq:common}, $a^\ast_\lambda(q,\dot q,u)=a^\ast_\lambda(q,\dot q,0)+u/I$, and the twins of Section~\ref{sec:twins} in the main paper share the coordinate $q$ and input $u$, so transport is the identity. (a) A fixed law transported to all mechanisms is invariant across twins and explained by $\bar\lambda$, but does not change with $\lambda$ and is wrong at $\omega_0$. (b) $\hat a=a^\ast_\lambda-g_0\,\delta h_m\sin q/I$ equals $a^\ast_{(I,h+\delta h_m,b,f,\varepsilon)}$, so (C3) holds; the added term does not depend on $(h,b,f)$, so differences between laws with equal $I$ are exact and (C2) holds; (C1) fails when $\delta h_A\ne\delta h_B$. (c) On actuated interventions outside $\mathcal I_A$ the predictor returns $a^\ast_{c\lambda}=a^\ast_\lambda(q,\dot q,0)+u/(cI)$. Twins are treated alike, so (C1) holds; for laws with equal $I$ the input terms cancel in differences, so (C2) holds; (A) holds by construction. Any $\theta$ explaining $\mathcal I_A$ is equivalent to $\lambda$ because $\mathcal I_A$ is identifying, yet $\lambda$ does not explain the actuated predictions outside $\mathcal I_A$, so (C3) fails. (d) Predicting with $c\lambda$ everywhere satisfies (C1)--(C3) as in (c); on passive interventions $a^\ast_{c\lambda}=a^\ast_\lambda$ by Proposition~\ref{prop:scale}(i), so the equality part of the anchor test holds on passive observations. That set is not identifying and does not satisfy (A); actuated predictions remain wrong.
\end{proof}

\begin{proof}[Proof of Theorem~\ref{thm:quantitative}]
On differences of fields, $(\psi,K)_\ast$ acts by $D\psi$, and the reference metrics are transported as in (T1); since the test sets of twins correspond, $\|(\psi,K)_\ast\delta a\|_{\mathcal I}=\|\delta a\|_{\mathcal I}$. At $\omega_0$, monotonicity of $\|\cdot\|_{\mathcal S}$ in $\mathcal S$ and \eqref{eq:kappa} give
\[
\begin{aligned}
\|\hat a-a^\ast_{\lambda_0}\|_{\mathcal I}
&\le\varepsilon_3+\kappa_A\|a^\ast_{\theta_0}-a^\ast_{\lambda_0}\|_{\mathcal I_A}\\
&\le\varepsilon_3+\kappa_A\big(\|a^\ast_{\theta_0}-\hat a\|_{\mathcal I_A}+\|\hat a-a^\ast_{\lambda_0}\|_{\mathcal I_A}\big)\\
&\le(1+\kappa_A)\,\varepsilon_3+\kappa_A\,\varepsilon_A .
\end{aligned}
\]
Along a law change, $\|\hat a_{\lambda_2}-a^\ast_{\lambda_2}\|_{\mathcal I}\le\|\hat a_{\lambda_1}-a^\ast_{\lambda_1}\|_{\mathcal I}+\varepsilon_2$. Along a twin pair, Proposition~\ref{prop:correspond} gives $(\psi,K)_\ast a^\ast_{\lambda,A}=a^\ast_{\lambda,B}$, so $\|\hat a_B-a^\ast_B\|_{\mathcal I}\le\varepsilon_1+\|(\psi,K)_\ast(\hat a_A-a^\ast_A)\|_{\mathcal I}=\varepsilon_1+\|\hat a_A-a^\ast_A\|_{\mathcal I}$. Induction along the path gives \eqref{eq:quantitative}. Conversely, take $\theta_0=\lambda_0$, and bound each relation by the triangle inequality through the true fields.
\end{proof}

\begin{proof}[Proof of Proposition~\ref{prop:kappa}]
For $\delta\beta=\beta(\theta)-\beta(\lambda_0)$, $\|\Phi\delta\beta\|_\iota^2=\delta\beta\Tr G_\iota\delta\beta$ and $\|\Phi\delta\beta\|^2_{\mathcal I_A}=\max_{\iota'\in\mathcal I_A}\delta\beta\Tr G_{\iota'}\delta\beta\ge|\mathcal I_A|^{-1}\delta\beta\Tr G_A\delta\beta$. If $\ker G_A\subseteq\ker G_\iota$ for all $\iota$, the generalized Rayleigh quotient $\delta\beta\Tr G_\iota\delta\beta/\delta\beta\Tr G_A\delta\beta$ is bounded by $\lambda_{\max}(G_A^{+1/2}G_\iota G_A^{+1/2})$, which gives \eqref{eq:kappa-bound}; the quotient is homogeneous, so the supremum over the open set of admissible $\delta\beta$ equals that over all directions. If some $\delta\beta\in\ker G_A$ has $\delta\beta\Tr G_\iota\delta\beta>0$, the ratio in \eqref{eq:kappa} is unbounded. For \eqref{eq:common}, $a^\ast_\lambda=(u-g_0h\sin q-b\dot q-f\tanh(\dot q/\varepsilon))/I=\Phi\beta$; on passive interventions the first column of $\Phi$ vanishes, so $e_1\in\ker G_A$ while $e_1\Tr G_\iota e_1>0$ for any actuated $\iota$.
\end{proof}

\begin{proof}[Proof of Proposition~\ref{prop:invariance}]
(i) The covariance used in the proof of Proposition~\ref{prop:correspond} holds for arbitrary curves, so $\rho_A=D\psi\Tr\rho_B$, and by (T1) $M_A^{-1}=D\psi^{-1}M_B^{-1}D\psi^{-\top}$; hence $\rho_A\Tr M_A^{-1}\rho_A=\rho_B\Tr M_B^{-1}\rho_B$. (ii) $\Lambda_A=B_A\Tr M_A^{-1}B_A=K\Tr B_B\Tr D\psi\,D\psi^{-1}M_B^{-1}D\psi^{-\top}D\psi\Tr B_BK=K\Tr\Lambda_BK$. (iii) By (T1) and (T2) the energies differ by a constant, by (T4) the input powers agree, and by (T3) the dissipated powers agree. (iv) A1 and B1 transmit forces through different joints and gear contacts, so their reactions differ although their laws agree.
\end{proof}

\begin{proof}[Proof of Proposition~\ref{prop:discrete}]
(i) For $\psi(s)=Ps+c$, any common linear velocity stencil that annihilates constants commutes with $P$. Endpoint averages and differences therefore satisfy $\bar s_B=P\bar s_A+c$, $\bar v_B=P\bar v_A$, and $a_B=Pa_A$. The continuous covariance identity for the Euler--Lagrange covector applies at these discrete arguments, giving $\rho_A=P^\top\rho_B$. Equation (T1) then gives equality of the dual metric norms.

(ii) Let $t_m$ be the interval midpoint. Endpoint position averages differ from the true position at $t_m$ by $O(\Delta t^2)$; endpoint averages of exact velocities have the same accuracy, as does their divided difference as an acceleration approximation. With a centered second-order velocity stencil, Taylor expansion on the common $C^4$ curve gives these estimates as well. The transformed curve $\psi\circ s$ has the same regularity. Smooth bounded coefficients imply that each reduced residual covector differs from its continuous counterpart at $t_m$ by $O(\Delta t^2)$. Uniform positive definiteness makes the corresponding dual norms locally Lipschitz, including at zero residual. By Proposition~\ref{prop:invariance}, the two continuous norms agree, so the triangle inequality yields $|r_k^A-r_k^B|=O(\Delta t^2)$. Each residual itself approximates $r(t_m)$; there is no factor of $\Delta t$ in this force-level discretization.
\end{proof}

\begin{proof}[Proof of Proposition~\ref{prop:decidable}]
(i) Take $\theta,\theta'\in\Theta_\eta(\mathcal E)$ with $\|a^\ast_\theta-a^\ast_{\theta'}\|_{\mathcal Q}\ge w_{\mathcal Q}-\epsilon$. A predictor that receives only $\mathcal E$ outputs the same $\hat a$ in both worlds, and the triangle inequality gives $\max\{\|\hat a-a^\ast_\theta\|_{\mathcal Q},\|\hat a-a^\ast_{\theta'}\|_{\mathcal Q}\}\ge(w_{\mathcal Q}-\epsilon)/2$; let $\epsilon\to0$. (ii) By Proposition~\ref{prop:correspond}, $(\psi,K)_\ast$ maps the field of each law on $A$ to the field of the same law on $B$, and it is an isometry on differences (proof of Theorem~\ref{thm:quantitative}); consistency within $\eta$ and widths are therefore preserved. (iii) For $\theta,\theta'\in\Theta_\eta(\mathcal E)$, $\|\Phi(\beta-\beta')\|_{\mathcal E}\le2\eta$ by the triangle inequality through $a_{\rm obs}$, and the Rayleigh-quotient bound of Proposition~\ref{prop:kappa} transfers it to $\mathcal Q$. If a feasible segment of parameter differences lies in the evidence null space but not the query null space, its endpoints have identical evidence and different query predictions, giving positive width. Infinite width requires an unbounded feasible direction of this kind; a bounded parameter set need not have infinite width.
\end{proof}

\begin{proof}[Proof of Proposition~\ref{prop:chain}]
(a) Expand $\rho(s+\beta,\dot s+\dot\beta,\ddot s+\ddot\beta)$ to first order. (b) With $\hat u=u+e_u$, $\hat\rho=\rho-B_Se_u$, so by the reverse triangle inequality in the $M_S^{-1}$ norm, $|\hat r-r|\le\|B_Se_u\|_{M_S^{-1}}=\|e_u\|_\Lambda$. (c) $|\|\hat P_Tx\|_g-\|P_Tx\|_g|\le\|(\hat P_T-P_T)x\|_g\le\varepsilon_T\|x\|_g$ with $x=D_t\dot{\hat q}-M^{-1}F$; for a correct motion $x=M^{-1}J\Tr\nu$ and $r=0$.
\end{proof}

\begin{proof}[Proof of Proposition~\ref{prop:scale}]
Multiplying \eqref{eq:common} by $c$ gives (i) when $u\equiv0$ and, after division by $c$, the statements (ii) and (iii). Under exact tracking of $r$, the required effort is the left-hand side of \eqref{eq:common} evaluated on $r$, which scales by $c$, giving (iv).
\end{proof}

\begin{proof}[Proof of Proposition~\ref{prop:response-structure}]
Necessity follows by differentiating $\hat a=\widehat M^{-1}(u-c[\widehat M]-\nabla V-d)$ with respect to $u$. The Christoffel term is even and quadratic in $v$, while $d$ is odd and vanishes at zero velocity. Even/odd decomposition gives \eqref{eq:response-decomposition}; the gradient is curl-free and dissipativity gives the power inequality. Conversely, the curl condition on the simply connected domain yields $V$ with $\nabla V=\widehat g$, unique up to a constant. The decomposition is an identity:
\[
-\widehat M\hat a(s,v,0)=\widehat g+\widehat c_{\rm even}+\widehat d_{\rm odd}.
\]
Input affinity and $A=\widehat M^{-1}$ then recover \eqref{eq:reduced} with $B_S=I$, the Christoffel force from \eqref{eq:metric-christoffel}, and $d=\widehat d_{\rm odd}$. The standard kinetic-energy identity gives $\dot H=u^\top v-v^\top d$. Finally, \eqref{eq:inverse-sensitivity} follows from the resolvent identity $(A+\Delta A)^{-1}-A^{-1}=-(A+\Delta A)^{-1}\Delta A A^{-1}$ and the eigenvalue bound $\lambda_{\min}(A+\Delta A)\ge m_A-\eta$.
\end{proof}

The curvature \eqref{eq:curvature} follows from the standard formula for the Gaussian curvature of a two-dimensional metric and was verified symbolically.

\section{Numerical Checks}\label{app:numerics}
This section verifies two analytical results numerically.

\paragraph*{Information in the tracking channel} We integrate the common law \eqref{eq:common} with $I=0.05$\,kg$\cdot$m$^2$, $h=0.02$\,kg$\cdot$m, $b=0.01$\,\Nm$\cdot$s, $f=0.02$\,\Nm, and $\varepsilon=0.05$\,rad/s under PD tracking of $r(t)=0.8\sin(\pi t)+0.3\sin(2.6\pi t)$ with $K_d=2\sqrt{K_pI}$ fixed at the nominal inertia when differentiating in $c$, using a stiff solver (Radau, relative tolerance $10^{-11}$), and compute sensitivities to the scale $c$ by central differences ($h=10^{-4}$) on samples in $[0.5,4]$\,s. Figure~\ref{fig:checks}b compares them with Corollary~\ref{cor:ladder}. The relative deviations decay as $K_p^{-1/2}$, as Theorem~\ref{thm:channels} predicts, and the normalized information $K_p^2\sum_j(\partial q/\partial c)^2/\sum_j\tau_\lambda^2$ approaches one, from 0.860 at $K_p=10^2$ to 0.9987 at $10^5$.

\paragraph*{Discrete invariance under a nonlinear map} Mechanism $B$ obeys $\ddot x+\sin x+0.2\dot x=u$, and $A$ is defined from $B$ by (T1)--(T4) with $\psi(s)=s+0.3\sin s$ and $K=1$. For the arbitrary predicted sequence $\hat s(t)=0.9\sin(1.7t)+0.2\cos(3.1t)$ under $u=0.4\cos t$ on $[0,4]$\,s, Fig.~\ref{fig:checks}c uses the reduced midpoint residual with centered nodal velocities. The discrepancy decreases by a factor of approximately four on each halving of $\Delta t$, from $2.9\times10^{-4}$ at $\Delta t=0.01$\,s to $4.6\times10^{-6}$ at $1.25\times10^{-3}$\,s, while the residual itself converges to a nonzero force-level quantity, 4.692. This checks Proposition~\ref{prop:discrete}(ii) and the normalization in \eqref{eq:discrete}.

\end{document}